\documentclass[10pt,journal,compsoc]{IEEEtran}

\newif\ifshowtodos
\showtodosfalse

\newif\ifFORM
\FORMfalse

\newif\iftr
\newif\ifcnf

\trtrue
\cnffalse

\newif\ifnohl
\nohlfalse

\newif\ifsq     

\newif\ifsqCAP
\newif\ifsqVS
\newif\ifsqEN
\newif\ifsqTIT

\newcommand{\ignore}[1]{}

\sqtrue
\sqCAPfalse
\sqENtrue
\sqVStrue
\sqTITtrue

\sqfalse
\sqCAPfalse
\sqENfalse
\sqVSfalse
\sqTITfalse

\usepackage{etex}
\usepackage{balance}
\usepackage{epstopdf}
\usepackage{placeins}
\usepackage{comment}

\usepackage{pdflscape}

\usepackage{graphicx}
\usepackage{float}
\usepackage{multirow}
\usepackage{rotating}
\usepackage{makecell}
\usepackage{tabulary}
\usepackage{parcolumns}
\usepackage{tikz}
\usetikzlibrary{tikzmark}

\usepackage{xpatch}
\expandafter\xpatchcmd
\csname pgfk@/tikz/every picture/.@cmd\endcsname
{\thepage}{\arabic{page}}{}{}

\tikzstyle{comment} = [draw, fill=blue!70, text=white, text width=3cm, minimum height=1cm, rounded corners, align=left, font=\scriptsize]
\tikzstyle{background_alg} = [draw, fill=blue!20, opacity=0.4, inner sep=4pt, rounded corners=2pt]

\usetikzlibrary{shapes}
\usetikzlibrary{plotmarks}
\usetikzlibrary{calc, fit}

\usepackage{enumitem}

\usepackage{amsthm}
\usepackage{amsmath,amssymb,amsfonts}
\usepackage{mathtools,mathrsfs}

\usepackage{soul}
\usepackage{fontawesome}
\usepackage{pifont}
\usepackage{textcomp}
\usepackage{booktabs}
\usepackage{url}
\usepackage{pbox}
\usepackage[normalem]{ulem}
\usepackage[10pt]{moresize}

\ifsqCAP
\usepackage[font={normalfont, scriptsize}]{caption}
\usepackage[font={normalfont, scriptsize}]{subcaption}
\else
\usepackage[font={normalfont, scriptsize}]{caption}
\usepackage[font={normalfont, scriptsize}]{subcaption}
\fi

\newcommand{\vspaceSQ}[1]{\ifsqVS\vspace{#1}\fi}
\newcommand{\enlargeSQ}[1]{\ifsqEN\enlargethispage{\baselineskip}\fi}

\ifsqTIT
\usepackage[compact]{titlesec}
\titlespacing*{\section}{0pt}{3pt}{-1pt}
\titlespacing*{\subsection}{0pt}{0pt}{-3pt}
\titlespacing*{\subsubsection}{0pt}{2pt}{1pt}
\fi

\usepackage{xcolor}
\definecolor{darkgrey}{RGB}{70,70,70}
\definecolor{lightgrey}{RGB}{200,200,200}
\definecolor{lyellow}{RGB}{255,255,100}
\definecolor{llyellow}{RGB}{250,250,180}
\definecolor{lgreen}{RGB}{144,238,144}
\definecolor{raphael_comments}{RGB}{13, 145, 24}

\usepackage[customcolors]{hf-tikz}
\hfsetbordercolor{white}
\hfsetfillcolor{vlgray}

\definecolor{vlgray}{rgb}{0.77 0.77 0.77}
\definecolor{ablack}{rgb}{0.2 0.2 0.2}
\definecolor{vllgray}{rgb}{0.9 0.9 0.9}
\definecolor{bblue}{rgb}{0.7 0.7 0.99}

\usepackage{colortbl}

\usepackage{inconsolata}
\usepackage{listings}

\ifsq
\else
\fi

\newcommand{\maciej}[1]{\textcolor{blue}{[Maciej: #1]}}

\newcommand{\greg}[1]{\textcolor{cyan}{[Greg: #1]}}
\newcommand{\tomasz}[1]{\textcolor{violet}{[Tomasz: #1]}}
\newcommand{\robert}[1]{\textcolor{teal}{[Robert: #1]}}

\newcommand{\julia}[1]{\textcolor{magenta}{[Julia: #1]}}

\newcommand{\eric}[1]{\textcolor{brown}{[Eric: #1]}}

\definecolor{hlL}{rgb}{0.8 0.8 0.99}

\newcounter{highlight}

\newcounter{hlLR}

\newcounter{hlLIR}

\newcounter{hlLIIR}

\newcounter{Ahighlight}

\usepackage{scalerel,stackengine}
\stackMath
\newcommand\rwh[1]{%
\savestack{\tmpbox}{\stretchto{%
  \scaleto{%
        \scalerel*[\widthof{\ensuremath{#1}}]{\kern-.6pt\bigwedge\kern-.6pt}%
                  {\rule[-\textheight/2]{1ex}{\textheight}}
                              }{\textheight}%
}{0.5ex}}%
\stackon[1pt]{#1}{\tmpbox}%
}

\usepackage[hang,flushmargin]{footmisc}

\renewcommand{\epsilon}{\ensuremath\varepsilon}

\renewcommand{\phi}{\ensuremath{\varphi}}

\DeclareMathOperator*{\argmax}{arg\,max}

\iftr
\usepackage{cite}
\fi
\usepackage{xurl}
\iftr
\usepackage[hidelinks]{hyperref}
\fi

\usepackage{fontawesome}
\usepackage{pifont}
\usepackage{textcomp}

\usepackage{xcolor}
\usepackage{soul}
\usepackage{dblfloatfix}

\usepackage{colortbl}

\usepackage{dblfloatfix}
\usepackage{eso-pic}

\renewcommand{\comment}[1]{\ignorespaces}

\usepackage{physics}

\nohltrue
\ifnohl

\renewcommand{\rowcolor}[1]{}
\renewcommand{\marginpar}[1]{}

\fi

\newcolumntype{y}{>{}l}

\IEEEaftertitletext{\vspace{-2\baselineskip}}

\newif\ifHL
\HLfalse

\newcommand{\faY}[0]{\faBatteryFull}
\newcommand{\faH}[0]{\faBatteryHalf}
\newcommand{\faN}[0]{\faTimes}

\IEEEoverridecommandlockouts

\usepackage{algorithm}
\usepackage{algorithmic}
\usepackage{comment}

\newcommand{\schemename}[0]{\textbf{x1}}
\newcommand{\schemenameS}[0]{\textbf{x1}\ }

\newcommand{\schemenameAS}[0]{\textbf{x1}\ }

\usepackage{bbm} 

\showtodosfalse

\begin{document}

%
%

\title{Reasoning Language Models: A Blueprint}

%

\author{Maciej Besta$^{1\dagger}$, Julia Barth$^1$, Eric Schreiber$^1$, Ales Kubicek$^1$, Afonso Catarino$^1$, Robert Gerstenberger$^1$, Piotr Nyczyk$^2$, Patrick Iff$^1$, Yueling Li$^3$, Sam Houliston$^1$, Tomasz Sternal$^1$, Marcin Copik$^1$, 
Grzegorz Kwa\'{s}niewski$^1$, Jürgen Müller$^3$,
Łukasz Flis$^4$,
Hannes Eberhard$^1$,
Zixuan Chen$^1$,
Hubert Niewiadomski$^2$, Torsten Hoefler$^1$\\
\vspace{0.5em}{\small $^\dagger$\textit{Corresponding author} \quad $^1$ETH Zurich \quad $^2$Cledar \quad $^3$BASF SE \quad $^4$Cyfronet AGH}}

\IEEEtitleabstractindextext{%
\begin{abstract}
Reasoning language models (RLMs), also known as Large Reasoning Models (LRMs), such as OpenAI's o1 and o3, DeepSeek-R1, and Alibaba's QwQ, have redefined AI's problem-solving capabilities by extending large language models (LLMs) with advanced reasoning mechanisms. Yet, their high costs, proprietary nature, and complex architectures—uniquely combining reinforcement learning (RL), search heuristics, and LLMs—present accessibility and scalability challenges. To address these, we propose a comprehensive blueprint that organizes RLM components into a modular framework, based on a survey and analysis of all RLM works. This blueprint incorporates diverse reasoning structures (chains, trees, graphs, and nested forms), reasoning strategies (e.g., Monte Carlo Tree Search, Beam Search), RL concepts (policy, value models and others), supervision schemes (Outcome-Based and Process-Based Supervision), and other related concepts (e.g., Test-Time Compute, Retrieval-Augmented Generation, agent tools). We also provide detailed mathematical formulations and algorithmic specifications to simplify RLM implementation. By showing how schemes like LLaMA-Berry, QwQ, Journey Learning, and Graph of Thoughts fit as special cases, we demonstrate the blueprint’s versatility and unifying potential. To illustrate its utility, we introduce \schemename, a modular implementation for rapid RLM prototyping and experimentation. Using \schemenameS and a literature review, we provide key insights, such as multi-phase training for policy and value models, and the importance of familiar training distributions. Finally, we discuss scalable RLM cloud deployments and we outline how RLMs can integrate with a broader LLM ecosystem. Our work demystifies RLM construction, democratizes advanced reasoning capabilities, and fosters innovation, aiming to mitigate the gap between ``rich AI'' and ``poor AI'' by lowering barriers to RLM design and experimentation.
\end{abstract}

}

\maketitle

\IEEEdisplaynontitleabstractindextext
\IEEEpeerreviewmaketitle

\iftr
\else
{\vspace{-1.0em}\noindent \textbf{An extended version: \url{http://arxiv.org/abs/2501.11223}}\vspace{1em}}
\fi


\section{Introduction}

\begin{figure*}[t]
\centering
\vspaceSQ{-0.5em}
\AddToShipoutPictureBG*{%
  \AtPageLowerLeft{%
    \put(\LenToUnit{0.5\paperwidth},\LenToUnit{3.6cm}){%
      \makebox[0pt]{
        \begin{minipage}{\textwidth}
          \centering
            \ifcnf
            \includegraphics[width=0.9\textwidth]{PDFs/summary_v4-cnf.pdf}
            \else
            \includegraphics[width=0.9\textwidth]{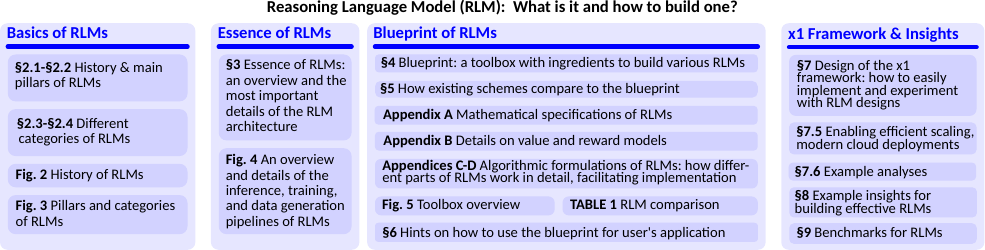}
            \fi
            \vspace{-0.75em}
            \ifcnf
            \captionof{figure}{\textbf{Summary of the contributions made by this paper. The \schemenameS framework can be found at https://github.com/spcl/x1.}}
            \else
            \captionof{figure}{\textbf{Summary of the contributions made by this paper. The \schemenameS framework can be found at \url{https://github.com/spcl/x1}.}}
            \fi
            \label{fig:packaging}
        \end{minipage}
      }
    }
  }
}
\vspaceSQ{-1.5em}
\label{fig:summary}
\end{figure*}

Reasoning Language Models (RLMs), such as OpenAI's o1~\cite{openai2024introducing}, o3~\cite{knight2024openai}, DeepSeek-R1~\cite{deepseek2024v3} and Alibaba's QwQ~\cite{qwen2024qwq}, also referred to as Large Reasoning Models (LRMs)\footnote{We use the term ``Reasoning Language Model'' instead of ``Large Reasoning Model'' because the latter implies that such models are always large. This does not necessarily have to be the case -- as a matter of fact, smaller RLM can outperform larger LLMs~\cite{guan2025rstar}.
\iftr
\vspace{20em}
\else
\vspace{16em}
\fi
},
represent a transformative breakthrough in AI, on par with the advent of ChatGPT~\cite{openai2022introducing}. These advanced systems have fundamentally redefined AI's problem-solving capabilities, enabling nuanced reasoning, improved contextual understanding, and robust decision-making across a wide array of domains, reshaping science~\cite{naqa2020artificial}, industries~\cite{burstrom2021ai}, governance~\cite{giest2024more}, and numerous other aspects of human life~\cite{kumar2019artificial, elliot2018culture, kislev2022relationships, tadeusiewicz2006cognitive, tadeusiewicz2008modern}. By extending the capabilities of standard large language models (LLMs) with sophisticated reasoning mechanisms, RLMs have emerged as the new cornerstone of cutting-edge AI, bringing us closer to AGI.

{
However, the high cost and proprietary nature of state-of-the-art RLMs, such as those developed by OpenAI, risk exacerbating the divide between ``rich AI'' and ``poor AI'', raising significant concerns about accessibility and equity.
Even the publicly available QwQ only comes with its model weights, and Alibaba does not disclose details about their training or data generation methodologies. Businesses and individuals unable to afford these advanced systems face a growing disadvantage, threatening to stifle innovation and reinforce systemic inequities. As RLMs become integral to critical applications, from healthcare to science,
\ifcnf
management, and beyond, it is imperative to address these
\else
manage-\parfillskip=0pt
\newpage
\noindent
ment, and beyond, it is imperative to address these
\fi
disparities and ensure that the benefits of advanced reasoning capabilities are broadly accessible.
}

\begin{figure*}[t]
\centering
\setcounter{figure}{1}  
\vspaceSQ{-0.5em}
\includegraphics[width=1.0\textwidth]{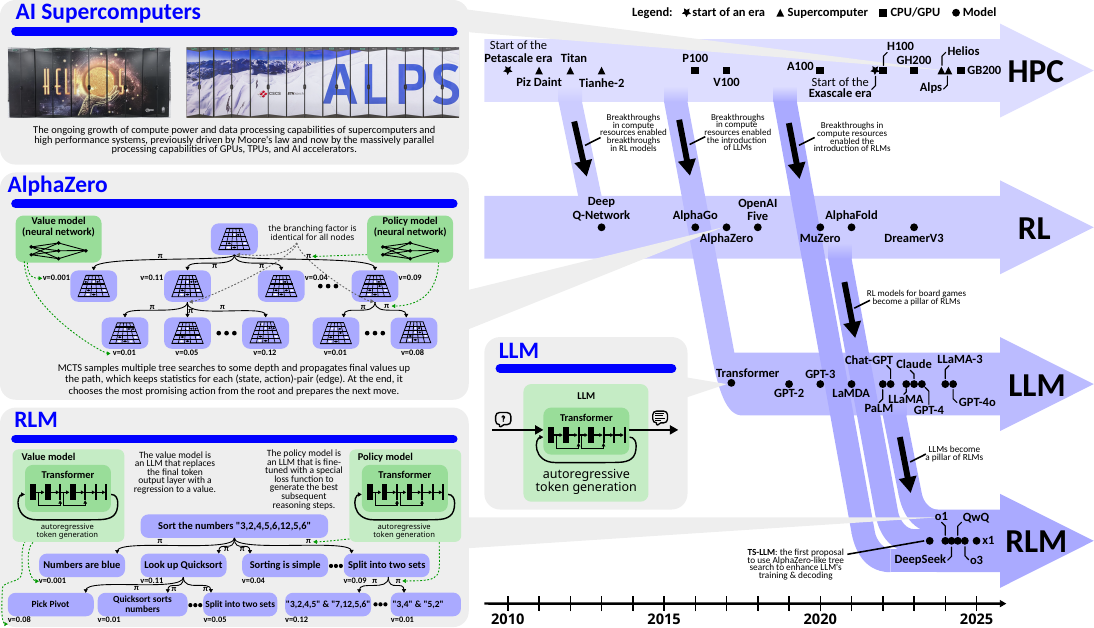}
\vspace{-1.0em}
\caption{\textbf{The history of RLMs.} This class of models has been the result of the development of three lines of works: (1) reinforcement learning based models such as AlphaZero~\cite{silver2018general}, (2) LLM and Transformer based models such as GPT-4o~\cite{openai2024hello}, and (3) the continuous growth of compute power and data processing capabilities of supercomputers and high performance systems.}
\vspaceSQ{-1.5em}
\label{fig:history}
\setcounter{figure}{0}  
\end{figure*}

\ifcnf
\newpage
\fi

The technical foundations of RLMs remain opaque and complex, compounding the accessibility challenge.
Emerging analyses suggest that their design likely integrates elements such as Monte Carlo Tree Search (MCTS) or Beam Search, reinforcement learning (RL), Process-Based Supervision (PBS)~\cite{uesato2022solving, lightman2023let, uesato2022solving, lightman2023let}, and advanced in-context learning (ICL) techniques like Chain-of-Thought (CoT)~\cite{wei2022chain} or Tree of Thoughts (ToT)~\cite{yao2023tree}, and possibly even retrieval-augmented generation (RAG)~\cite{li2025search, lewis2020retrieval, guu2020realm, besta2024multi}.

Additionally, these architectures employ multiple specialized subcomponents—such as synthetic data generation engines and policy, value, and reward models—trained through some form of novel loss functions and possibly several fine-tuning schemes.
However, the intricate interplay of these components and their integration into a cohesive and effective architecture remains poorly understood. Here, \textit{the ``holy-grail question'' is: what is the detailed design of an RLM and how to make it simultaneously achieve effectiveness (i.e., high accuracy in delivered answers), low cost, and scalability?}

To help answer this question and to address the above challenges, we propose a comprehensive blueprint for constructing, analyzing, and experimenting with RLMs (\textbf{contribution~\#1}\iftr; a roadmap of all the contributions and the paper is in Figure~\ref{fig:packaging}\fi). Our approach identifies and crystallizes the fundamental building blocks of RLMs, organizing them into a cohesive framework. This blueprint is presented with increasing levels of granularity, starting from high-level overview, finishing at low-level details that can be directly harnessed when implementing. Further, to maximize the clarity and comprehensiveness, we present the blueprint using three perspectives: (1) architecture diagrams and descriptions, (2) detailed mathematical formulations, and (3) in-depth algorithmic specifications. By employing these complementary perspectives, we aim to provide a clear and actionable guide for developing RLMs tailored to specific applications, settings, and constraints.

Our blueprint comprehensively encompasses the potential building blocks of RLMs, offering a flexible and modular framework. It incorporates a variety of reasoning structures, such as chains, trees, graphs, and even higher-order structures such as hierarchical (or nested) trees, along with numerous operations that transform and advance the reasoning process. The blueprint supports different granularities of reasoning steps, ranging from individual tokens to full sentences or structured segments. Additionally, it enables diverse training schemes, including Outcome-Based Supervision (OBS) and PBS, and the related Outcome \& Process Reward Models (ORMs \& PRMs). Next, in order to illustrate the capability of the blueprint to accommodate novel design ideas, we describe several novel schemes and how they fit within the blueprint. One such example is Trace-Based Supervision (TBS), which extends PBS by incorporating labeled traces of traversal paths through entire reasoning structures, rather than just linear chains of reasoning steps.
By unifying all these components, our blueprint serves as a versatile toolbox for constructing RLMs—ranging from simple models to sophisticated designs—tailored to specific reasoning tasks and performance objectives.

\begin{figure*}[t]
\centering
\setcounter{figure}{2} 
\vspaceSQ{-0.5em}
\includegraphics[width=1.0\textwidth]{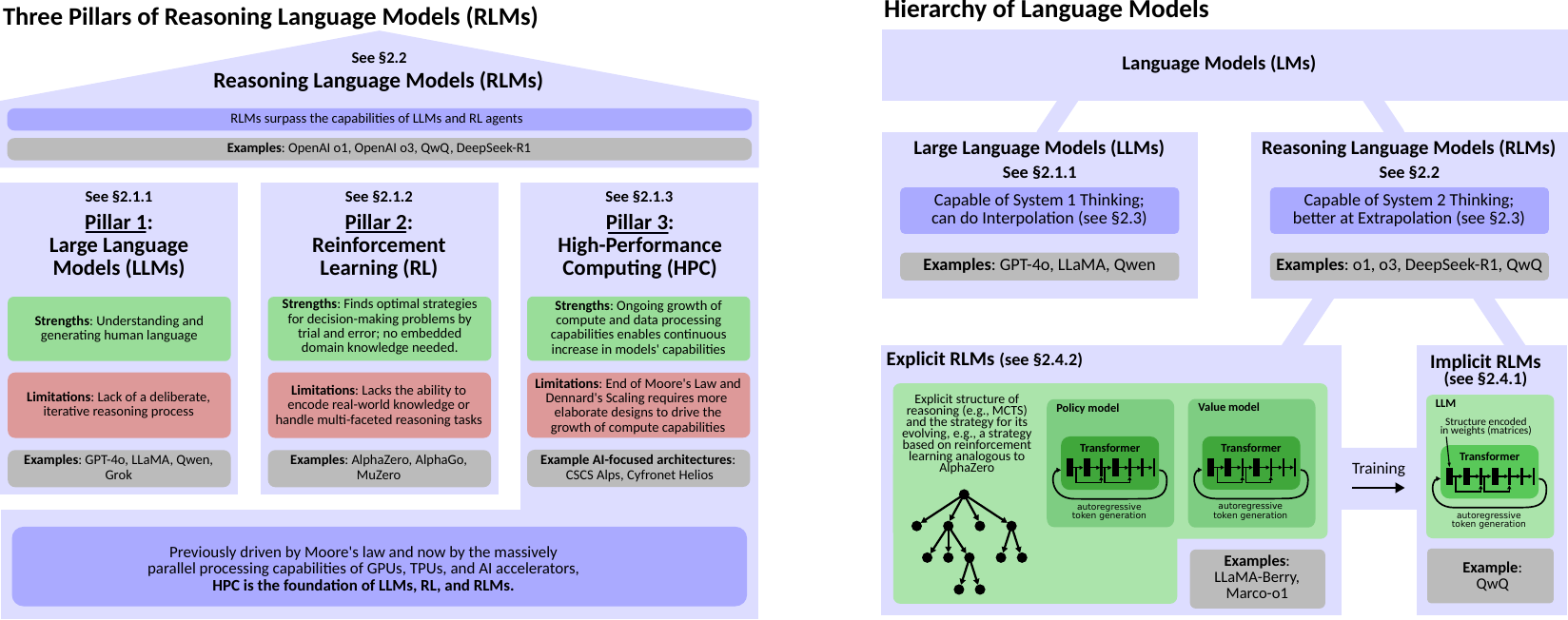}
\vspace{-1.5em}
\caption{\textbf{Hierarchy of language models (right) and the three pillars of RLMs (left).}}
\vspaceSQ{-1.5em}
\label{fig:pillars}
\end{figure*}

We conduct a broad analysis of existing reasoning schemes (\textbf{contribution~\#2}), demonstrating how they fit into our blueprint as special cases. This analysis encompasses not only standard MCTS and reinforcement learning-based designs, such as LLaMA-Berry~\cite{zhang2024llamaberry}, but also models like QwQ~\cite{qwen2024qwq}. Additionally, we include paradigms diverging from standard MCTS, such as Journey Learning~\cite{qin2024o1} or Beam Search, which redefines reasoning through implicit long-chain structures, and advanced structured prompting techniques like CoT~\cite{wei2022chain}, ToT~\cite{yao2023tree}, and Graph of Thoughts~\cite{besta2024graph}. We also consider \textit{reasoning utilities} such as Retrieval-Augmented Generation (RAG) and data stores, tools, and others. By mapping these diverse approaches to one blueprint, we showcase its versatility and expressive power, highlighting its ability to unify a wide range of reasoning methodologies within a coherent framework.



To demonstrate the utility of our framework, we introduce \schemename, a modular and user-friendly implementation\footnote{\url{https://github.com/spcl/x1}} designed to simplify the process of developing and experimenting with new RLM architectures, covering not only training and inference, but also synthetic data generation (\textbf{contribution~\#3}). We design \schemenameS to facilitate supporting various optimizations, design decisions, and overall scalability, such as batch processing, making it a well-suited foundation of experimentation infrastructure. 
We also discuss key aspects of deployment in cloud environments, ensuring that \schemenameS can be seamlessly integrated into modern infrastructure for both research and production use cases.
%
%

By providing both theoretical insights and practical tools, this work aims to democratize access to advanced RLMs, enabling researchers and practitioners to design, train, and deploy sophisticated reasoning models with reduced complexity and cost. Our blueprint offers a clear and adaptable framework that lowers the barriers to entry, fostering broader experimentation and innovation. Additionally, the modular implementation of \schemenameS serves as a foundation for rapid prototyping and large-scale experimentation, empowering users to explore new reasoning paradigms and optimize performance across diverse applications. By bridging the gap between conceptual advancements and practical implementations, this work seeks to accelerate progress in the field, unlock new possibilities for intelligent systems across research, industry, and education, and to mitigate the risk of the growing gap between ``rich AI'' and ``poor AI''.


\section{Evolution \& Foundations of RLMs}

We first summarize the evolution and foundations of reasoning language models. Figure~\ref{fig:history} shows an overview of the history of the development of these models.


\subsection{Basic Pillars of Reasoning LMs}

The development of reasoning-capable LLMs represents a convergence of three critical threads: (1) advances in LLMs such as GPT-4, (2) RL designs such as AlphaZero, and (3) High-Performance Computing (HPC) resources. Together, these threads have shaped models capable of efficient \textit{System 2 Thinking} -- a level of reasoning that combines explicit deliberation with novel problem-solving abilities, distinct from the intuitive, fast, and automatic heuristics of \textit{System 1 Thinking}. Figure~\ref{fig:history} compares example designs in these pillars while Figure~\ref{fig:pillars} (left side) further discusses the details of these pillars.

\subsubsection{Large Language Models: A Reservoir of Knowledge}

LLMs such as GPT-4o~\cite{openai2024hello} or LLaMA~\cite{grattafiori2024llama} represent an extraordinary leap in the field of AI, constituting a vast repository of world knowledge encoded directly in their weights. Trained on huge corpora of text from diverse sources, LLMs are capable of understanding and generating human language with remarkable fluency. However, their reasoning abilities largely align with the fast, automatic, and intuitive {System 1 Thinking}. While they can generate coherent responses and even perform simple reasoning tasks, LLMs have limitations. The reasoning they exhibit is often shallow, rooted in the simple mechanism of predicting the next most probable token in a sequence rather than engaging in explicit problem-solving or structured analysis. While LLMs may generate plausible-sounding solutions to a problem, these outputs are the result of statistical language modeling rather than a deliberate, iterative reasoning process. This distinction highlights the need for integrating more advanced mechanisms capable of explicit reasoning into AI systems—paving the way for hybrid designs that combine the knowledge-rich foundation of LLMs with structured reasoning methodologies.

\subsubsection{Reinforcement Learning: Exploring and Innovating}

RL has historically provided a framework for decision-making and exploration in environments where an agent must learn optimal strategies through trial and error. Landmark systems like AlphaZero~\cite{silver2018general} and a long line of others such as AlphaGo~\cite{silver2016mastering} or MuZero~\cite{schrittwieser2020mastering} demonstrated the profound potential of RL by achieving superhuman performance in games such as chess, shogi, and Go. Unlike traditional AI systems, AlphaZero began with no embedded domain knowledge. Instead, it mastered these games purely through self-learning, discovering novel strategies that even human experts had not considered.

One of the most striking examples of RL's innovative capacity came during an AlphaZero match, where the system made a move initially deemed a mistake by human observers. This move~\cite{metz2016two} later proved to be both surprising and strategically brilliant, illustrating the capacity of RL agents to explore unconventional solutions that lie outside the bounds of human intuition. Such capabilities are fundamentally rooted in RL's ability to navigate vast search spaces effectively.

However, traditional RL systems lacked the ability to encode real-world knowledge or handle complex, multi-faceted reasoning tasks. This limitation spurred the integration of RL principles with LLMs, combining the structured exploration and optimization capabilities of RL with the knowledge-rich reasoning foundation of language models.

\subsubsection{HPC: Scalability \& Efficiency}

The growth of LLM and RL systems has been propelled by advancements in High-Performance Computing (HPC). Initially driven by Moore's Law, which enabled a doubling of transistor density approximately every two years, HPC benefited from both technological advancements and the economic feasibility of manufacturing smaller transistors. However, as the costs of further miniaturization have risen sharply, Moore's Law has reached practical limits, necessitating alternative strategies like parallelism and heterogeneous computing.

Modern HPC systems rely heavily on GPUs, TPUs, and AI accelerators for their parallel processing capabilities, alongside CPUs for sequential and general-purpose tasks. Heterogeneous computing leverages these components to optimize task-specific performance. Distributed frameworks, employing techniques such as data, model, and pipeline parallelism~\cite{ben2019demystifying, besta2023parallel, besta2023high}, further enable the training of enormous models across thousands of compute nodes.

Energy efficiency innovations, including sparsity, quantization, and pruning, mitigate the growing energy demands of scaling AI systems. These advancements ensure that HPC remains a cornerstone for developing and deploying AI models, supporting the combination of vast knowledge, reasoning capabilities, and computational scalability -- allowing AI evolution to continue beyond the limits of traditional Moore's Law scaling.

\subsection{The Convergence: System 2 Thinking in AI}

The intersection of these three threads -- LLMs, RL, and HPC -- has culminated in the emergence of models capable of System 2 Thinking. These advanced systems combine the knowledge-rich foundation of LLMs with the exploratory and optimization capabilities of RL, all supported by the scalability and performance of modern HPC. The result is a new class of AI models that can engage in explicit, deliberate reasoning processes.

These models possess a world model encoded in the weights of their LLM components, allowing them to reason about complex scenarios and contexts. Their RL capabilities combined with the HPC capabilities enable them to navigate truly immense decision spaces, evaluate multiple strategies, and iteratively refine solutions.

\subsection{Interpolation (LLMs) vs.~Extrapolation (RLMs)}

Standard LLMs, driven by their autoregressive token prediction mechanism, primarily perform interpolation within the vast search space of solutions. They excel at generating responses that align with patterns seen in their training data, effectively synthesizing knowledge from known contexts. However, this process limits them to producing outputs that remain within the boundaries of their training distribution. In contrast, reasoning LMs enable extrapolation beyond these boundaries. By combining structured exploration, reasoning LMs navigate uncharted areas of the solution space, generating novel insights and solutions that extend past the limits of their training data. This enables a shift from basic pattern completion to active problem-solving.

\subsection{Hierarchy of Reasoning-Related Models}


The evolution of RLMs can be understood as a hierarchical progression, with earlier models such as GPT-4o being less capable in terms of reasoning, and the o1-like architectures demonstrating increasing sophistication and explicit reasoning abilities. This hierarchy reflects the integration of System 1 (LLMs) and System 2 (RLMs) Thinking.
RLMs can be further divided based on how reasoning is implemented into \textit{Implicit RLMs} and \textit{Explicit RLMs}; the details of this categorization can be found in Figure~\ref{fig:pillars} (the right side).

\subsubsection{Implicit Reasoning Models}

In this subclass, the reasoning structure is embedded entirely within the model's weights. Models such as QwQ~\cite{qwen2024qwq} operate as ``black boxes'', where reasoning is implicit and cannot be explicitly disentangled or manipulated. While these models exhibit improved reasoning capabilities compared to standard LLMs, their reasoning processes are opaque and rely on the internalized patterns learned during training.

\subsubsection{Explicit Reasoning Models}

These models introduce explicit reasoning mechanisms external to the model's core weights. Examples include designs such as LLaMA-Berry~\cite{zhang2024llamaberry}, Marco-o1~\cite{zhao2024marco}, and potentially OpenAI's o3, which incorporate mechanisms like explicit MCTS combined with RL for decision-making. This explicit structure enables the model to simulate, evaluate, and refine solutions iteratively, facilitating novel problem-solving and extrapolation. By separating reasoning from the static knowledge encoded in the weights, these models achieve greater flexibility and interpretability in their reasoning processes.
Note that the explicit reasoning can be internalized via training making it implicit -- we discuss it later in the blueprint.

\begin{figure*}[hbtp]
\centering
\vspace{-1.5em}
\includegraphics[width=0.98\textwidth]{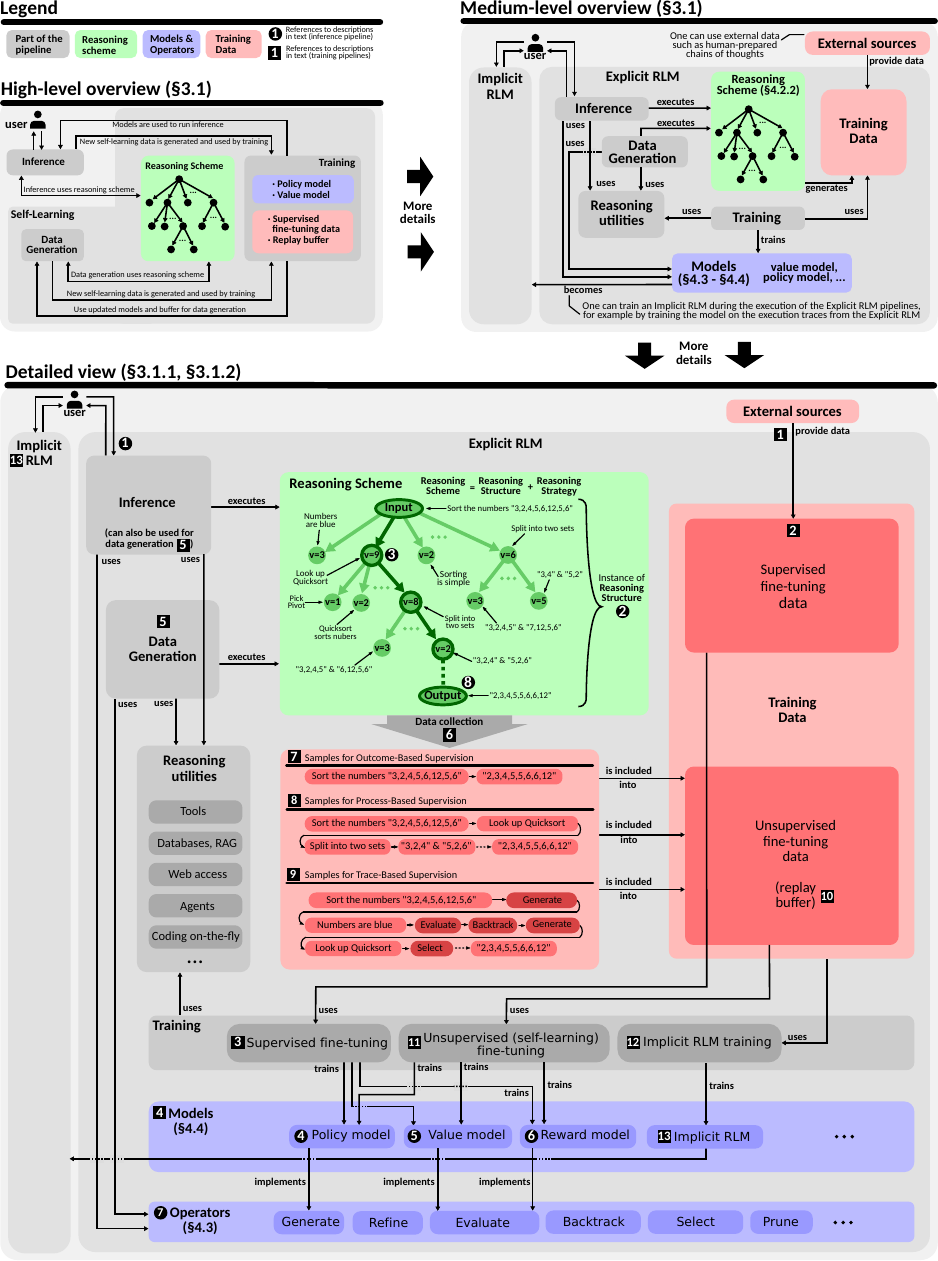}
\vspace{-1em}
\caption{\textbf{Overview of a general RLM design and core concepts.} We provide a high-level overview \textbf{(the top-left part)}, a more detailed medium-level overview \textbf{(the top-right part)}, and a very detailed diagram showing the inference and training pipelines \textbf{(the bottom part)}. A detailed specification of the inference pipeline can be found in \iftr Appendix~\ref{sec:mcts_algo_description} and in Algorithm~\ref{alg:mcts_star}\else Appendix~C.1 and in Algorithm~1\fi. Details on the pipelines for different training phases and paradigms can be found in \iftr Appendices~\ref{app:phase1_training} and~\ref{sec:phase2-appendix} as well as in Algorithms~\ref{algo:sft_policy_model}--\ref{algo:value_model_update}\else Appendices~C.2 and C.3 as well as in Algorithms~2--7\fi.
The data generation pipeline is detailed in \iftr Appendix~\ref{sec:appendix_data}\else Appendix D\fi.}
\vspaceSQ{-1.5em}
\label{fig:basic-pipeline}
\end{figure*}

\section{Essence of Reasoning LMs}
\label{sec:essence-basic}

We now describe the general architecture of RLMs, which we summarize in Figure~\ref{fig:basic-pipeline}.
%
In the following sections, we generalize this description to the full RLM blueprint. 
%
%


\subsection{Basic Architecture, Pipelines, \& Concepts}
\label{sec:pipeline}

We now outline the foundational architecture, operational pipelines, and core concepts. 
Figure~\ref{fig:basic-pipeline} offers three levels of detail. In general (the top-left part), the whole RLM architecture consists of three main pipelines: inference, training, and data generation. The inference serves user requests, using models (e.g., the value or policy model) provided by the training pipeline. Data generation mirrors the inference pipeline in its internal design; the main difference is that it runs independently of the user requests, generating data that is then used to re-train the models. As such, training combined with data generation from various domains~\cite{saparov2023language, zhang2024accessing} offers \textit{self-learning} capabilities and is analogous to the self-play setting of AlphaZero~\cite{silver2018general}.

\subsubsection{Inference}

The inference process begins when the user provides an input prompt \includegraphics[scale=0.2,trim=0 16 0 0]{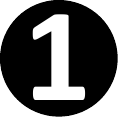}, which typically describes the problem or question to be addressed by the RLM. This input serves as the root of the reasoning process and initiates the construction of a \textbf{reasoning structure} \includegraphics[scale=0.2,trim=0 16 0 0]{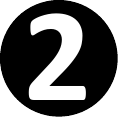} that organizes RLM's progress. The structure is usually represented as a tree. The root of this tree corresponds to the user's input, and subsequent nodes are generated to explore the search space -- the domain of possible reasoning paths or solutions. The purpose of this reasoning structure is to systematically investigate potential solutions, progressively refining and extending reasoning paths to converge on an optimal or satisfactory answer.

An individual \textbf{point} in the search space, represented as a \textbf{node} in the reasoning structure, corresponds to a \textbf{reasoning step} \includegraphics[scale=0.2,trim=0 16 0 0]{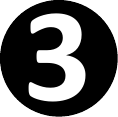}. A reasoning step is defined as a coherent and self-contained unit of thought -- a sequence of tokens that advances the solution by either exploring a new branch of the problem or building upon existing progress. These steps form the building blocks of the reasoning process.
%

The details of how the structure evolves are usually governed by the \textbf{MCTS scheme}, enhanced with \textbf{policy and value models} (we also distinguish other \textbf{reasoning strategies}, described below). This approach, inspired by methods used in AlphaZero, ensures that the search process is both efficient and directed toward promising solutions. The \textbf{policy model} \includegraphics[scale=0.2,trim=0 16 0 0]{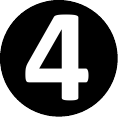} is responsible for generating new reasoning steps at each node, predicting the next most likely and logical steps to expand the reasoning process. Meanwhile, the \textbf{value model} \includegraphics[scale=0.2,trim=0 16 0 0]{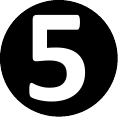} evaluates the quality of a reasoning path starting at a given node, helping the system prioritize the most promising steps to follow. Sometimes, a reward model\footnote{We use a naming scheme in which a model used to estimate the quality of a whole reasoning path starting at a given node, is called the \textit{value model}, while a model used to estimate the quality of a given reasoning step, is called the \textit{reward model}.}~\includegraphics[scale=0.2,trim=0 16 0 0]{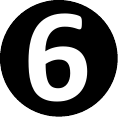} is used instead, to assess the quality of an \textit{individual} specific node and its corresponding reasoning step.
In our blueprint, as detailed in the next section, we abstract the models into a more general notion of \textit{operators}~\includegraphics[scale=0.2,trim=0 16 0 0]{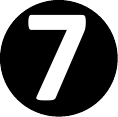} to enable more flexibility in how they are implemented.

The search and reasoning processes continue iteratively until a \textbf{terminal step} is reached \includegraphics[scale=0.2,trim=0 16 0 0]{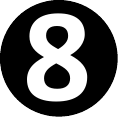}. This terminal step represents a completion of the reasoning chain that forms the final answer to the posed problem. It serves as the leaf node in the tree, concluding that particular reasoning path.

This architecture provides a unified framework that accommodates a wide range of reasoning tasks. Whether reasoning steps are fine-grained (e.g., individual token sequences) or coarse-grained (e.g., entire reasoning chains treated as single nodes), the architecture adapts seamlessly. By structuring the search space explicitly and guiding exploration with policy and value models, the RLM achieves a level of reasoning capability bridging intuitive pattern recognition and deliberate problem-solving.

A detailed specification of the inference pipeline can be found in \iftr Appendix~\ref{sec:mcts_algo_description} and in Algorithm~\ref{alg:mcts_star}\else Appendix~C.1 and in Algorithm~1\fi.








\subsubsection{Training}

Training details depend on what model is trained (value, policy, reward, ...). In general, we assume fine-tuning a model such as LLaMA. Here, we follow an approach where one first harnesses supervised data, usually coming from existing datasets such as PRM800K~\cite{lightman2023let} \includegraphics[scale=0.2,trim=0 16 0 0]{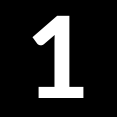}, which becomes a part of the supervised training data \includegraphics[scale=0.2,trim=0 16 0 0]{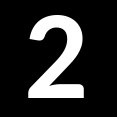} used in the \textbf{supervised training pipeline} \includegraphics[scale=0.2,trim=0 16 0 0]{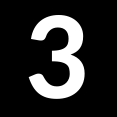} of the framework to train some, or all, of the models \includegraphics[scale=0.2,trim=0 16 0 0]{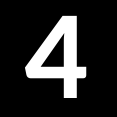} considered in the blueprint.
The second part of the overall training framework in RLMs is the \textbf{unsupervised (self-learning) training pipeline}, in which training data is being continually generated \includegraphics[scale=0.2,trim=0 16 0 0]{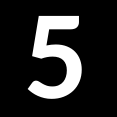} 
and used to improve the models. The data can be obtained from inference, assuming quality control~\cite{guan2025rstar}, but also from a dedicated synthetic data generation pipeline that mirrors that of the inference. To collect the data, one executes the respective RLM pipeline for a given input task and gathers the results \includegraphics[scale=0.2,trim=0 16 0 0]{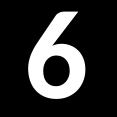}; depending on how detailed the gathering process is, the data collected can contain only \textbf{outcome-based labels}~\includegraphics[scale=0.2,trim=0 16 0 0]{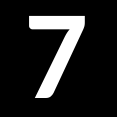}, \textbf{process-based labels} \includegraphics[scale=0.2,trim=0 16 0 0]{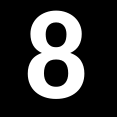}, or some other variant such as \textbf{trace-based labels}~\includegraphics[scale=0.2,trim=0 16 0 0]{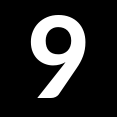} suggested in our blueprint, that generalize process-based samples to samples that contain also information about operators applied during the task solution process.
All this data becomes a part of the replay buffer~\includegraphics[scale=0.2,trim=0 16 0 0]{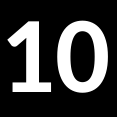} and is used in the unsupervised training scheme~\includegraphics[scale=0.2,trim=0 16 0 0]{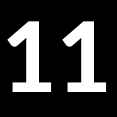} or it can also be used to train \includegraphics[scale=0.2,trim=0 16 0 0]{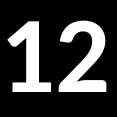} a model that would become an \textbf{Implicit RLM}~\includegraphics[scale=0.2,trim=0 16 0 0]{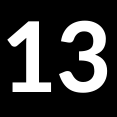}.

A detailed specification of the pipelines for different training phases and paradigms can be found in \iftr Appendices~\ref{app:phase1_training} and~\ref{sec:phase2-appendix} as well as in Algorithms~\ref{algo:sft_policy_model}--\ref{algo:value_model_update}\else Appendices C.2 and C.3 as well as in Algorithms 2–7\fi.
The data generation pipeline is detailed in \iftr Appendix~\ref{sec:appendix_data}\else Appendix~D\fi.

\subsection{Encompassing Diverse RLM Architectures}

The above-described design is applicable to many RLM designs. However, there are numerous other variants of architectures, some of which do not fully conform to this framework. In this section, we discuss these variants, highlighting how our blueprint accommodates such variations.

In some RLM designs~\cite{zhang2024llamaberry}, a single node in the MCTS tree could represent \textit{an entire reasoning structure}, such as a complete chain of reasoning steps. In this case, the action space involves transitioning between different reasoning structures rather than individual steps. This approach changes the nature of the search, as the focus shifts from iteratively constructing a single reasoning path to evaluating and refining entire structures within the search space. Our blueprint accommodates this with the concept of \textbf{nesting}, where a node in the reasoning structure can contain another reasoning structure.

Other architectures introduce even more novel paradigms. For instance, Journey Learning~\cite{qin2024o1} adds an additional layer of complexity by incorporating a transformation step that ``rewires'' the search or reasoning structure. This transformation consolidates multiple paths in the tree, synthesizing them into a new form that is used as input for subsequent reasoning iterations.

Despite these variations, our blueprint is sufficiently general to encompass all these cases and beyond, as we illustrate more formally in the following. This generality ensures that the blueprint is not only applicable to existing designs but also provides a foundation for future innovations in RLM development.

\subsection{Integration with Broader LLM Agent Ecosystems}

The integration of RLMs into broader LLM agent ecosystems would enable these models to interact dynamically with external tools, databases, and resources during execution. This interaction can occur within the inference or data generation pipeline, leveraging value or policy models to extend the reasoning process through access to retrieval-augmented generation (RAG), web queries, and specialized tools. For example, during a reasoning task, the value or the reward model could query a database to verify intermediate steps, ensuring factual correctness or retrieving additional context to refine its reasoning. Similarly, these models could utilize computational tools for mathematical or symbolic computations, thereby expanding the scope and accuracy of their reasoning.

\section{Blueprint for Reasoning LMs}
\label{sec:essence-general}

\begin{figure*}[hbtp]
\centering
\vspaceSQ{-0.5em}
\includegraphics[width=1.0\textwidth]{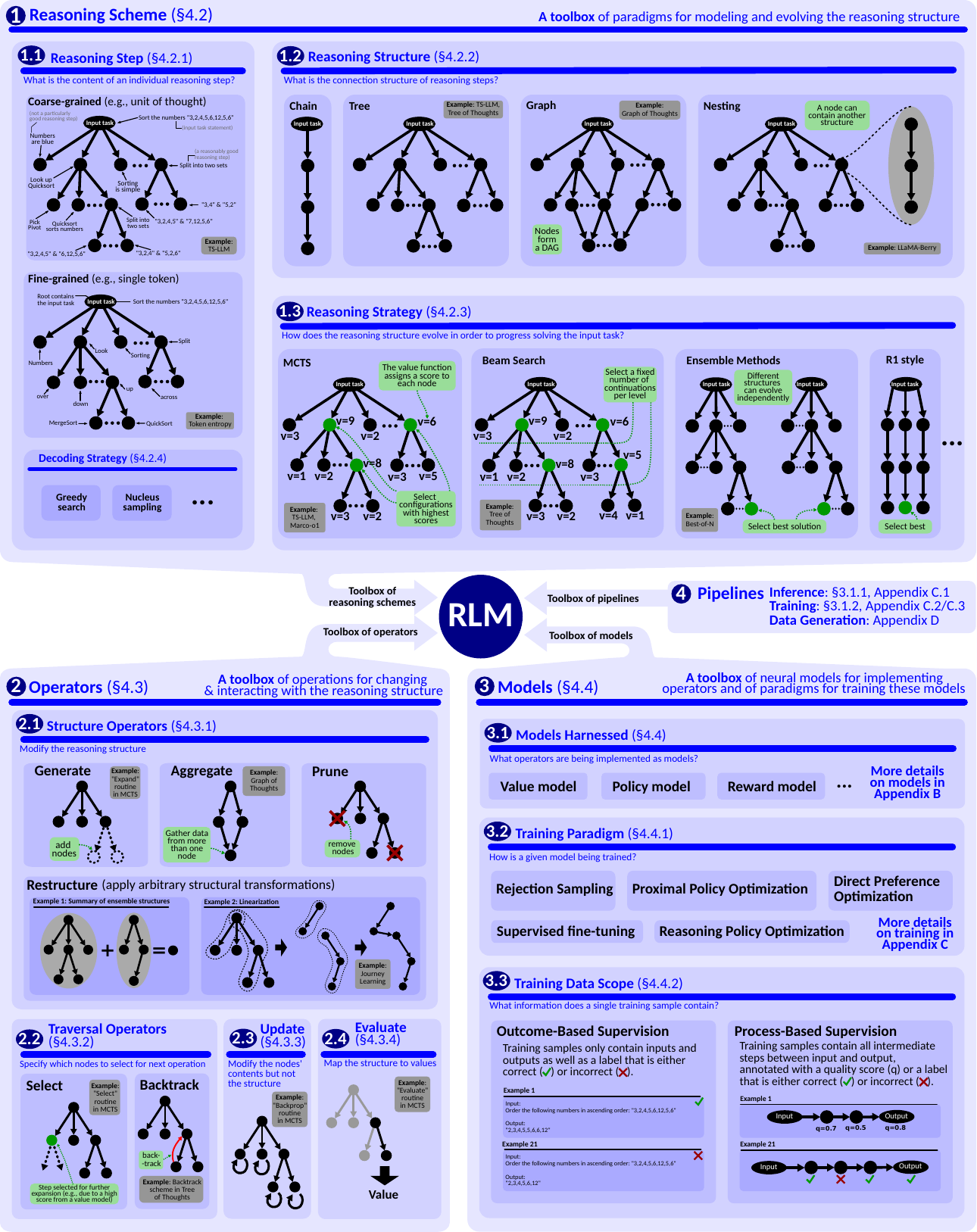}
%
%
\caption{\textbf{A blueprint for reasoning LMs.} It consists of four main toolboxes: the reasoning scheme \textbf{(the top part)}, operators \textbf{(the bottom-left part)}, and models \textbf{(the bottom-right part)}; pipelines are mentioned in the center and detailed in \iftr Appendix~\ref{sec:mcts_algo_description} and in Algorithm~\ref{alg:mcts_star} (the inference pipeline), Appendix~\ref{app:phase1_training}, Appendix~\ref{sec:phase2-appendix}, and in Algorithms~\ref{algo:sft_policy_model}--\ref{algo:value_model_update} (the training pipelines), and in Appendix~\ref{sec:appendix_data} \else Appendix~C.1 and in Algorithm~1 (the inference pipeline), Appendix~C.2, Appendix~C.3, and in Algorithms~2–7 (the training pipelines), and in Appendix~D\fi (the data generation pipeline).}
\vspaceSQ{-0.5em}
\label{fig:blueprint}
\end{figure*}

We now introduce our RLM blueprint that can be used to develop novel reasoning models and to provide ground for analysis, evaluation, and comparison of such designs. We overview the blueprint in Figure~\ref{fig:blueprint}.

\subsection{Overview \& Main Components}

The blueprint specifies a toolbox of components that can be used to build an arbitrary RLM.
We identify several classes of such components.
First, an RLM includes a \textbf{reasoning scheme}, which specifies a \textbf{reasoning structure} (e.g., a tree) together with a \textbf{reasoning strategy} (e.g., MCTS) of how this structure evolves in order to solve a given input task.
Second, there is a set of \textbf{operators} (e.g., refine) that can be applied to the reasoning structure (as specified by the reasoning strategy) in order to evolve it and make progress towards solving the input task. 
Operators are specified based on \textit{what they do} (i.e., what effect they have on the reasoning structure). \textit{How} this effect is achieved, depends on how a given operator is implemented. Here, many operators rely on neural \textbf{models} (e.g., policy model), which -- together with their training paradigms -- form the third class of the blueprint components.
Finally, we also distinguish a set of \textbf{pipelines}, i.e., \textit{detailed specifications of operations} that orchestrate the interaction between the reasoning scheme and the operators in order to achieve a specific objective, such as training, inference, or data generation.
\textit{Hence, an RLM can be defined as a composition of a reasoning scheme, a set of operators and associated models, and a set of pipelines.}



\subsection{Reasoning Scheme}
\label{sec:reasoning}

A reasoning scheme is the part of the blueprint that specifies the details of the reasoning steps progressing toward the solution, how they are interconnected to form coherent chains, trees, or more complex reasoning structures, and how these structures evolve in the course of solving the input task.

\subsubsection{Reasoning Step}

A reasoning step is a fundamental unit of the reasoning structure -- a sequence of tokens that advances the RLM towards the solution. Reasoning steps can vary in length, ranging from a \textbf{single token} to \textbf{entire segments} of text. The variability in their granularity depends on the user design choice. 
In existing schemes, a reasoning step is typically conceptualized as a ``coherent and self-contained unit of thought''. For instance, in mathematical proofs, this may correspond to an individual logical argument or deduction. 

The flexibility in defining reasoning steps allows models to adapt to different problem domains, balancing fine-grained and coarse-grained reasoning. Coarse steps, such as logical arguments (or even complete reasoning pathways~\cite{zhang2024llamaberry}), simplify preparation and adoption of training data, enhance interpretability, and -- as we discuss in Section~\ref{sec:principles} -- reduce computational overhead. On the other hand, single-token steps enable the utilization of concepts like token entropy~\cite{malinin2021uncertainty} to incorporate the model's uncertainty, as well as the integration of advanced decoding schemes (e.g., speculative decoding~\cite{leviathan2023fast} or contrastive decoding~\cite{li2023contrastive}) explicitly into the RLM design. Yet, while making the reasoning steps more fine-grained allows for a more detailed exploration of solution paths, this increased flexibility results in greater computational demands, particularly when combined with search algorithms such as MCTS.

\subsubsection{Reasoning Structure}

The reasoning structure specifies how individual reasoning steps are connected and organized. Common structures include chains (linear sequences), trees (hierarchical branching), and graphs (arbitrary connections). 

\textbf{Chains} are sequential reasoning flows, where each step builds directly on the preceding one. Chain structures are prevalent in CoT-based models, where each reasoning step follows logically from the previous step in a linear progression.
In \textbf{tree} structures, each reasoning step can branch into multiple continuations, forming a decision tree. This structure is commonly used in MCTS-based frameworks, where multiple potential paths are explored before selecting a branch that will be further investigated. It enables more effective exploration of the space of reasoning steps, but simultaneously makes the RLM design more complex and costly.
Finally, \textbf{graph} structures allow for arbitrary dependencies between reasoning steps, enabling graph-based reasoning, such as that found in the Graph of Thoughts (GoT) framework~\cite{besta2024graph}.

Further generalization involves \textbf{nested structures}, where reasoning nodes themselves may contain substructures. For example, a node in a tree structure may represent a CoT chain, as proposed in LLaMA-Berry~\cite{zhang2024llamaberry}. This hierarchical organization could be particularly useful for multi-step tasks where high-level decisions guide low-level computations, such as meta-reasoning frameworks~\cite{zhang2024llamaberry}. One could harness any other \textit{higher-order structures}, such as hypergraphs, motifs, and others~\cite{besta2023hot, besta2022motif, besta2021graphminesuite, besta2024demystifying}.

\subsubsection{Reasoning Strategy}

The reasoning strategy governs how the reasoning structure evolves, specifying the process by which new reasoning steps are added and integrated. Example strategies include: 

\begin{itemize}[noitemsep, leftmargin=0.75em]
\item \textbf{MCTS}~\cite{kocsis2006bandit} A popular approach that balances exploration and exploitation by simulating multiple reasoning paths and selecting the most promising one based on a scoring function.
\item \textbf{Beam Search}~\cite{snell2024scaling} A breadth-limited search that keeps a fixed number of top-ranked continuations at each step. While commonly used for decoding token sequences, beam search can also be applied to reasoning steps.
\item \textbf{Ensemble Methods} These methods involve aggregating multiple independent reasoning strategies, such as combining chains and trees to enhance robustness and accuracy. One example is {Best-of-N}~\cite{wang2023self, feng2023alphazero} -- a strategy where multiple independent reasoning paths are generated, and the most effective solution is selected based on predefined criteria, e.g., accuracy or completeness. Another example is tree ensemble (Forest)~\cite{bi2024forest} where, instead of a single reasoning tree, a reasoning ``forest'' consists of multiple disconnected trees, which may eventually converge at a shared solution node. This approach supports diverse reasoning pathways that parallelize exploration. 
\end{itemize}

\noindent
\textbf{Reasoning Strategy vs.~Decoding Strategy. }
It is crucial to distinguish reasoning strategies from token-level decoding strategies. While decoding strategies, such as greedy search and nucleus sampling~\cite{holtzman2020curious}, generate the internal token sequences within a reasoning step, reasoning strategies focus on the higher-level process of integrating and expanding reasoning steps within the reasoning structure.

\subsection{Operators}
\label{sec:operators}

Operators specify operations that can be applied to various parts of the reasoning structure to progress the reasoning process.
We now provide an extensive toolbox of operators. Many of them have been widely used in RLM-related designs, but some -- to our best knowledge -- are still unexplored, we include them to foster innovation and propel the design of more effective and more efficient RLMs.

\subsubsection{Structure Operators}

Structure operators transform the reasoning structure by taking it as input and producing a modified version, typically through addition of reasoning steps. For instance, they may add new children to a specific node, facilitating the exploration of alternative reasoning paths.

\begin{itemize}[noitemsep, leftmargin=0.75em]
\item \textbf{Generate} The generate operator adds one or more new reasoning steps to a reasoning structure. Within the MCTS reasoning strategy, this operator is typically implemented as a policy model to generate new steps. In other strategies, the generation operator may involve sequentially appending steps (CoT) or exploring multiple candidate steps in parallel (Beam Search).
%
%
%
\item \textbf{Aggregate} This operator combines multiple reasoning steps, paths, or structures into the next individual step. This enables consolidating information or improving coherence. It is used in Ensemble Methods~\cite{bi2024forest} or in Graph of Thoughts~\cite{besta2024graph}.
%
%
\item \textbf{Prune} This operator removes nodes or reasoning steps from the structure that are deemed suboptimal or irrelevant based on evaluation metrics. It enables optimizing the reasoning structure in order to, e.g., reduce token costs.
\item \textbf{Restructure} The restructure operator applies arbitrary transformations to the reasoning structure, enabling flexible reorganization of its components. A notable example is the conversion of a reasoning tree into a linear chain by rearranging its branches into a sequential series of steps, as done in Journey Learning~\cite{qin2024o1}. This restructuring facilitates the integration of insights from diverse branches into a cohesive flow, ``flattening'' it and making it easier for the model to process and utilize information within a single, unified context.
\end{itemize}

\noindent
\textbf{Discussion on Diversity }
In structure operators, there is a notion of how \textit{diverse} the outcomes of the operator are. For example, when generating $k$ new reasoning steps, one may want to make the contents of these steps as different to one another as possible. While different mechanisms to steer diversity exist, a typical approach is the use of the \textbf{policy model temperature}. We additionally propose to consider the \textbf{Diverse Beam Search}~\cite{vijayakumar2018diverse} which promotes diversity by maintaining multiple diverse candidate sequences during decoding.
In MCTS, there is also a distinction between exploitation (expanding the structure by applying generation operators within an already established tree branch) and exploration (generating new branches). Here, one impacts diversity by manipulating the \textbf{exploitation-exploration tradeoff}, as determined by the Upper Confidence Bound
for Trees (UCT) formula~\cite{kocsis2006bandit} or its variants.

\subsubsection{Traversal Operators}

Traversal operators define how the reasoning process navigates through the \textit{existing} reasoning structure. These operators play a crucial role in shaping the flow of reasoning by determining which paths to pursue.

\begin{itemize}[noitemsep, leftmargin=0.75em]
\item \textbf{Select} The select operator determines which reasoning steps to pick for further exploration, evaluation, or refinement within the reasoning process. It evaluates existing elements based on predefined criteria, such as heuristic scores, likelihood estimates, performance metrics or search strategies like PUCT~\cite{rosin2011multi} or UCT~\cite{kocsis2006bandit}, selecting the most promising candidates to guide the next stages of reasoning. By balancing exploration (considering diverse alternatives) and exploitation (focusing on high-potential paths), the selection operator optimizes resource allocation and ensures efficient reasoning progression.
%
%
\item \textbf{Backtrack} The backtrack operator enables the model to explicitly return to a previous reasoning step and continue along a different reasoning path. This operator supports error correction, divergence handling, and hypothesis revision by abandoning unproductive directions in favor of alternative trajectories. The QwQ model output indicates that the reasoning structures used as training data in this model harnessed backtrack.
\end{itemize}

\subsubsection{Update Operators}

The update operator enhances specific parts of the reasoning structure without altering the structure itself. A common example is the \textbf{backpropagation} phase in MCTS, where evaluation scores are propagated and updated along existing reasoning steps to inform future decisions. Another form of update involves \textbf{refining} the content of individual nodes or subsets of nodes, replacing their original versions with improved iterations. The refine operator could address ambiguities, correct errors, and optimize inefficiencies, resulting in a more robust version of the step~\cite{madaan2023self}. It could also integrate suggestions from self-critique~\cite{saunders2022self} (evaluates steps to identify weaknesses and suggest targeted improvements), summarization~\cite{zhu2025understanding} (condenses key elements into concise representations to streamline the reasoning structure), or rephrasing~\cite{deng2023rephrase} (reformulates steps to improve clarity and coherence while preserving their logical integrity). An example would be the ``improve'' thought transformation in Graph of Thoughts~\cite{besta2024graph}.

%

\subsubsection{Evaluate Operators}

Evaluate operators take as input a segment of the reasoning structure and output a value without any modifications to the structure. They are widely used with reasoning strategies, such as MCTS.

One important type of evaluation occurs when the reasoning structure reaches a terminal state, allowing the full reasoning sequence to be assessed against a known solution—applicable to tasks with definitive answers, such as mathematical problems. This \textbf{terminality evaluation} verifies whether the final step provides a correct and complete solution. 
%

One can also \textbf{evaluate intermediate steps} (i.e., non-terminal ones). This can involve estimating the reward associated with specific reasoning steps, using heuristics, aggregated simulation outcomes, or a trained reward model for more efficient assessments. Other methods such as embedding-based verification could also potentially be harnessed~\cite{besta2024checkembed}.

Another form of evaluation employs a \textbf{value estimator}, which judges a given reasoning step based on its expected contribution to a correct final outcome. This method evaluates both the correctness of the step and its alignment with the overall solution goal. Such evaluations can be performed through simulations, as in the original MCTS algorithm, or more efficiently using a learned value model~\cite{silver2017mastering}.

A critical aspect of evaluation is the selection of \textbf{appropriate metrics}. For instance, in value estimation, an ideal metric considers both the correctness of a reasoning step and the extent of progress it represents toward the final solution, ensuring a balanced assessment of its contribution.

\subsubsection{Discussion: Test-Time Compute}

One of the recent trends in next-generation LLMs~\cite{wang2024openr, mai2024improving} is to shift from merely increasing model sizes to enhancing computational strategies during inference, a concept known as the test-time compute (TTC). This approach allocates additional computational resources during a model's execution to improve performance, particularly in complex reasoning tasks. This methodology mirrors human cognitive processes, where increased deliberation is applied to more challenging problems. 

Recent studies~\cite{snell2024scaling} indicate that optimizing test-time compute can be more effective than merely increasing model size. For instance, employing a compute-optimal strategy—where computational resources are adaptively allocated based on the problem's complexity—can enhance efficiency by over four times compared to traditional methods. Moreover, in scenarios where smaller base models achieve moderate success rates, augmenting test-time compute enables them to outperform models up to 14 times larger.

While test-time compute offers significant benefits, it also presents challenges, related to -- among others -- resource allocation (determining the optimal amount of computational resources for each inference task requires sophisticated strategies to balance performance gains against computational costs), dynamic scaling (implementing adaptive compute strategies necessitates models capable of assessing problem difficulty in real-time and adjusting their computational efforts accordingly)~\cite{manvi2024adaptive}, and hardware implications (the shift towards increased test-time computation may influence hardware requirements, putting more pressure on delivering specialized inference-focused hardware solutions).

\noindent
\textbf{Test-Time Compute in the Context of the Blueprint. }
Our blueprint offers mechanisms to dynamically allocate computational resources during inference to improve performance, particularly for more complex problems. By leveraging the modular structure of the blueprint, TTC can be effectively implemented through specific operators designed for reasoning tasks. We now provide several examples.

\begin{itemize}[noitemsep, leftmargin=0.75em]
\item The \textbf{generate operator} can be used to implement TTC by dynamically increasing the number of next reasoning steps generated for harder problems. For simpler tasks, the operator may only generate a minimal set of continuations. However, for more complex problems, the operator can be used to create a larger set of potential reasoning steps, thereby expanding the search space.
\item The \textbf{refine operator} provides another avenue for implementing TTC by enhancing a given reasoning step multiple times for harder problems. In this approach, the operator iteratively improves the quality of a reasoning step, addressing ambiguities, rectifying errors, or improving clarity. For simpler tasks, the operator might only refine a step once, while for more complex reasoning, it can perform multiple enhancement iterations to ensure the output meets a higher standard of precision and robustness.
\item The \textbf{traversal operators}, such as select, enable the exploration of multiple reasoning paths at test time, offering another key mechanism for implementing TTC~\cite{zhang2024generative}. By using select on several next reasoning steps, the model can dynamically expand its search tree for more challenging problems, thereby increasing the diversity and depth of reasoning paths under consideration.
For example, in a complex task, the model might select multiple high-probability steps and explore their corresponding continuations in parallel. This approach facilitates broader exploration of the reasoning space, ensuring that promising paths are not prematurely discarded.
\item To efficiently manage the expanded set of possibilities, the blueprint allows integration with the \textbf{aggregate operator}. This operator evaluates the generated reasoning paths and selects the most promising ones based on predefined criteria, such as the likelihood of correctness or the quality of intermediate steps. This combination ensures that while more computational resources are allocated for challenging tasks, only the most relevant paths are explored further, optimizing both accuracy and efficiency.
\end{itemize}

\subsection{Models}

Models are used to implement various types of operators. Most common are the \textbf{value model} (implementing the value evaluation operator) and the \textbf{policy model} (implementing the generate operator).

Models are further categorized and discussed in detail in \iftr Appendix~\ref{sec:value-reward-models}\else Appendix~B\fi; we discuss the variants of the value model (\textbf{Q-Value Model}, \textbf{V-Value Model}), we compare Process-Based and Outcome-Based Reward Models, and we formally identify a new variant of models, the \textbf{Outcome-Driven Process-Based Reward Model}.

\subsubsection{Training Paradigm}

Each model must be trained according to a specified paradigm, which outlines the methodology for optimizing its performance. This paradigm defines key training components such as the loss function, data generation and labeling procedures, and other critical training details.

A wide range of training schemes has been developed for models used in RLMs, with early foundational work stemming from advancements related to AlphaZero. These schemes have since evolved to support the complex requirements of reasoning tasks within LLMs.
Common training paradigms include \textbf{supervised fine-tuning (SFT)}, where models are trained on reasoning sequences labeled with q-values; \textbf{rejection sampling}~\cite{stiennon2020learning,charniak2005coarse}, which involves filtering generated outputs based on quality criteria; and \textbf{RL-based methods} such as \textbf{Proximal Policy Optimization (PPO)}~\cite{schulman2017proximal}, \textbf{Direct Preference Optimization (DPO)}~\cite{rafailov2023direct}, and reasoning-specific variants like \textbf{Reasoning Policy Optimization (RPO)}~\cite{pang2024iterative}.
Several training paradigms also incorporate \textbf{self-learning}, where the model iteratively improves by generating and evaluating its own reasoning sequences, thereby simulating competitive or cooperative reasoning scenarios.

\subsubsection{Training Data Scope}

The training data for RLMs can vary significantly in terms of how much of the reasoning structure it captures.
We now outline two established approaches, \textbf{Outcome-Based Supervision (OBS)} and \textbf{Process-Based Supervision (PBS)}.
More details regarding both OBS and PBS can be found in \iftr Appendix~\ref{PBSvsOBS}\else Appendix~B.1\fi.

In \textbf{Outcome-Based Supervision} (also known as a \textbf{sparse training signal})~\cite{cobbe2021training, uesato2022solving} each training sample consists solely of the input and the corresponding output. For example, in mathematical problem-solving, a sample may include the task statement and the final solution, labeled as correct or incorrect. This approach is straightforward to implement, and the required data is relatively easy to collect. However, it can limit the model's reasoning accuracy, as it provides minimal insight into the intermediate steps that led to the solution~\cite{lightman2023let}.

An alternative approach is \textbf{Process-Based Supervision} (also known as a \textbf{dense training signal})~\cite{lightman2023let, wang2024math}, where a training sample reflects the entire reasoning structure. In this case, the sample contains not only the input and final output but also all intermediate reasoning steps, annotated with labels indicating the quality of each step. This richer training data allows the model to learn more granular reasoning patterns, improving its ability to generate accurate and interpretable solutions by understanding the reasoning process in detail. However, such data is much more challenging to generate or gather~\cite{lightman2023let}. 

\textbf{OBS vs.~PBS} By varying the training data scope, developers can strike a balance between ease of data collection and the depth of reasoning insights provided to the model, with dense supervision generally offering improved performance at the cost of increased data complexity. We detail these, and additional aspects of ORMs and PRMs in regards to pipelines for different training phases and paradigms in \iftr Appendix~\ref{sec:value-reward-models}, Appendix~\ref{app:phase1_training}, Appendix~\ref{sec:phase2-appendix}, and in Algorithms~\ref{algo:sft_policy_model}--\ref{algo:value_model_update}\else Appendix~B, Appendix~C.2, Appendix~C.3, and in Algorithms~2–7\fi.
%

\textbf{Trace-Based Supervision (TBS)} is a potential way to extend PBS by incorporating detailed information about the sequence of applied operators, including traversal operators, within the reasoning structure. By capturing the full trace of how reasoning steps are generated, refined, or revisited, TBS would provide richer supervision that teaches the model to internalize not just the reasoning steps but also the process of navigating and manipulating the reasoning structure itself. This approach could enable the training of more powerful Implicit RLMs by guiding them to replicate the reasoning dynamics of explicit structures, improving their ability to reason flexibly and efficiently.

\iftr

\begin{table*}[t]
\vspaceSQ{-0.5em}
\centering
\setlength{\tabcolsep}{1.5pt}
\ifsq\renewcommand{\arraystretch}{0.9}\fi
\footnotesize
\scriptsize
\ssmall
\sf
\begin{tabular}{lllllllllllcllllllll}
\toprule
& \multicolumn{3}{c}{\textbf{Reasoning}} & \multicolumn{10}{c}{\textbf{Reasoning Operator}} & \multicolumn{2}{c}{\textbf{Models}} & \multicolumn{3}{c}{\textbf{Pipeline}} & \\
\multirow{2}{*}{\textbf{Scheme}} & & & & \multicolumn{4}{c}{{\textbf{Structure}}} & \multicolumn{2}{c}{{\textbf{Traversal}}} & \multicolumn{2}{c}{\textbf{Update}} & \multicolumn{2}{c}{\textbf{Evaluation}} & & & & & & \multirow{2}{*}{\textbf{Remarks}} \\
\cmidrule(lr){2-4} \cmidrule(lr){5-8} \cmidrule(lr){9-10} \cmidrule(lr){11-12} \cmidrule(lr){13-14} \cmidrule(lr){15-16} \cmidrule(lr){17-19}
& \textbf{Structure} & \textbf{Step} & \textbf{Strategy} & \textbf{Gen.} & \textbf{Agg.} & \textbf{Pr.} & \textbf{Res.} & \textbf{Sel.} & \textbf{BT} & \textbf{Ref.} & \textbf{Bp.} & \textbf{Inter.} & \textbf{Term.} & \textbf{PM} & \textbf{VM} & \textbf{Inf.} & \textbf{Tr.} & \textbf{DG} & \\
\midrule
\multicolumn{20}{l}{\textbf{Explicit RLMs (Section~\ref{sec:rlms-existing-typical})}} \\
\midrule
rStar-Math~\cite{guan2025rstar} &
\textbf{E} Tree & \textbf{C} Thought + Code Block & \textbf{E} MCTS & 
\faY & \faN & \faN & \faN & \faY & \faN & \faN & \faY & \faY & \faY & 
\faY & \faY & 
\faY & \faY & \faY & 
\\ 
PRIME~\cite{yuan2024free, cui2025process} &
\textbf{E} Multiple Chains & \makecell[tl]{\textbf{F} Token \\ \textbf{C} Thought} & \textbf{E} Best-of-N & 
\faY & \faN & \faN & \faN & \faY & \faN & \faN & \faN & \faY & \faY & 
\faY & \faY & 
\faY & \faY & \faY & 
\\ 
Marco-o1~\cite{zhao2024marco} &
\textbf{E} Tree & \makecell[tl]{\textbf{F} Token Sequence \\ \textbf{C} Thought} & \textbf{E} MCTS & 
\faY & \faN & \faN & \faN & \faY & \faN & \faY & \faY & \faY & \faY & 
\faY & \faN & 
\faY & \faY & \faY & 
\\ 
%
%
Journey Learning (Tr.)~\cite{qin2024o1} &
\textbf{E} Tree & \textbf{E} Thought & \textbf{E} Tree Search & 
\faY & \faN & \faY & \faY & \faN & \faY & \faY & \faN & \faY & \faY & 
\faY & \faY & 
\faH* & \faY & \faY & 
*Separate entry \\ 
OpenR~\cite{wang2024openr} &
\textbf{E} Tree & \textbf{C} Thought & \makecell[tl]{\textbf{E} Best-of-N \\ \textbf{E} Beam Search \\ \textbf{E} MCTS}   & 
\faY & \faN & \faY & \faN & \faY & \faN & \faN & \faY & \faY & \faY & 
\faY & \faY & 
\faY & \faY & \faY & 
\\ 
LLaMA-Berry~\cite{zhang2024llamaberry} &
\textbf{E} Tree of Chains & \textbf{C} Solution &\textbf{E} MCTS &  
\faY & \faN & \faN & \faN & \faY & \faN & \faY & \faY & \faN & \faY & 
\faY & \faY & 
\faY & \faY & \faY & 
\\ 
ReST-MCTS*~\cite{zhang2024restmcts} &
\textbf{E} Tree & \textbf{C} Thought & \textbf{E} MCTS & 
\faY & \faN & \faN & \faN & \faY & \faN & \faH* & \faY & \faY & \faY & 
\faY & \faY & 
\faY & \faY & \faY & 
*Advice by critic\\ 
AlphaMath~Almost~Zero~\cite{chen2024alphamath} &
\textbf{E} Tree & \textbf{F} Thought & \textbf{E} MCTS & 
\faY & \faN & \faN & \faN & \faY & \faN & \faN & \faY & \faY & \faY & 
\faH* & \faH* & 
\faY & \faY & \faY & 
*Single model\\ 
MCTS-DPO~\cite{xie2024monte} &
\textbf{E} Tree & \textbf{F} Token Sequence & \textbf{E} MCTS & 
\faY & \faN & \faN & \faN & \faY & \faN & \faN & \faY & \faY & \faY & 
\faH* & \faH* & 
\faY & \faY & \faY & 
*Single model \\ 
AlphaLLM~\cite{tian2024toward} &
\textbf{E} Tree & \textbf{C} Option & \textbf{E} MCTS & 
\faY & \faN & \faY & \faN & \faY & \faN & \faN & \faY & \faY & \faY & 
\faY & \faY & 
\faY & \faY & \faY & 
\\ 
TS-LLM~\cite{feng2023alphazero} &
\textbf{E} Tree & \makecell[tl]{\textbf{F} Token \\ \textbf{F} Sentence} & \makecell[tl]{\textbf{E} MCTS \\ \textbf{E} Tree Search} & 
\faY & \faN & \faY & \faN & \faY & \faN & \faN & \faY & \faY & \faY & 
\faY & \faY & 
\faY & \faY & \faY & 
\\ 
\midrule
\multicolumn{20}{l}{\textbf{Implicit RLMs (Section~\ref{sec:rlms-implicit})}} \\
\midrule
DeepSeek-R1~\cite{guo2025deepseek} &
\textbf{I} Chain & \textbf{F} Token & \faN & 
\faY & \faN & \faN & \faN & \faN & \faH & \faN & \faN & \faN & \faN & 
\faN & \faN & 
\faY & \faN & \faN & 
\\ 
QwQ~\cite{qwen2024qwq} &
\textbf{I} Chain* & \textbf{F} Token & \faN & 
\faY & \faN & \faN & \faN & \faN & \faH & \faN & \faN & \faN & \faN & 
\faN & \faN & 
\faY & \faN & \faN & 
*Linearized Tree\\ 
%
%
Journey Learning (Inf.)~\cite{qin2024o1} &
\textbf{I} Chain* & \textbf{C} Thought & \textbf{I} DFS & 
\faY & \faN & \faN & \faN & \faN & \faH & \faN & \faN & \faN & \faN & 
\faN & \faN & 
\faY & \faN & \faN & 
*Linearized Tree \\ 
\midrule
\multicolumn{20}{l}{\textbf{Structured Prompting Schemes (Section~\ref{sec:rlms-prompting})}} \\
\midrule
Graph of Thoughts (GoT)~\cite{besta2024graph} &
\textbf{E} Graph* & \textbf{C} Thought & \textbf{E} Controller & 
\faY & \faY & \faN & \faN & \faY & \faY & \faY & \faN & \faY & \faY & 
\faN & \faH & 
\faY & \faN & \faN & 
*DAG\\ 
Tree of Thoughts (ToT)~\cite{yao2023tree} &
\textbf{E} Tree & \textbf{C} Thought & \textbf{E} DFS & 
\faY & \faN & \faY & \faN & \faY & \faY & \faN & \faN & \faY & \faY & 
\faN & \faH & 
\faY & \faN & \faN & 
\\ 
&
& & \textbf{E} Beam Search & 
& & & & & & & & & & 
& & 
& & & 
\\ 
Self-Consistency (SC)~\cite{wang2023self} &
\textbf{E} Multiple Chains & \textbf{C} Thought & \textbf{E} Majority Voting & 
\faY & \faN & \faN & \faN & \faY & \faN & \faN & \faN & \faN & \faN & 
\faN & \faN & 
\faY & \faN & \faN & 
\\ 
Chain of Thought (CoT)~\cite{wei2022chain} &
\textbf{I} Chain & \textbf{C} Thought & \faN & 
\faY & \faN & \faN & \faN & \faN & \faN & \faN & \faN & \faN & \faN & 
\faN & \faN & 
\faY & \faN & \faN & 
\\ 
\bottomrule
\end{tabular}
\caption{\textbf{Comparison of RLMs with respect to the provided taxonomy (Section~\ref{sec:essence-general} and Figure~\ref{fig:blueprint}).}
``\textbf{Reasoning}'': Details of the reasoning approach, specifically what is its \textbf{Structure} and its \textbf{Strategy}?
``\textbf{Reasoning Operator}'': Does a given scheme support operators on the reasoning structure? If yes, which classes (and specific functionalities) are supported \textbf{Structure} (``\textbf{Gen.}'': generate, ``\textbf{Agg.}'': aggregate, ``\textbf{Pr.}'': prune, ``\textbf{Res.}'': restructure), \textbf{Traversal} (``\textbf{Sel}'': select, ``\textbf{BT}'': backtrack), \textbf{Update} (``\textbf{Ref.}'': refine, ``\textbf{Bp.}'': backpropagate), and \textbf{Evaluation} of ``\textbf{Inter.}'': intermediate steps and ``\textbf{Term.}'': terminal steps?
``\textbf{Model}``: Does a given scheme use models to implement its operators and if so, which ones (``\textbf{PM}'': policy model, ``\textbf{VM}'': value model)?
``\textbf{Pipeline}'': Which pipelines are harnessed by a given scheme (``\textbf{Inf.}'': inference, \textbf{Tr.}'': training, ``\textbf{DG}'': data generation)? 
When describing representations, we use the following abbreviations:
``\textbf{E}'': explicit,
``\textbf{I}'': implicit.
``\textbf{F}'': fine-grained.
``\textbf{C}'': coarse-grained.
``\faY'': full support (i.e., YES),
``\faH'': partially [supported],
``\faN'': no support (i.e., NO).
}
\label{tab:schemes}
\end{table*}

\fi

\subsection{Pipelines}

A pipeline is a detailed specification of operations that orchestrates the details of the interaction between the reasoning scheme and the operators and models to achieve a specific objective. Typically, an RLM would incorporate a single \textbf{pipeline for inference} and a separate \textbf{pipeline for training} each model used in an RLM. Moreover, there could also be \textbf{pipelines for synthetic data generation} used for training models. One can also distinguish a pipeline that trains an Implicit RLM using the provided reasoning traces from the Explicit RLM.

The details of pipelines depend on arbitrary design choices. In Section~\ref{sec:essence-basic}, we provided a general description of how these pipelines work. In \iftr Appendix~\ref{sec:appendix-algo}\else Appendix~C\fi, we present detailed algorithmic specifications of our pipelines, along with insights into the reasoning behind these design choices.
\iftr
Specifically, the inference pipeline can be found in Appendix~\ref{sec:mcts_algo_description} and in Algorithm~\ref{alg:mcts_star}. Pipelines for different training phases and paradigms can be found in Appendix~\ref{app:phase1_training}, Appendix~\ref{sec:phase2-appendix}, and in Algorithms~\ref{algo:sft_policy_model}--\ref{algo:value_model_update}.
\else
Specifically, the inference pipeline can be found in Appendix~C.1 and in Algorithm~1. Pipelines for different training phases and paradigms can be found in Appendix~C.2, Appendix~C.3, and in Algorithms~2--7.
\fi
The data generation pipeline is detailed in \iftr Appendix~\ref{sec:appendix_data}\else Appendix~D\fi.

\ifcnf

\fi

\section{Expressing Existing Schemes}

We now showcase the expressivity of our blueprint, by illustrating how it can be used to model a broad scope of existing RLMs and other related works. We summarize the outcomes of the analysis in Table~\ref{tab:schemes}.
We start with typical and most prevalent Explicit RLM architectures based on MCTS and policy and/or value models, where a single reasoning step is an individual logical argument (Section~\ref{sec:rlms-existing-typical}).
We also discuss there schemes that generalize this typical design, by harnessing nesting or restructure operators.
%
Finally, we study Implicit RLMs (Section~\ref{sec:rlms-implicit}) and various structured prompting schemes such as CoT or ToT (Section~\ref{sec:rlms-prompting}), showing that they also fit our blueprint.


\subsection{Explicit RLMs}
\label{sec:rlms-existing-typical}

We start with the most widespread variant of RLMs that follows the architecture outlined in Section~\ref{sec:pipeline}.
These reasoning models such as TS-LLM \cite{feng2023alphazero}, AlphaLLM \cite{tian2024toward}, MCTS-DPO~\cite{xie2024monte}, and others \cite{chen2024alphamath, zhang2024restmcts, zhang2024llamaberry, wang2024openr, zhao2024marco, guan2025rstar} generally employ an explicit tree structure in which a node represents a distinct reasoning step. The reasoning strategy is based on the MCTS and focuses on iterative exploration, expansion and evaluation of nodes within the tree. By incorporating value mechanisms—such as prompt-based evaluation or dedicated value models, the system identifies and prioritizes promising branches, facilitating more informed decision-making and refinement of the reasoning process.
All MCTS based reasoning models implement at least a next-step generation operator, an evaluation operator, and the update operator for back-propagating the values. In addition, ReST-MCTS*, LLaMA-Berry, and Marco-o1 support a refinement operator to further improve produced reasoning steps.





\textbf{Journey Learning}~\cite{qin2024o1} exhibits two main differences to typical MCTS-based RLMs. First, it harnesses a linearization restructure operator, in which the tree reasoning structure is transformed into a chain, by extracting several selected reasoning chains from it and combining them together into an individual long chain. This way, the scheme attempts to harness insights from different tree branches. By maintaining a chain-based structure, Journey Learning preserves the simplicity of linear reasoning while embedding the capacity for self-correction and exploration of multiple hypotheses.
Additionally, Journey Learning introduces a pipeline for the internalization of such long reasoning chains into its weights. This enables the final model to generate such long reasoning chains, possibly containing different reasoning branches, directly from its weights, illustrating the path towards the construction of an implicit RLM.
%

\subsection{Implicit RLMs}
\label{sec:rlms-implicit}

\textbf{Qwens's QwQ}~\cite{qwen2024qwq} embodies a fully implicit reasoning model, characterized by an implicit reasoning structure that is generated autoregressively directly by the model weights.
The reasoning strategy in QwQ -- as indicated by the model output -- harnesses next-step generation, backtracking, summarization, and critique generation to derive the final solution. At each step, the model implicitly generates a new node within the chain by employing one of these four implicit operators, presumably implemented using special tokens.

\subsection{Structured Prompting Schemes}
\label{sec:rlms-prompting}

Finally, we also illustrate that advanced \textit{structured} prompting schemes, such as CoT, ToT, and GoT, constitute a fully explicit RLM structure without any implicit reasoning than what is originally presented in the used LLM, i.e., no models nor training or data generation pipelines.

\textbf{CoT}~\cite{wei2022chain} utilizes an implicit reasoning structure consisting of a chain of reasoning steps. The reasoning strategy employed in CoT is oriented towards constructing a single coherent chain of reasoning, culminating in a solitary solution, thus only needing the generate operator. CoT serves as the foundational framework for a range of advanced reasoning strategies, including prompting methodologies such as Self-Consistency and Self-Refinement, among others.

\textbf{Self-Consistency (SC)}~\cite{wang2023self} extends the CoT framework by introducing redundancy into the reasoning process. It generates multiple reasoning chains and employs a majority-voting mechanism to determine the most consistent solution, which implements a select operator from our blueprint.

\textbf{ToT}~\cite{yao2023tree} adopts an explicit reasoning structure organized in a hierarchical, tree-based format. Within this framework, each node corresponds to a distinct reasoning step, and branching facilitates exploration across multiple inferential pathways (the generate operator). Additionally, an evaluation operator, implemented via a specialized prompt and the LLM itself, assesses branches of the tree.

\textbf{GoT}~\cite{besta2024graph} introduces a more intricate reasoning structure by employing an explicit graph-based representation. In this framework, nodes represent individual reasoning steps, and the graph architecture supports non-linear, interdependent relationships between these steps. The reasoning strategy in GoT is orchestrated by an external controller, realized as a separate LLM, which guides the exploration, refinement and aggregation of the graph’s nodes.

\section{How to Use the Blueprint}

We now outline how to use our blueprint for the user's application; we keep this section in a tutorial style.

\subsection{Part 1: Define the Reasoning Scheme}

The first step in using the blueprint is to define the reasoning scheme, which specifies the foundational structure and strategy of your RLM. Start by selecting the reasoning structure. Chains are the most affordable in terms of token costs, at least when it comes to ICL~\cite{besta2024demystifying}. Trees, while the most expensive, offer rich branching that enhances exploratory reasoning. Graphs, though slightly cheaper than trees, introduce additional challenges in implementation but can yield significant accuracy gains due to their flexibility.

Next, decide on the granularity of reasoning steps. Coarse-grained steps, such as thoughts or sentences, are widely used due to their simplicity and ease of scaling. However, token-based granularity, which operates at the level of individual tokens, offers the potential for greater precision and unexplored accuracy improvements. This approach, while promising, demands significantly more computational resources and careful design. This decision defines your action space (possible operations) and state space (configuration of the reasoning structure).

Another decision is choosing a reasoning strategy to govern how the reasoning structure evolves. MCTS combined with some variants of policy and value models remains the most widely adopted approach due to its balance of exploration and exploitation. However, alternative strategies that have not been deeply studied, such as ensembles of reasoning structures, may offer untapped potential.

Finally, determine the specific details of your chosen strategy, including parameters like exploration coefficients, decoding strategy, scoring functions, and step evaluation methods. These choices will significantly impact the model's reasoning dynamics, scalability, and overall effectiveness. Each decision at this stage lays the foundation for tailoring the RLM to your specific application requirements.






\subsection{Part 2: Define the Operators}

The next step is to specify the set of operators that will govern the reasoning process. For an MCTS-based design, the simplest approach is to implement the core operators: generate (often called expand for MCTS), select, and backpropagate. These fundamental operations suffice for many scenarios, providing a straightforward framework for reasoning.

Beyond the basics, consider whether you want to incorporate less mainstream operators, such as backtrack. By explicitly including backtrack, you enable a clearer tracking of progress within the search tree, making it potentially easier to revisit and refine earlier reasoning steps. This approach also facilitates advanced training schemes, like Trace-Based Supervision, by generating richer and more structured data. Consider using this and other operators within our toolbox.

You will also need to determine the implementation details for each operator. Decide which operators will be implemented as neural models—such as using a policy model to guide selection or a value model for backpropagation—and which will rely on non-neural methods. This choice affects both the computational complexity and the flexibility of the system, so it's important to align these decisions with your reasoning scheme and performance goals.





\subsection{Part 3: Determine the Training Details}

In this phase, you need to outline the specifics of training for the models that will implement operators. For an MCTS-based design, consider the typical approach of using the policy model to implement generate (expand) and the value model for evaluation. If necessary, you might also train a separate model to calculate the reward at individual nodes, enhancing the precision of the reward signals.

Identify the application or training domain in order to address generalization requirements. This step ensures that your models are trained on data representative of the tasks you want them to handle.

Define the models, including their architectures and the selection of suitable base models. Consider how the design of these models—such as transformer-based architectures or more specialized designs—aligns with your reasoning structure and overall objectives.

Collect training data for both the policy and value models. For the policy model, consider generating data automatically with our pipeline or using a scheme such as CoT prompting, and include a special end-of-step token to ensure clean segmentation. For the value model, generate data through full MCTS simulations, which provide rich, structured information about reasoning paths and outcomes.

Fine-tune the models as needed. If using coarse reasoning steps, perform supervised fine-tuning (SFT) on the policy model to teach it how to reason step-by-step. Similarly, apply SFT to the value model to initialize it as a reliable evaluator.

Run MCTS with initialized models to collect additional data. You might filter this data to keep only high-quality reasoning paths (terminal states) or strong signals (high absolute advantages) for further training.

Finally, train both models either by additional SFT rounds or with reinforcement learning methods such as Proximal Policy Optimization (PPO). This ensures that the models are optimized not only for accuracy but also for the efficiency and robustness needed in complex reasoning tasks.

\section{Framework \schemename: Design \& Implementation}
\label{subsec:framework-design}

We now introduce \schemename\footnote{https://github.com/spcl/x1}, an extensible and minimalist framework that can serve as ground to design and experiment with RLMs, and currently provides one example of the blueprint.\footnote{We are working continuously on expanding the framework as well as adding more RLMs.}
An overview of the framework is in Figure~\ref{fig:framework-server-infra}.

\subsection{Reasoning Scheme}

The \schemenameS framework employs a tree reasoning structure in conjunction with MCTS as the reasoning strategy. This combination allows for a systematic exploration of reasoning paths while balancing exploration of new possibilities and exploitation of promising solutions judged by a value model. The framework achieves this alignment through the implementation of a series of operators that guide the construction, traversal, evaluation, and updating of the reasoning tree.

\subsection{Operators}

The \textbf{generate} operator plays a crucial role in expanding the tree by adding new children to a selected node. To improve the diversity of these newly generated nodes, we employ Diverse Beam Search~\cite{vijayakumar2018diverse}, which ensures variability among the children. Alternatively, high-temperature sampling can be used to introduce stochasticity into the generation process, fostering the exploration of different reasoning paths.

Traversal of the reasoning tree is managed by the \textbf{select} operator, which uses the PUCT function to identify the next node to expand. This operator balances a trade-off between exploration, favoring less-visited nodes, and exploitation, reinforcing nodes with higher potential based on previous evaluations. Always starting from the root node, the traversal mechanism ensures that the system can dynamically explore alternative paths and recover from suboptimal decisions by backtracking and selecting new branches.

The \textbf{backpropagation} update operator refines the q-values which can be used as guidance for the select operator along the path from an expanded node back to the root. This process incorporates new information from downstream nodes, leading to progressively more accurate q-values for the intermediate nodes. These refined q-values subsequently inform future decisions, making the reasoning process increasingly robust over time.

The framework implements two different evaluate operators. First, the \textbf{value estimate} operator predicts the discounted expected future reward for a chain extending from the root to a specific node. This prediction is derived from the q-value model, offering a quantitative measure of the path's quality. Second, when the ground truth is available, the \textbf{terminality evaluation} operator directly evaluates leaf nodes for correctness, assigning fixed rewards to verified solutions. These rewards are incorporated into the q-values of upstream nodes, ensuring that the reasoning process is informed by both model predictions and objective validation.

\begin{figure*}[t]
    \centering
    \includegraphics[width=1.0\linewidth]{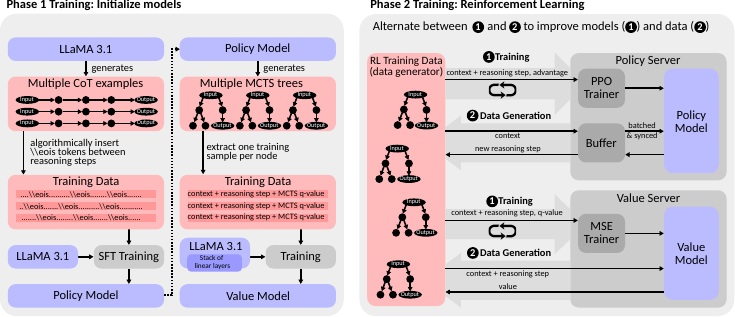}
    \caption{An overview of the \schemenameS framework is presented, highlighting its two-phase training process. In phase 1, the models are initialized, while in phase 2, the models are iteratively refined by alternating between constructing a sufficient number of MCTS trees and training the models on data derived from these trees.}
    \label{fig:framework-server-infra}
\end{figure*}

\subsection{Models \& Training Paradigms}

Both the value and the policy model in \schemenameS are fine-tuned versions of an LLM\footnote{We currently use Llama-3.1-8B-Instruct as base model.}, without reliance on prompting, which is used in several other RLM architectures~\cite{guan2025rstar,zhang2024restmcts}. This design decision aims to maximize the quality of results.
We now outline briefly selected key aspects of how we train these models, full details can be found in \iftr Appendices~\ref{sec:value-reward-models}, \ref{sec:appendix-algo}, and \ref{sec:appendix_data}\else Appendices~B, C, and D\fi.

\subsubsection{Training the Policy Model}

The policy model also leverages an LLM to generate new nodes during the MCTS. It is fine-tuned to output an individual next reasoning step instead of a whole chain of thoughts towards a completion (which LLMs commonly do). We achieve this by introducing \textit{a novel token}, \textit{the end of intermediate step (\textbf{eois}) token}, which denotes the completion of each reasoning step. The eois token complements the standard end of sequence (eos) token, which indicates the conclusion of an entire reasoning chain.
By incorporating the eois token, the framework enables the explicit identification of intermediate reasoning steps, allowing for greater interpretability and precise determination of whether the reasoning process is complete or ongoing. This dual-token strategy enhances the LLM’s capability to decompose complex problems into manageable substeps while ensuring the model recognizes when a solution has been reached.

\subsubsection{Training the Value Model}

The value model is designed to estimate the sum of the expected discounted future rewards for a sequence of reasoning steps and a newly proposed reasoning step, quantifying the value of the node modeling this step. For a given node in the MCTS tree, its value (referred to in the MCTS literature as state action value or q-value) is defined as the expected cumulative reward discounted by the number of steps required to achieve it. Formally, the q-value $Q_{\pi}(s_t, a_t)$ for traversing the edge to node $s_{t+1}$ when taking action $a_t$ from $s_t$ at depth $t$ in the MCTS tree is expressed as

\begin{align}
    Q_{\pi}(s_t, a_t) &= \mathbb{E}\left[\gamma^{T-t} r(s_T, a_T) \mid s_{t}, a_{t}\right] \nonumber \\
    &\approx \frac{1}{N} \sum_{i=1}^N \gamma^{T-t} r(s_{T}^{(i)}, a_{T}^{(i)}),  \label{eq:emp}
\end{align}

\noindent
where $\gamma$ is the discount factor, $T$ marks the last reasoning step $a_T$ that is added, resulting in the terminal state $s_{T+1}$ containing the complete reasoning structure and rewards are modeled sparse.
The terminal state $s_{T+1}$ is defined as the state in which no additional reasoning steps can be added. It typically represents the state containing the final solution to the problem at hand. Accordingly, $r(s_{T}, a_{T})$ is the terminal reward. We chose to model rewards as sparse, where only the final reasoning step receives a non-zero reward, since for most reasoning tasks, only the final answer can be evaluated against the true solution. As a result, one can only obtain a reward signal when the last step is reached. We can approximate the q-value by sampling $N$ reasoning chains until the terminal state, as in Eq.~\ref{eq:emp}, and averaging the terminal rewards discounted by the depth required.


The q-value model is trained using data from completed MCTS searches. Initially, when the q-value model is unavailable, $N$ simulations (complete rollouts) are performed, and the average discounted reward is used to initialize the q-values for each node. More information can be found in the \iftr Appendix~\ref{value_model_phase1_data}\else Appendix~D.2\fi.

\subsection{Enabling Scalability and Efficiency}

The current implementation is built to scale to multiple GPUs on multiple nodes. To further enhance the scalability and computational efficiency, several architectural and operational improvements have been implemented. 
%

One design decision involves the decoupling of the value and policy models. The deployment of dedicated value and policy servers confers several advantages:

\begin{itemize}[noitemsep, leftmargin=0.75em]
    \item \textbf{Scalability} The decoupling of value and policy servers from the MCTS instance facilitates scalability and the execution of multiple parallel MCTS instances.
    \item \textbf{Batch Processing} The policy server incorporates batching capabilities, allowing the concurrent processing of multiple queries, thereby enhancing throughput.
    \item \textbf{Resource Optimization} The independent allocation of computational resources to the value and policy models is inherently supported by the framework's architecture, enhancing efficient resource utilization.
    \item \textbf{Replication and Distribution} The separation of value and policy models facilitates the application of distinct replication and distribution strategies.
\end{itemize}

Figure~\ref{fig:framework-server-infra} illustrates the implementation of the framework as a server architecture, demonstrating how these structural enhancements contribute to improved scalability and efficiency. Building on these architectural enhancements, we employ the following strategies to further optimize the framework's efficiency and scalability, focusing on inference and parallelization.

In the framework, we incorporate the standard optimizations of batching, quantization, and KV caching.
Inference calls are batched in the policy model, enabling simultaneous processing of multiple queries.
To expedite the reasoning process, the framework creates multiple child nodes in parallel during the node expansion phase. Specifically, \(N\) new nodes are generated concurrently in each expansion step, reducing computational overhead and enhancing overall system performance.
Further optimization of inference speed is achieved through KV caching and quantization. KV caching mechanisms mitigate redundant computations, while quantization techniques reduce the memory consumption of both policy and value models.

\subsection{Blueprint for Efficient Scaling}

Our blueprint can be deployed to AI HPC systems and clouds, as both systems provide the performance and resources necessary to scale RLMs.
Deployment on HPC systems is straightforward: compute tasks are distributed across statically allocated nodes, connected with a low-latency and high-bandwidth interconnect, and with training data being available on a high-performance parallel filesystem.
On the other hand, the cloud provides many configurable services that offer different trade-offs between performance, cost, and reliability.
There, it becomes the user's responsibility to choose the storage options and compute granularity that provides the best match for
expected performance and cost.
The architecture of our blueprint fits into the \textit{microservice} architecture, with a clear separation of {compute tasks}, {data storage}, and {coordination}.
This architecture helps to ease the configuration process, as different components of the system can be deployed, scaled, and optimized independently.
In particular, the separation of value and policy servers allows them to be scaled separately according to the complexity of reasoning steps that might require different resource allocations to handle task generation and evaluation.

First, we outline the major decisions users must make before deploying the \schemenameS scaling blueprint:

\begin{itemize}[noitemsep, leftmargin=0.75em]
    \item \textbf{Deployment} Training and inference tasks are typically allocated to virtual machines and containers, with the latter typically deployed as managed services with an orchestrator such as Kubernetes. There, \schemenameS can benefit from modern frameworks like Ray~\cite{moritz2018ray} that hide the complexity of managing a service in a Kubernetes cluster.
    \item \textbf{Data Storage} In the cloud, object storage provides automatic bandwidth scalability that allows to scale computations operating on the same data. To overcome latency and power constraints, data can also be placed in in-memory caches like Redis and hybrid solutions that combine disks with flash memory~\cite{zhao2023tectonic}.
    \item \textbf{Communication} Requirements of the \schemenameS blueprint differ from classical microservices, that rely on high-level abstractions like RPC and REST interfaces. RLM must utilize high-performance network fabrics offered by modern clouds, such as InfiniBand on Azure and Elastic Fabric Adapter (FBA) on AWS, both capable of achieving throughput of 400 Gb/s~\cite{desensi2022noise}. These are also available to training processes distributed across many GPUs, e.g., through specializations of the NVIDIA collectives library NCCL.
    \item \textbf{Parallelism} We apply parallelism at multiple blueprint levels, including the classic data, model, and pipeline parallelism. These can scale horizontally across a larger number of virtual machines and containers. On the other hand, reasoning steps can benefit from elastic scaling, like in distributed MCTS and Beam Search, where each path can be explored in parallel. There, containers can be allocated on the fly to support new paths and deallocated as soon as the parallelism scale of the computation decreases.
\end{itemize}

New developments in the machine learning infrastructure can significantly impact RLM deployment strategies:

\begin{itemize}[noitemsep, leftmargin=0.75em]
    \item \textbf{Elastic Compute} Computing tasks can be executed on ephemeral resources that trade the guaranteed lifetime
    and reliability for lower costs, such as spot virtual machines~\cite{miao2024spotserve}. Serverless functions provide elasticity scalability with fine-grained pricing models~\cite{copik2021sebs}, which can be a good fit for dynamically generated reasoning steps. However, serverless functions are stateless and suffer from cold starts, which requires optimization techniques dedicated to LLMs~\cite{fu2024serverlessllm}. Furthermore, restricted network communication in functions forces the adoption of new communication protocols~\cite{jiang2021lambdaml,copik2022fmi}.
    
    \item \textbf{GPU Management} 
    Cloud rental of GPU devices is particularly expensive, and procuring a sufficient number of devices can be challenging, specifically when constrained to a single cloud region. Given the large compute and memory requirements of base models, space-sharing might not be feasible. On the other hand, time-sharing of GPU devices between different \schemenameS services could be a viable alternative, but it is currently constrained by large memory allocations and the cost of swapping model checkpoints between CPU and GPU memory. To increase resource utilization, new techniques for efficient GPU checkpoint and restore are needed~\cite{fu2024serverlessllm}.
    
    \item \textbf{Parameter-Efficient Resource Sharing} Resource-sharing can be further enhanced by utilizing a shared base model architecture for the policy and value models, while dynamically swapping task-specific parameter layers - such as Low-Rank Adaptation~\cite{hu2022lora}, prefix tuning~\cite{li2021prefix}, or other adapter layers - on the GPU during inference. These modular strategies keep the base model loaded in device memory and replace only the lightweight task-specific layers, eliminating redundant loading and reducing both latency and memory usage. An example of an RLM, which uses a shared base model with separate additional linear layers for policy and value model, is AlphaMath~\cite{chen2024alphamath}.

    \item \textbf{Cross-Region Deployment} Cloud applications are often deployed in a single region to avoid the performance and cost of cross-region data access. However, workloads can be scheduled globally, suspended, and migrated across regions to avoid hardware resource exhaustion and achieve lower carbon emissions~\cite{choudhury2024mast,wiesner2021let}.
\end{itemize}

\iftr
\subsection{Example Analysis: Token Probability Distributions}
\label{sec:token_analysis}

As an illustrative example, we use the framework to directly leverage the \textit{token probability distribution}, thereby facilitating the use of associated properties—such as entropy and variance—for guiding subsequent reasoning decisions. By focusing on these probabilistic characteristics, the framework can help identify when to expand a given reasoning step. Using token probability distributions can be used for navigating the reasoning based on both coarse and fine steps. To support this analysis, the \schemenameS implementation includes scripts that provide insights into token-level metrics, such as entropy fluctuations and distribution patterns, to inform reasoning strategies.



\subsubsection{Relevance of Token Probability Distribution}

The token probability distribution provides critical information about the likelihood of different next-step candidates in a reasoning process. By examining this distribution, we can gain insight into how certain tokens dominate or diversify the reasoning space, and in turn, guide more informed decisions about which step to take next. 


We now list a few scenarios where different token distributions offer insights into which reasoning decision is best to take at a given step.

\begin{itemize}[noitemsep, leftmargin=0.75em]
\item \textbf{Flat Token Distribution. }
A flat probability distribution occurs when all tokens have roughly equal probabilities. In this scenario, there is significant uncertainty about which step is the best to choose because no single token stands out as a clear candidate. This can make the reasoning process more exploratory, as the model may need to consider multiple tokens equally and rely on additional strategies—such as external heuristics or learned policies—to identify the most promising step. While this can foster exploration, it may also lead to inefficiencies since the model might need to evaluate many equally plausible paths before finding an optimal solution. Another decision that could be taken in such a scenario, is to delay initiating a reasoning step till the token distribution changes to be more skewed.
\item \textbf{Skewed Distribution with One Dominant Token. }
When one token has a much higher probability than others, the distribution is highly skewed. This often signals that the model is confident about the next step in the reasoning process. If the dominant token corresponds to a logical or well-supported continuation, this confidence can streamline decision-making and reduce computational overhead. However, if the model's confidence is misplaced—perhaps due to biases in the training data or a lack of context—relying on a single dominant token may cause the reasoning process to follow a suboptimal path. In such cases, it's crucial to assess whether the high-probability token genuinely represents the most logical next step or if additional validation is needed.
\item \textbf{Skewed Distribution with Multiple High-Probability Tokens. }
In some cases, the distribution may be skewed with a small set of tokens receiving much higher probabilities than others. This indicates that the model sees several plausible continuations, each with a reasonable chance of being correct. While this is generally a positive sign—offering a diversity of credible options—it also complicates the decision-making process. The reasoning strategy must weigh the trade-offs between these top candidates, considering not only their individual probabilities but also how each choice impacts the subsequent reasoning trajectory. This scenario highlights the need for effective evaluation metrics (like entropy or Gini coefficient) to help select the step that contributes most to reaching the correct or desired outcome.
\end{itemize}


By analyzing token probability distribution and identifying the cases above and others, reasoning strategies can, for example, improve efficiency (identifying when a distribution is flat allows the reasoning algorithm to focus on  diversification or introduce additional constraints to narrow down choices), enhance decision confidence (recognizing when one token is dominant can help expedite decisions, provided the model's confidence is well-founded), or foster balanced exploration (detecting multiple high-probability tokens facilitates exploring various credible paths without being overly committed to a single option).


\begin{figure*}[t]
  \begin{subfigure}{0.49\textwidth}
      \vspace{0.5em}
      \centering
      \includegraphics[width=1.0\columnwidth]{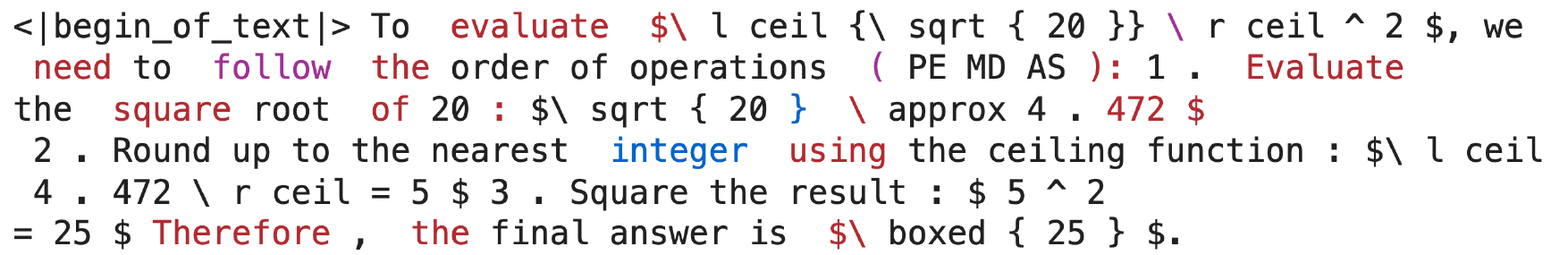}
      \vspace{0.2em}
      \subcaption{1st example}
        \vspace{1em}
      \label{fig:model_output:0}
  \end{subfigure}
  \begin{subfigure}{0.49\textwidth}
      \centering
      \includegraphics[width=1.0\columnwidth]{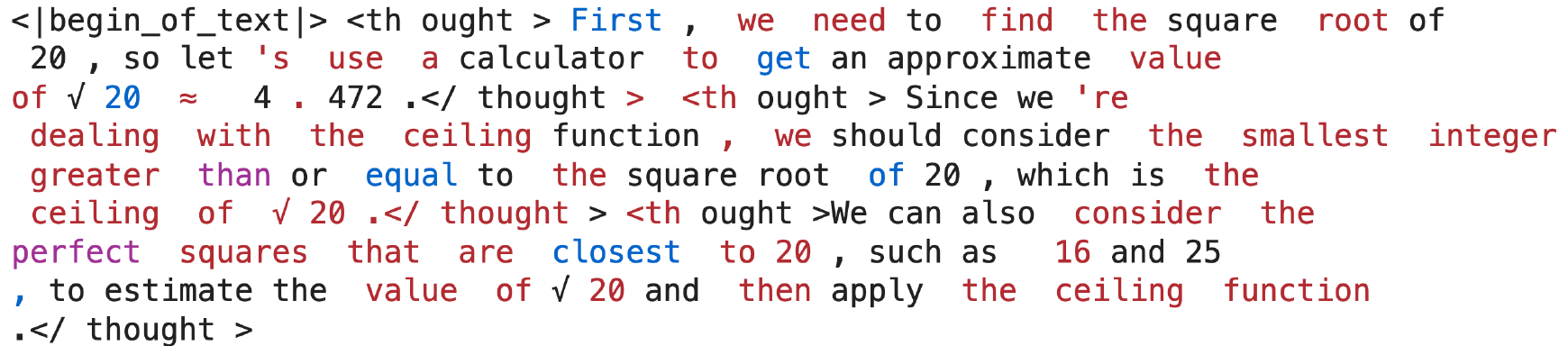}
      \subcaption{2nd example}
       \vspace{1em}
      \label{fig:model_output:1}
  \end{subfigure}
  \begin{subfigure}{0.49\textwidth}
      \centering
      \includegraphics[width=1.0\columnwidth]{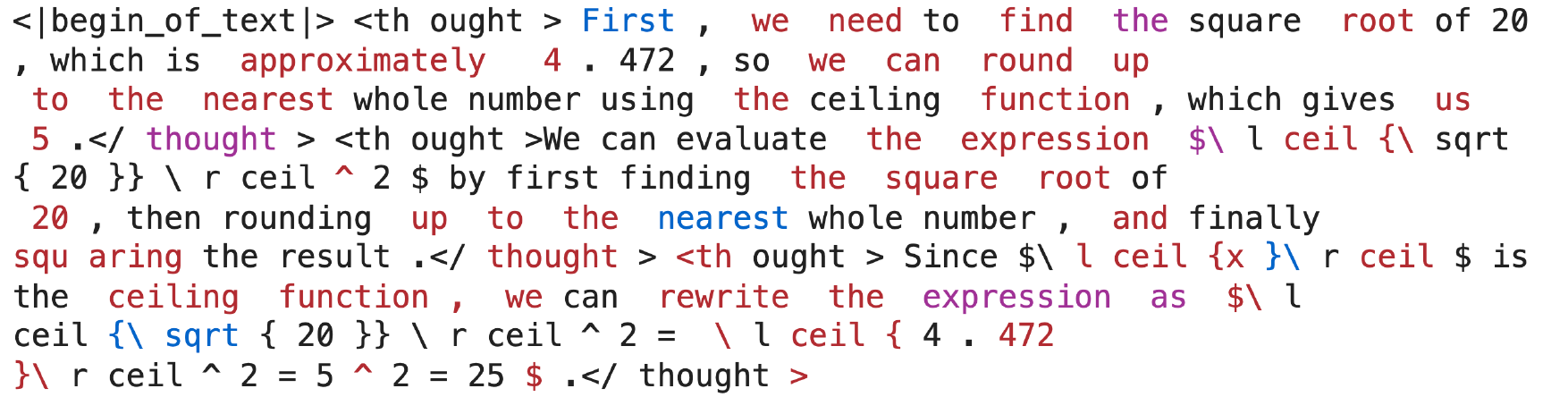}
      \subcaption{3rd example}
      \label{fig:model_output:2}
  \end{subfigure}
  \begin{subfigure}{0.49\textwidth}
      \vspace{0.3em}
      \centering
      \includegraphics[width=1.0\columnwidth]{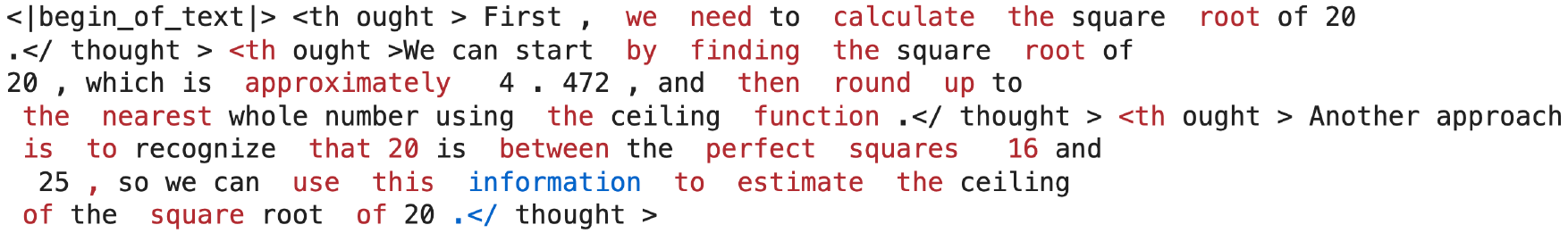}
      \vspace{0.3em}
      \subcaption{4th example}
      \label{fig:model_output:3}
  \end{subfigure}
  \caption{\textbf{Four examples of model output with highlighted tokens indicating uncertainty levels.}
  The outputs have been color-coded to reflect the confidence levels of the model's token predictions. Tokens are highlighted \textbf{\textcolor{purple}{in purple}} when the highest probability is below 0.8 (indicating lower certainty without significant contention), \textbf{\textcolor{blue}{in blue}} when the second-highest probability exceeds 0.1 (indicating contention, where another token is a close alternative), and \textbf{\textcolor{red}{in red}} when both conditions are met (indicating high uncertainty). These examples illustrate varying levels of prediction confidence and contention in reasoning steps, emphasizing regions of high ambiguity or competition between plausible continuations. This type of visual analysis is useful for identifying points in the reasoning process where the model lacks confidence or is torn between alternatives, guiding refinements in reasoning strategies and model design. It also helps pinpoint critical areas where additional supervision or context may improve model performance.}
  \label{fig:model_output}
\end{figure*}

\subsubsection{Analyzing Token Probability Distribution}

To understand the form of a token probability distribution, we examine variance, entropy, VarEntropy, and the Gini coefficient as key metrics that offer distinct perspectives on the distribution's shape and characteristics.

\textbf{Variance} provides a broad measure of uncertainty by reflecting how spread out the probabilities are across the vocabulary. When variance is low, the probabilities are nearly uniform, indicating a flat distribution. However, variance alone does not capture the specific structure or shape of the distribution. For example, two distributions can have the same variance but differ in their overall form, such as one having multiple minor peaks versus another being nearly uniform with a single dominant token. To address this, we consider further measures below.

\textbf{Entropy} has long been a standard measure of uncertainty and information content in a probability distribution. Higher entropy corresponds to greater unpredictability—requiring more information to describe the system's state. For instance, if all tokens have nearly equal probabilities, the entropy is high, reflecting a flat distribution. In contrast, low entropy occurs when a small number of tokens dominate, resulting in a skewed distribution. The entropy of a distribution is given by $H = - \sum_i p_i \log_2(p_i),$ where $p_i$ is the probability of the $i$-th token. This metric provides valuable insight into whether the distribution is diffuse and exploratory or concentrated and decisive.

\textbf{VarEntropy} extends this analysis by measuring the variability of entropy itself, thus offering a dynamic view of how uncertainty changes. A high VarEntropy combined with low entropy often indicates a sharp, focused distribution with a few dominant outcomes. Conversely, low VarEntropy and high entropy typically reflect a flat, uniform distribution where no single token stands out. The VarEntropy is defined as $\sum_i p_i(|\log(p_i)|-|H|)^2$. This metric captures the nuanced shifts in distribution shape, helping to pinpoint how tightly probabilities cluster around certain tokens versus how broadly they spread.

The \textbf{Gini Coefficient}, traditionally used to measure inequality, provides another lens on the form of the distribution. A perfectly equal distribution has a Gini coefficient of 0, signifying that all tokens have identical probabilities. A Gini coefficient closer to 1 indicates high inequality, where a few tokens hold most of the probability mass. By visualizing the cumulative distribution of sorted probabilities, the Gini coefficient highlights how the probability is concentrated or dispersed.

Together, these metrics—variance, entropy, VarEntropy, and Gini—enable a detailed examination of token probability distributions. By leveraging each metric’s unique strengths, we can effectively characterize whether a distribution is flat, skewed with a dominant token, or skewed across several highly probable tokens, ultimately guiding more informed decisions in reasoning and model development.

\subsubsection{Example Results}

Figure~\ref{fig:model_output} and~\ref{fig:token_uncertainty} illustrate example model outputs and their respective token probability distributions. By analyzing the highest probabilities, the second-highest probabilities, and the sum of the remaining probabilities, we gain valuable insights into the underlying token distribution, which can subsequently be quantified through the uncertainty metrics discussed earlier.


\begin{figure*}[hbtp]
  \begin{subfigure}{\textwidth}
      \centering
      \includegraphics[width=1.0\textwidth]{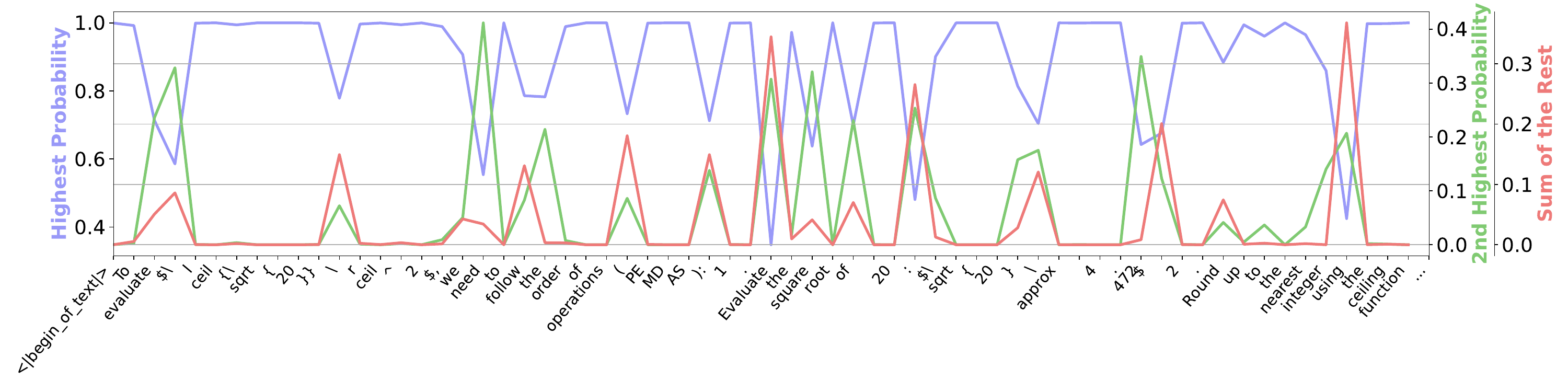}
      \subcaption{\centering To evaluate $\lceil{\sqrt{20}}\rceil^2$, we need to follow the order of operations (PEMDAS):1. Evaluate the square root of 20: $\sqrt{20} \approx 4.472$ \break 2. Round up to the nearest integer using the ceiling function: $\lceil 4.472 \rceil = 5$ \break 3. Square the result: $5^2 = 25$Therefore, the final answer is $\boxed{25}$.}
      \label{fig:token_uncertainty:0}
  \end{subfigure}
  \begin{subfigure}{\textwidth}
      \centering
      \includegraphics[width=1.0\textwidth]{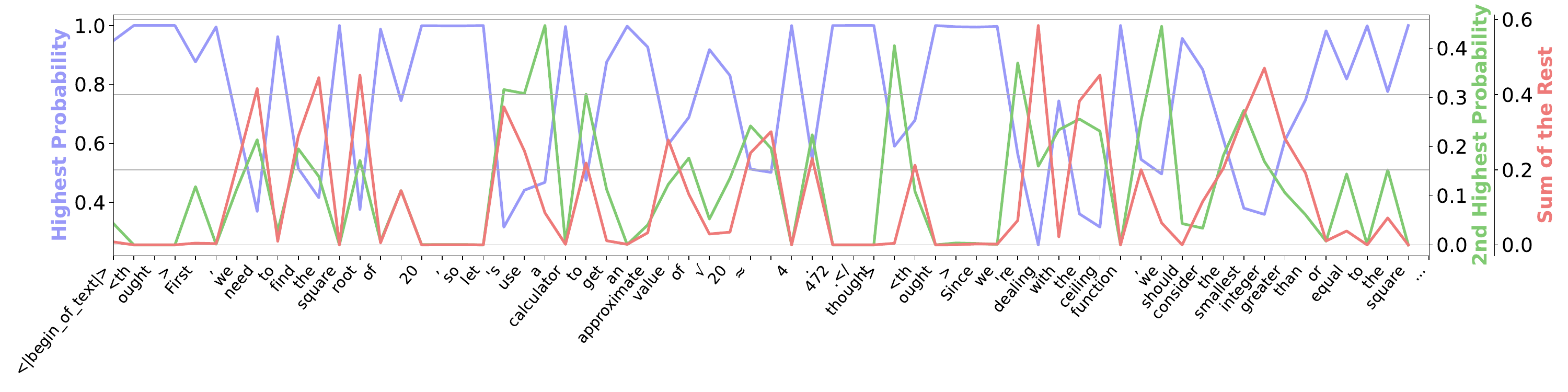}
      \subcaption{\centering \textless thought\textgreater First, we need to find the square root of 20, so let's use a calculator to get an approximate value of $\sqrt{20} \approx 4.472$.\textless thought\textgreater \break \textless thought\textgreater Since we're dealing with the ceiling function, we should consider the smallest integer greater than or equal to the square root of 20, which is the ceiling of $\sqrt{20}$. \textless /thought\textgreater \break \textless thought\textgreater We can also consider the perfect squares that are closest to 20, such as 16 and 25, to estimate the value of $\sqrt{20}$ and then apply the ceiling function.\textless /thought\textgreater}
      \label{fig:token_uncertainty:1}
  \end{subfigure}
  \begin{subfigure}{\textwidth}
      \centering
      \includegraphics[width=1.0\textwidth]{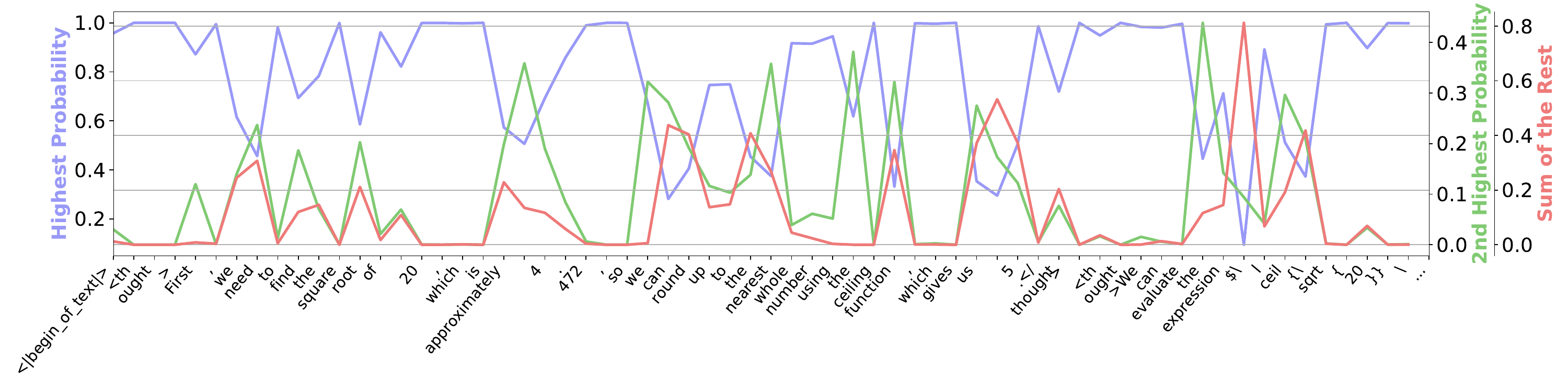}
      \subcaption{\centering \textless thought\textgreater First, we need to find the square root of 20, which is approximately 4.472, so we can round up to the nearest whole number using the ceiling function, which gives us 5.\textless /thought\textgreater \break \textless thought\textgreater We can evaluate the expression $\lceil{\sqrt{20}}\rceil^2$ by first finding the square root of 20, then rounding up to the nearest whole number, and finally squaring the result.\textless /thought\textgreater \break \textless thought\textgreater Since $\lceil{x}\rceil$ is the ceiling function, we can rewrite the expression as $\lceil{\sqrt{20}}\rceil^2 = \lceil{4.472}\rceil^2 = 5^2 = 25$.\textless /thought\textgreater}
      \label{fig:token_uncertainty:2}
  \end{subfigure}
  \begin{subfigure}{\textwidth}
      \centering
      \includegraphics[width=1.0\textwidth]{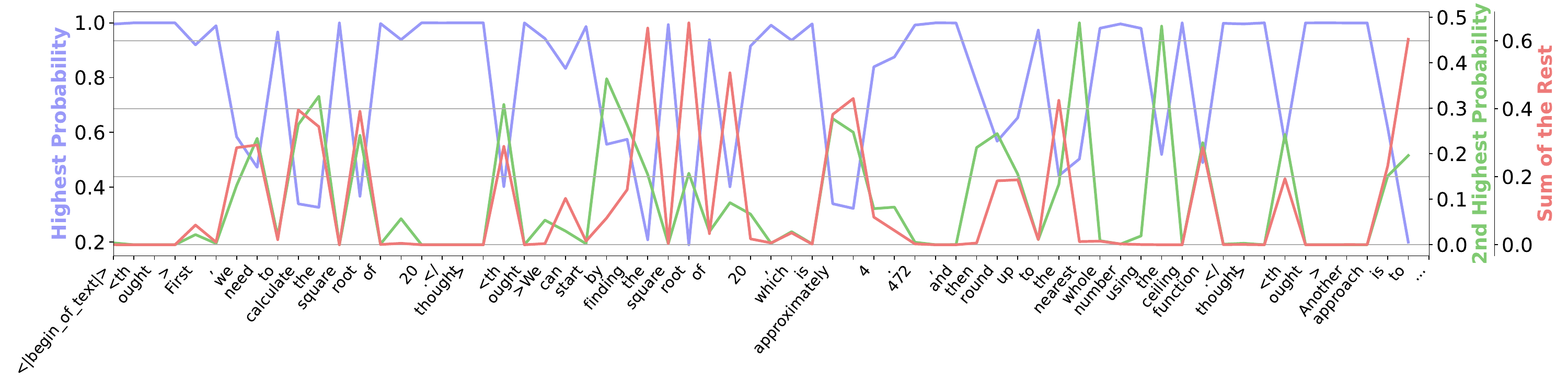}
      \subcaption{\centering \textless thought\textgreater First, we need to calculate the square root of 20.\textless /thought\textgreater \break \textless thought\textgreater We can start by finding the square root of 20, which is approximately 4.472, and then round up to the nearest whole number using the ceiling function.\textless /thought\textgreater \break \textless thought\textgreater Another approach is to recognize that 20 is between the perfect squares 16 and 25, so we can use this information to estimate the ceiling of the square root of 20.\textless /thought\textgreater}
      \label{fig:token_uncertainty:3}
  \end{subfigure}
  \caption{\textbf{Probabilities of the first 64 tokens of example model outputs.} We show the
  two highest probabilities as well as the sum of the other probabilities.}
  \label{fig:token_uncertainty}
\end{figure*}

In Figures~\ref{fig:token_uncertainty:0} and~\ref{fig:token_uncertainty:3}, specific regions emerge where the top two probabilities are very close, while the remaining probabilities are significantly smaller. Such regions likely indicate scenarios where forking the reasoning process (e.g., exploring multiple paths) could disproportionately benefit future outcomes, as the competing high-probability tokens suggest alternative plausible continuations. Conversely, in instances where the first probability is notably high, with much lower second and remaining probabilities, the model exhibits strong confidence in a single continuation. These cases are conducive to more deterministic reasoning, as forking may be unnecessary.

Additionally, regions with a relatively high sum of the remaining probabilities (close to the top two) highlight flatter distributions with high uncertainty. These scenarios signal a need for cautious reasoning, where clarification or additional contextual refinement may help reduce ambiguity. For instance, such uncertainty may suggest that the model has not yet committed to a specific path and could benefit from revisiting earlier reasoning steps to address potential errors or misalignments.


\begin{figure*}[hbtp]
  \begin{subfigure}{\textwidth}
      \centering
      \includegraphics[width=1.0\textwidth]{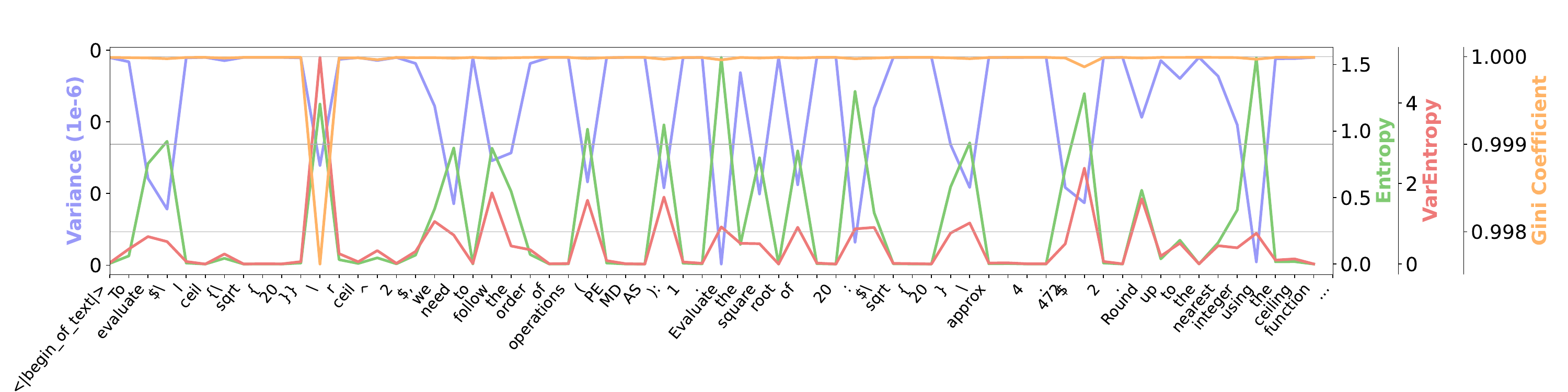}
      \subcaption{\centering To evaluate $\lceil{\sqrt{20}}\rceil^2$, we need to follow the order of operations (PEMDAS):1. Evaluate the square root of 20: $\sqrt{20} \approx 4.472$ \break 2. Round up to the nearest integer using the ceiling function: $\lceil 4.472 \rceil = 5$ \break 3. Square the result: $5^2 = 25$Therefore, the final answer is $\boxed{25}$.}
      \label{fig:token_uncertainty_metrics:0}
  \end{subfigure}
  \begin{subfigure}{\textwidth}
      \centering
      \includegraphics[width=1.0\textwidth]{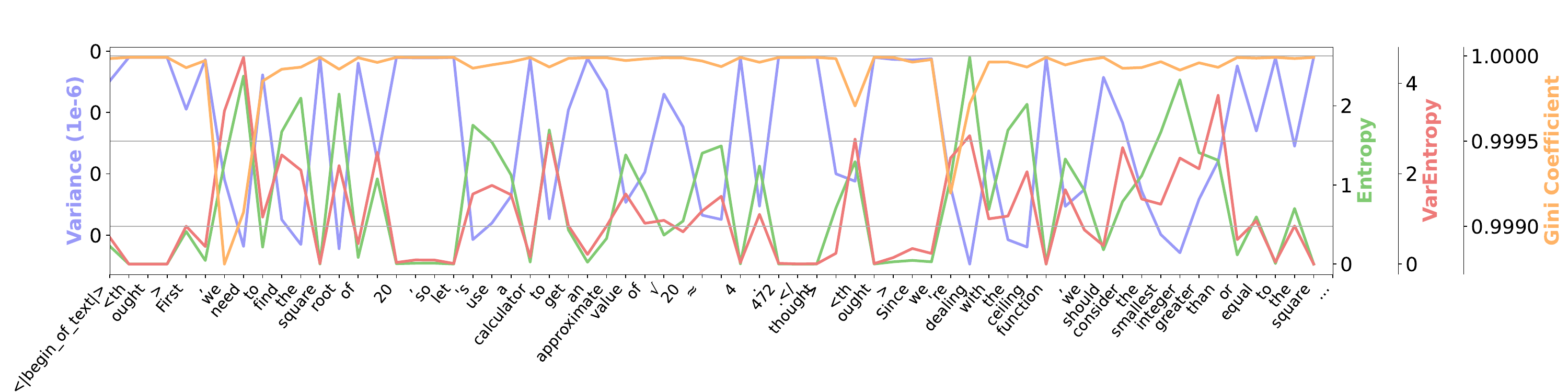}
      \subcaption{\centering \textless thought\textgreater First, we need to find the square root of 20, so let's use a calculator to get an approximate value of $\sqrt{20} \approx 4.472$.\textless thought\textgreater \break \textless thought\textgreater Since we're dealing with the ceiling function, we should consider the smallest integer greater than or equal to the square root of 20, which is the ceiling of $\sqrt{20}$. \textless /thought\textgreater \break \textless thought\textgreater We can also consider the perfect squares that are closest to 20, such as 16 and 25, to estimate the value of $\sqrt{20}$ and then apply the ceiling function.\textless /thought\textgreater }
      \label{fig:token_uncertainty_metrics:1}
  \end{subfigure}
  \begin{subfigure}{\textwidth}
      \centering
      \includegraphics[width=1.0\textwidth]{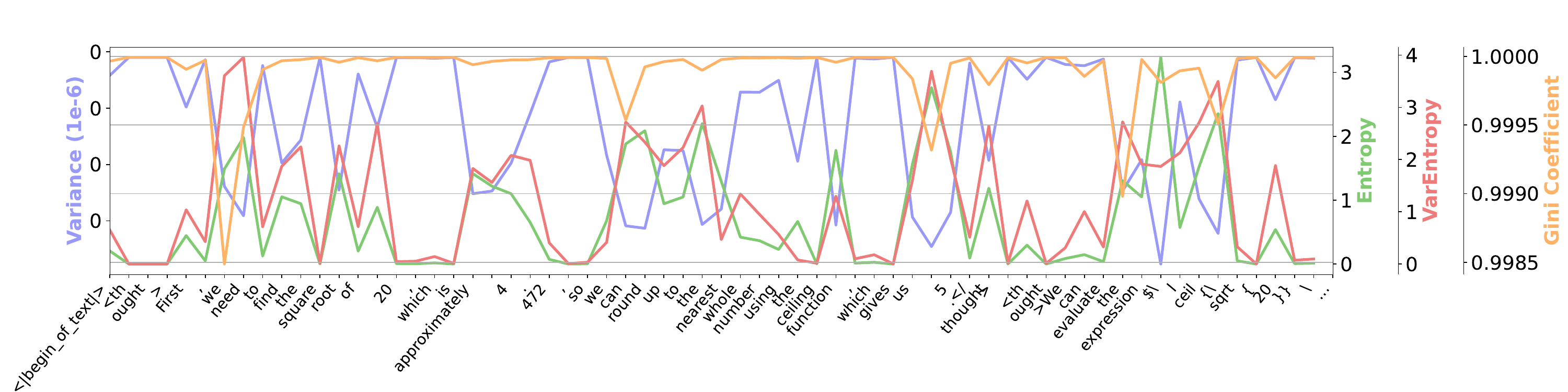}
      \subcaption{\centering \textless thought\textgreater First, we need to find the square root of 20, which is approximately 4.472, so we can round up to the nearest whole number using the ceiling function, which gives us 5.\textless /thought\textgreater \break \textless thought\textgreater We can evaluate the expression $\lceil{\sqrt{20}}\rceil^2$ by first finding the square root of 20, then rounding up to the nearest whole number, and finally squaring the result.\textless /thought\textgreater \break \textless thought\textgreater Since $\lceil{x}\rceil$ is the ceiling function, we can rewrite the expression as $\lceil{\sqrt{20}}\rceil^2 = \lceil{4.472}\rceil^2 = 5^2 = 25$.\textless /thought\textgreater }
      \label{fig:token_uncertainty_metrics:2}
  \end{subfigure}
  \begin{subfigure}{\textwidth}
      \centering
      \includegraphics[width=1.0\textwidth]{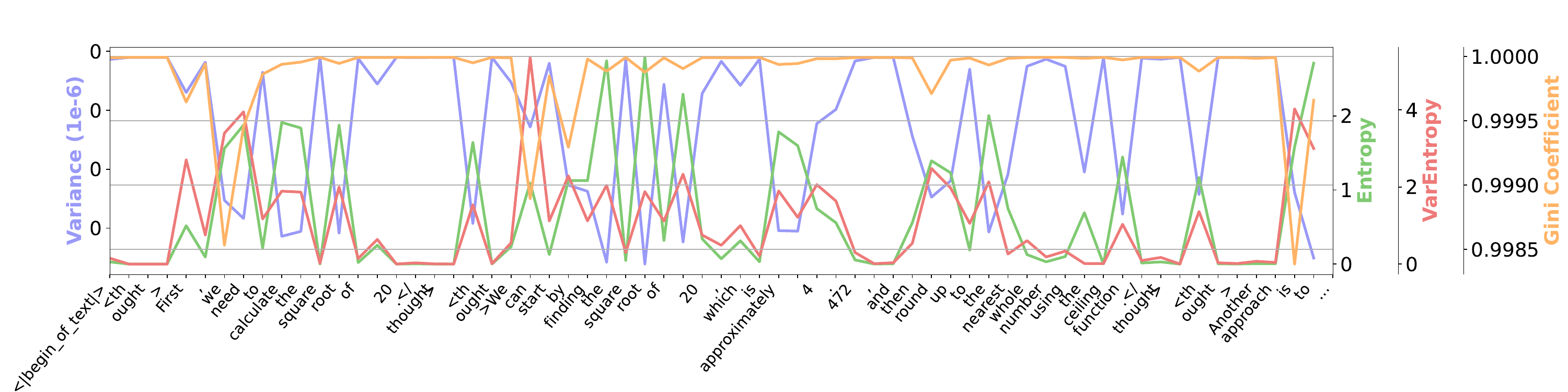}
      \subcaption{\centering \textless thought\textgreater First, we need to calculate the square root of 20.\textless /thought\textgreater \break \textless thought\textgreater We can start by finding the square root of 20, which is approximately 4.472, and then round up to the nearest whole number using the ceiling function.\textless /thought\textgreater \break \textless thought\textgreater Another approach is to recognize that 20 is between the perfect squares 16 and 25, so we can use this information to estimate the ceiling of the square root of 20.\textless /thought\textgreater }
      \label{fig:token_uncertainty_metrics:3}
  \end{subfigure}
  \caption{\textbf{Uncertainty metrics (variance, entropy, VarEntropy, and the
  Gini coefficient) plotted against the first 64 tokens of the output token sequence.}}
  \label{fig:token_uncertainty_metrics}
\end{figure*}

Figure~\ref{fig:token_uncertainty_metrics} further analyzes these results using metrics such as variance, entropy, VarEntropy, and the Gini coefficient. In Figure~\ref{fig:token_uncertainty_metrics:0}, a zero-shot prompt demonstrates lower uncertainty overall, suggesting that it yields more confident predictions and potentially higher-quality outputs. However, the presence of specific high-probability tokens (e.g., ``472'') raises concerns about potential data leakage into the training set or the tokenizer, which could bias the results. Another notable observation is the high uncertainty associated with \textless thought\textgreater tokens, which appear challenging for the model to predict accurately. This highlights the complexity introduced by token granularity, where most words correspond to single tokens, resulting in a roughly even distribution for the next token across the vocabulary in some contexts.

The uncertainty metrics provide actionable insights for reasoning strategy design. For example, cases with high VarEntropy and low entropy indicate a distribution where a few outcomes dominate, making tree-based search strategies effective. These strategies prioritize exploring high-probability outcomes while avoiding unnecessary evaluations of less probable branches. In contrast, low VarEntropy and high entropy reflect a flat distribution where no clear outcome dominates. Such cases could benefit from clarification mechanisms or intermediate step refinements to reduce ambiguity before proceeding further.

Interestingly, the Gini coefficient often highlights critical regions more effectively than other metrics. In vital reasoning areas, it captures the inequality in token probabilities, helping to identify tokens that significantly influence the reasoning process. This contrasts with metrics like entropy and VarEntropy, which may also flag tokens related to formatting or stylistic choices, providing less task-specific utility.

Overall, these visualizations and metrics emphasize the importance of analyzing token probability distributions to design effective reasoning strategies. By leveraging the nuanced patterns revealed by these metrics, models can better adapt to uncertainty, balance exploration and exploitation, and optimize decision-making during the reasoning process.

\fi

\subsection{Benchmarking RLMs}

Our experience with benchmarking RLMs highlights critical considerations for ensuring fair and reliable performance comparisons. Incorporating multiple models within a reasoning scheme often increases output variance, emphasizing the need for benchmarking on sufficiently large sample sizes. Benchmarks with limited sample sizes, such as AIME or AMC, which often provide only a two-digit range of samples, risk selective reporting. This occurs when researchers focus on subsets of results where their models perform well, rather than reflecting the true variability of their systems.

\begin{figure}[t]
    \centering
    \includegraphics[width=0.9\linewidth]{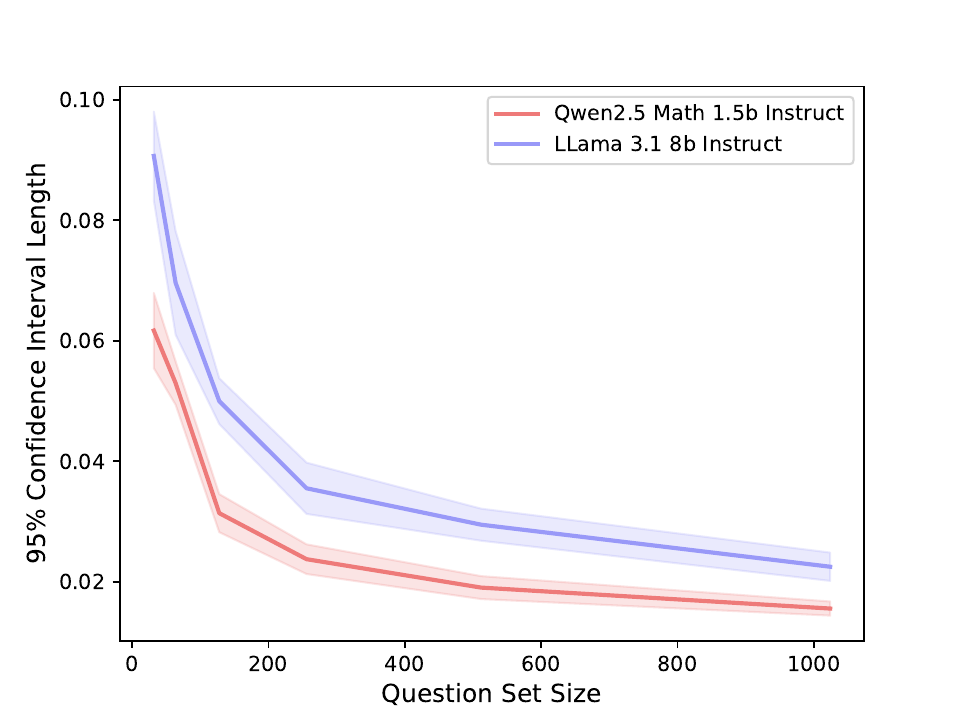}
    \caption{Estimated 95\%-confidence interval length for different question set sizes using sampled generated answers from a subset of 1000 questions with eight generated answers per question at temperature 1. The confidence interval is calculated over the eight different pass@1 subsets of each question with 32 sets randomly sampled with replacement for each set size.} 
    \label{fig:estimated_variance_of_LLM_gen}
\end{figure}

Experimental findings (Figure~\ref{fig:estimated_variance_of_LLM_gen}) demonstrate that achieving low error variability, within a single-digit percentage range, requires evaluation across at least 500 samples. Given the inherent complexity of RLMs, which often exhibit greater variability than simpler LLM setups, these results suggest specific sample size thresholds. We recommend that individual benchmarks contain at least 200 samples per category, with a minimum of 500 samples evaluated across all categories to ensure statistically robust comparisons. Adhering to these guidelines would in many cases mitigate variability-driven biases and facilitate more transparent assessments of RLM performance across different approaches.



%


\section{Example Insights for Effective RLMs}
\label{sec:principles}


We provide example insights gathered from the literature and from our analyses of design decisions using \schemename.

\textbf{Use Process-Based Evaluation}
Process-based evaluation, in which the reasoning structure as a whole is assessed, has been shown to be more reliable than alternative methods such as Outcome-Based Reward Models (ORMs). By examining the reasoning steps and their relationships within the structure, process-based evaluation provides a richer signal that helps models refine their reasoning paths and improve overall accuracy. This approach ensures that each intermediate step contributes positively to the final outcome, resulting in more robust reasoning and better generalization across tasks.

\textbf{Use Two Phases for Training}
Adopting a two-phase training strategy—splitting SFT and RL—has proven effective in several contexts. This phased approach allows the model to first learn a solid foundation of reasoning patterns in phase one, followed by fine-tuning under more complex, adaptive conditions in phase two. For instance, research on process reinforcement through implicit rewards~\cite{cui2025process} demonstrates that models trained with a dedicated SFT phase can maintain performance on standard benchmarks while achieving improved reasoning capabilities during RL. This separation also helps mitigate instability and ensures that each phase targets specific learning objectives, ultimately leading to more robust RLMs.

\textbf{Train on Familiar Distributions}
Training on familiar data distributions can significantly influence a model's initial performance and subsequent improvements.
For example, PRIME~\cite{yuan2024free, cui2025process} shows that training on a carefully curated token sequence (such as the eois token approach) avoids performance degradation. Similarly, in tasks like rStar-Math~\cite{guan2025rstar}, models trained on well-defined, familiar distributions tend to stabilize more quickly and produce higher-quality reasoning outputs. By focusing on familiar distributions, researchers can ensure that the models effectively internalize the fundamental reasoning patterns before moving on to more diverse or challenging tasks.

\textbf{Be Careful with Prompting LLMs to Critique and Evaluate}
Relying on prompting alone to encourage large language models to critique and evaluate their own outputs often leads to instability. Research indicates that models struggle to self-correct reliably when prompted to refine their reasoning without external guidance. For example, a recent study~\cite{huang2023large} illustrates that such prompting typically fails to produce consistently improved results. Another work~\cite{qu2024recursive} demonstrates that explicitly training the model to output better responses through iterative refinement outperforms simple prompting. These findings highlight the importance of structured training approaches and careful operator design when aiming for self-improvement capabilities in RLMs.

\iftr
\section{Benchmarks for RLMs}

We now outline benchmarks related to RLMs.
Sun et al.~\cite{sun2023survey} provide a clear distinction between various types of reasoning including mathematical, logical, casual, and commonsense. Below, we highlight a selection of benchmarks for each category. We also include additional categories related to the realm of RLMs, namely, coding related benchmarks and benchmarks that involve reasoning utilities such as tools or RAG. We show the benchmarks in Figure~\ref{fig:benchmarks}.

\begin{figure*}[t]
\centering
\vspaceSQ{-0.5em}
\iftr
\includegraphics[width=1.0\textwidth]{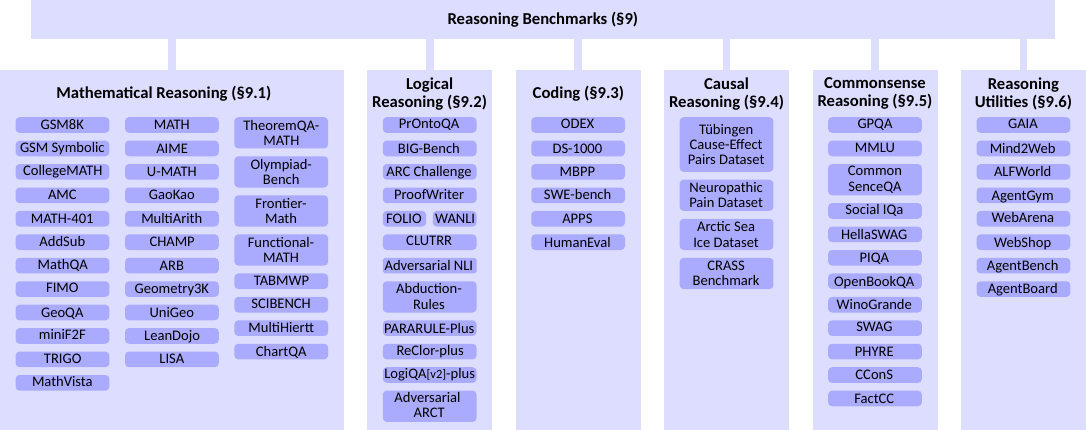}
\else
\includegraphics[width=1.0\textwidth]{PDFs/benchmark_overview_figure-cnf.pdf}
\fi
\vspace{-1.5em}
\caption{\textbf{Overview of benchmarks for RLMs.}}
\vspaceSQ{-1.5em}
\label{fig:benchmarks}
\end{figure*}

\subsection{Mathematical Reasoning}
Mathematical reasoning benchmarks involve arithmetic, geometry, and other mathematical tasks that use logical constructs and symbolic computation. They can be further categorized into benchmarks with fixed datasets and template-based benchmarks~\cite{mirzadeh2024gsmsymbolic, srivastava2024functional}. 

\textbf{GSM8K}~\cite{cobbe2021training} consists of a training set (7,473 samples) and a test set (1,319 samples) of high-quality grade school-level mathematical word problems. Early breakthroughs in mathematical problem-solving by language models were achieved by training on the training subset of this benchmark.

\textbf{GSM Symbolic}~\cite{mirzadeh2024gsmsymbolic} introduces a generator that can use 100 templated questions, which are derived from the questions of the GSM8K dataset. This approach emphasizes the limited generalization capabilities of current RLMs and highlights the importance of templated benchmarks in evaluating LLMs' performance in mathematical reasoning.

\textbf{MATH}~\cite{hendrycks2021math} benchmark contains questions ranging in difficulty from high school to competition-level mathematics, containing 12,500 problems, split into 7,500 for training and 5,000 for testing. These problems are sourced from various mathematics competitions such as the AMC 10, AMC 12, and AIME (Level 5).

\textbf{Functional MATH}~\cite{srivastava2024functional} builds upon the MATH dataset by introducing templated problem formats designed to assess the functional understanding of mathematical concepts by LLMs. However, the code and templates remain inaccessible to the public, limiting its broader adoption.

\textbf{AIME}~\cite{aimo2024amc}, \textbf{AMC}~\cite{aimo2024aime}, and \textbf{GaoKao}~\cite{liao2024mario} feature mathematical tasks ranging from Olympiad level to college entrance level difficulty. The AMC is generally easier, the GaoKao offers a broader range of difficulty levels, while the AIME is likely the most challenging. AIME consists of 30 problems, the AMC includes 40 problems and the GaoKao contains around 300 questions.

\textbf{OlympiadBench}~\cite{he2024olympiadbench} is a more advanced benchmark that spans Olympiad-level mathematics and physics problems, comprising 8,476 problems sourced from international and Chinese Olympiad competitions, as well as the Chinese College Entrance Exam (GaoKao).

\textbf{CollegeMATH}~\cite{tang2024mathscale} is designed for evaluating college-level mathematics, with a dataset that contains 1,281 training problems and 2,818 test problems. These problems are sourced from textbooks, extracted with the help of LLMs.

\textbf{U-MATH}~\cite{chernyshev2024umath} benchmark features 880 university-level test problems without images sourced from ongoing courses across various institutions, currently available through the Gradarius platform. This benchmark presents unpublished, open-ended problems balanced across six core subjects. 

\textbf{FrontierMath}~\cite{glazer2024frontiermath} is an expert-level benchmark containing exceptionally challenging mathematics problems covering a wide array of modern mathematical domains. The dataset size remains undisclosed, but the problems have been carefully crafted and tested by expert mathematicians. Notably, current state-of-the-art models can solve less then 2\% of the problems, revealing a still significant gap between AI capabilities and human expertise in the field of mathematics.

In general, it is recommended to utilize templated versions of these benchmarks where available, rather than relying solely on question-answer (QA) pairs. Templated benchmarks minimize the likelihood of contamination from prior exposure during model training, thus providing a more accurate measure of performance \cite{mirzadeh2024gsmsymbolic, srivastava2024functional}.

Other related benchmarks include MATH-401~\cite{yuan2023how},  MultiArith~\cite{roy2015solving}, AddSub~\cite{hosseinii2014learning}
CHAMP~\cite{mao2024champ}, MathQA~\cite{amini2019mathqa}, ARB~\cite{sawada2023advanced}, FIMO~\cite{liu2023fimo},
Geometry3K~\cite{lu2021inter}, GeoQA~\cite{chen2021geoqa}, UniGeo~\cite{chen2022unigeo}, miniF2F~\cite{zheng2022minif2f}, LeanDojo~\cite{yang2023leandojo}, TheoremQA-MATH~\cite{chen2023theoremqa}, TRIGO~\cite{xiong2023trigo}, LISA~\cite{jiang2021lisa}, MathVista~\cite{lu2024mathvista}, ChartQA~\cite{masry2022chartqa}, TABMWP~\cite{lu2023dynamic}, MultiHiertt~\cite{zhao2022multihiertt}, and SCIBENCH~\cite{wang2023scibench}.

\subsection{Logical Reasoning} 
Logical reasoning emphasizes formal processes, from propositional and predicate logic to automated theorem proving.

\textbf{PrOntoQA}~\cite{saparov2023language} generates ontology graphs, similar to causality graphs, which do not necessarily reflect natural patterns. From these graphs, it constructs statements and poses questions that necessitate logical reasoning for resolution. Due to the abstract and artificial nature of some ontology graphs, models must focus more on step-by-step logical reasoning rather than relying on commonsense inference to derive correct conclusions.

\textbf{BIG-Bench}~\cite{srivastava2022beyond} is one of the most extensive benchmarks for reasoning tasks encompassing over 200 tasks, each potentially comprising numerous questions. It encompasses a broad range of domains and employs templated question formats, enabling a systematic evaluation of reasoning capabilities across diverse contexts.

\textbf{ARC Challenge}~\cite{chollet2019measure} assesses the ability to understand formal patterns, rules, and transformations within structured, grid-based environments. Tasks focus on identifying logical structures such as conditional relationships and sequences. For instance, deducing transformations between grids based on abstract rules exemplifies the application of formal logical reasoning paradigms.

Other benchmarks include ProofWriter~\cite{tafjord2021proofwriter}, FOLIO~\cite{han2022folio}, WANLI~\cite{liu2022wanli}, CLUTRR~\cite{sinha2019clutrr}, Adversarial NLI~\cite{nie2020adversarial}, AbductionRules~\cite{young2022abduction}, PARARULE-Plus~\cite{bao2022multi}, ReClor-plus~\cite{bao2024assessing}, LogiQA-plus~\cite{bao2024assessing}, LogiQAv2-plus~\cite{bao2024assessing} and Adversarial ARCT~\cite{niven2019probing}.

\subsection{Coding}

There also exist benchmarks related to how well a given model can code. These include ODEX~\cite{wang2023execution}, SWE-bench~\cite{jimenez2024swebench}, DS-1000~\cite{lai2023ds1000}, APPS~\cite{hendrycks2021apps}, MBPP~\cite{austin2021program}, and HumanEval~\cite{chen2021evaluating}.

\subsection{Causal Reasoning}  
Causal reasoning involves understanding and analyzing cause-effect relationships, including counterfactual reasoning and causal inference. This domain challenges models to predict or reason about events based on causal dynamics. 

\textbf{Tübingen Cause-Effect Pairs Dataset}~\cite{mooij2016distinguishing} comprises 108 cause-effect pairs drawn from diverse domains such as meteorology, biology, medicine, engineering, and economics. It serves as a comprehensive benchmark for assessing causal reasoning across various contexts.  

\textbf{Neuropathic Pain Dataset}~\cite{tu2019neuropathic} captures complex relationships between nerve function and symptoms in patients. It requires a domain-specific knowledge and causal inference to accurately interpret the data.  

\textbf{Arctic Sea Ice Dataset}~\cite{huang2021benchmarking} consists of a 12-variable graph that models the dynamics of Arctic sea ice based on satellite data generated since 1979. It provides a structured environment to explore causal relationships within climatological systems.  

\textbf{CRASS Benchmark}~\cite{frohberg2022crass} focuses on counterfactual reasoning tasks using 274 sample multiple choice questions. It evaluates models' abilities to answer counterfactual questions, using top-k accuracy as the primary performance metric. 

Many of these benchmarks have either been largely solved by current state-of-the-art models, or their applicability in real-world language model tasks remains limited, rendering them unsuitable for benchmarking current RLMs.

\subsection{Commonsense Reasoning}  
Commonsense reasoning encompasses tasks that require the application of everyday knowledge, including questions that rely on implicit cultural, social, or contextual understanding. This category also extends to specialized domain knowledge tasks.  

\textbf{GPQA (Diamond)}~\cite{rein2023gpqa} is a multiple-choice benchmark spanning disciplines such as chemistry, genetics, biology, and physics. The questions are designed to be solvable by experts (PhDs) within their respective fields but remain challenging for experts from unrelated domains. The diamond subset contains 198 samples.

\textbf{MMLU (STEM)}~\cite{hendrycks2021mmlu} incorporates questions across a spectrum of difficulty, ranging from general commonsense reasoning to highly specialized domain knowledge.  

Other related benchmarks include Social IQa~\cite{sap2019social}, SWAG~\cite{zellers2018swag}, HellaSWAG~\cite{zellers2019hellaswag}, CommonSenceQA~\cite{talmor2019commonsenseqa}, PIQA~\cite{bisk2020piqa}, PHYRE~\cite{bakhtin2019phyre}, OpenBookQA~\cite{mihaylov2018can}, CConS~\cite{kondo2023probing}, WinoGrande~\cite{sakaguchi2020winogrande}, and FactCC~\cite{kryscinski2020evaluating}.

\subsection{Reasoning Utilities}

Benchmarking capabilities of RLMs related to reasoning utilizies involve testing the capabilities of an RLM in how it acts as an agent. This includes benchmarks such as GAIA~\cite{mialon2023gaia}, WebArena~\cite{zhou2024webarena}, Mind2Web~\cite{deng2023mind2web}, WebShop~\cite{yan2022webshop}, ALFWorld~\cite{shridhar2021alfworld}, AgentBench~\cite{liu2023agentbench}, AgentGym~\cite{xi2024agentgym}, and AgentBoard~\cite{chang2024agentboard}. Another line of related benchmarks tests the RAG capabilities~\cite{chen2024benchmarking, xiong2024benchmarking, lyu2024crud, es2023ragas}.




\fi
\section{Related Analyses}




RLMs have been explored from several angles in prior works, yet significant gaps remain in providing a systematic blueprint and open-sourced framework for their construction. Below, we categorize prior efforts and describe how our work advances the field.

\subsection{Reasoning with Standard LLMs}

Several works explore techniques for enhancing the reasoning capabilities of standard LLMs. These approaches use straightforward mechanisms applied during pre-training, fine-tuning or inference.

\textbf{Enhancing Reasoning with Training}
Huang and Chang~\cite{huang2023towards} outline pre-training and fine-tuning on reasoning datasets, and advanced prompting strategies. Sun et al.~\cite{sun2023survey} contribute additional insights, including techniques such as alignment training and the integration of Mixture of Experts architectures. Furthermore, Huang et al.~\cite{huang2022large} demonstrate the possibility of self-improvement on reasoning tasks with additional training on self-generated labels.

\textbf{Reasoning with Prompting \& In-Context Learning}
Qiao et al.~\cite{qiao2023reasoning} provide an overview of prompting-only techniques, classifying prompting methods into two main categories: strategy-enhanced reasoning and knowledge-enhanced reasoning. Besta et al.~\cite{besta2024demystifying} provide a taxonomy of different advanced in-context reasoning topologies. These include the Chain-of-Thought (CoT)~\cite{wei2022chain}, Tree of Thoughts (ToT)~\cite{yao2023tree}, and Graph of Thoughts (GoT)~\cite{besta2024graph}. 

Some of these works further provide overviews of different reasoning tasks, reasoning datasets, and reasoning benchmarks~\cite{huang2023towards, sun2023survey, qiao2023reasoning}. Others focus on enhancing domain-specific reasoning, such as mathematical~\cite{lu2023survey, ahn2024large, yan2024survey} or logical reasoning~\cite{luo2023towards}.

These studies remain largely limited to reviewing existing literature. Therefore, they lack code implementation and rarely employ formal language. Most importantly, they rarely cover explicit reasoning models. Our blueprint integrates most of these techniques within a broader, modular structure.

\subsection{Explicit Reasoning Models}

The following works explore techniques that extend beyond basic mechanisms applied during pre-training or inference. These methods involve iteratively refining reasoning paths, often increasing computational demands during training and/or inference.

Dong et al.~\cite{dong2024survey} provide a taxonomy and survey of inference-time self-improvement methods, including independent, context-aware, and model-aided approaches. Guan et al.~\cite{guan2024search} propose verifier engineering, a post-training paradigm for foundation models involving three stages: Search, Verify, and Feedback, to enhance model outputs with scalable supervision signals.
%
Zeng et al.~\cite{zeng2024scaling} provide a comprehensive roadmap for reproducing OpenAI's o1 reasoning model from a reinforcement learning perspective. Although the work thoroughly examines all core components: policy initialization, reward design, search, and learning, no implementation is provided. Various specific implementations of RLMs exist, we provide a summary in Table~\ref{tab:schemes}. 
There are also other works related to Explicit RLMs, considering both coarse-grained~\cite{xie2024monte, wang2024towards} and fine-grained~\cite{delorenzo2024make, xie2024monte, wang2024towards} reasoning steps.

Our blueprint provides a more foundational and universally applicable framework for RLMs. We further supplement the theoretical and algorithmic overview with a modular and scalable implementation to enable practical development and experimentation.

\section{Conclusion}

This work introduces a comprehensive blueprint for reasoning language models (RLMs), providing a flexible and modular toolbox that demystifies the intricate design and operation of these advanced systems. By encompassing diverse reasoning structures, operations, and training schemes, the blueprint establishes a robust foundation for constructing, analyzing, and extending RLMs tailored to various applications. The accompanying \schemenameS implementation enhances this contribution, offering a modular, minimalist, and user-friendly platform for experimentation and rapid prototyping of novel RLM architectures.

Our blueprint and \schemenameS pave the way for several exciting avenues of future research and development in reasoning AI. One example is Trace-Based Supervision (TBS), which extends Process-Based Supervision by incorporating labeled traces of traversal through reasoning structures. TBS has the potential to train more powerful implicit RLMs capable of internalizing reasoning structures and improving generalization.

The work also explores new directions in value and reward modeling, introducing a hierarchy of models and formally identifying several recent designs as instances of a new class of models, namely the Outcome-Driven Process-Based Reward Model. This model class bridges the gap between outcome-based and process-based evaluation by dynamically connecting intermediate reasoning steps to terminal outcomes, enabling more granular feedback during training without the need for extensive labels.

Additionally, the blueprint's extensive set of operators can inspire the development of innovative reasoning strategies, such as advanced tree-based searches, multi-step refinement processes, or hybrid search algorithms that adapt dynamically to the task's complexity. \iftr These strategies can be tailored using the token probability distribution analysis tools provided, leading to more effective generation strategies that optimize reasoning steps through probabilistic insights.\fi The blueprint also provides a foundation for developing nested architectures where reasoning structures such as trees and graphs are embedded hierarchically. These designs can address multi-layered reasoning tasks, expanding the scope of RLM applications to domains requiring deep, structured reasoning processes.

Scalability remains a key focus of this work. The blueprint's modular design supports future scalable cloud deployments that enable efficient distribution of compute-intensive tasks across cloud infrastructures. These deployments will not only enhance scalability but also optimize cost and resource utilization, making RLMs more accessible for real-world applications. 

By exploring and integrating these ideas, this work aims to empower the next generation of reasoning language models, democratize access to advanced reasoning capabilities, and foster innovation across research and industry. The blueprint's versatility, combined with the \schemenameS platform, will make it one of the factors in the progress in RLM research and applications.

\section*{Acknowledgements}
We thank Nicolas Dickenmann for writing the initial MCTS codebase.
We thank Hussein Harake, Colin McMurtrie, Mark Klein, Angelo Mangili, and the whole CSCS team granting access to the Ault, Piz Daint and Alps machines, and for their excellent technical support. 
We thank Timo Schneider for help with infrastructure at SPCL.
This project received funding from the European Research Council (Project PSAP, No.~101002047), and the European High-Performance Computing Joint Undertaking (JU) under grant agreement No.~955513 (MAELSTROM). This project received funding from the European Union’s HE research and innovation programme under the grant agreement No.~101070141 (Project GLACIATION).
We gratefully acknowledge Polish high-performance computing infrastructure PLGrid (HPC Center: ACK Cyfronet AGH) for providing computer facilities and support within computational grant no.~PLG/2024/017103.

\clearpage
\ifcnf
\setcounter{page}{1}
\fi
\appendices
\section{Mathematical Foundation of Markov Decision Processes for Reasoning Tasks}
\label{sec:appendix_math}

\begin{table*}[ht!]
    \centering
    \caption{Overview of mathematical notation used in the paper.}
    \label{tab:notation}
    \begin{tabular}{@{}ll@{}} 
        \toprule
        \textbf{Symbol} & \textbf{Description} \\ 
        \midrule
        \( \mathcal{M} = (\mathcal{S}, \mathcal{A}, p, r, \gamma) \) & Markov Decision Process (MDP) definition. \\
        \( s \in \mathcal{S} \) & A state in the state space, representing a sequence of reasoning steps. \\
        \( a \in \mathcal{A} \) & An action in the action space, corresponding to selecting the next reasoning step. \\
        \(\mathcal{A}_s \subseteq \mathcal{A} \) & A set of actions available in state $s$. \\
        \( p(s'\mid s, a) \) & The probability of transitioning to state $s'$ from state $s$ by taking action \( a \). \\
        \( r(s) \) & The reward received when arriving in state $s$. \\
        \( \gamma \in [0,1] \) & Discount factor, determining the present value of future rewards. \\
        \( \pi_\theta(a \mid s) \) & Policy parameterized by \( \theta \), representing the probability of taking action \( a \) in state \( s \). \\
        \( V_{\pi}(s) \) & Value function under policy \( \pi \), representing the expected return starting from state \( s \). \\
        \( Q_{\pi}(s, a) \) & State-action value function under policy \( \pi_\theta \), representing the expected return of taking action \( a \) in state \( s \). \\
        \( \tau_\pi \) & A trajectory consisting of states and actions, \( (s_0, a_0, s_1, \ldots, s_{T+1}) \) following policy $\pi$. \\
        \(\mathcal{C}(s)\) & The set of children of state \( s \). \\
        \bottomrule
    \end{tabular}
\end{table*}

In this section, we provide a rigorous mathematical framework for RLMs. We achieve this by integrating the theory of Markov Decision Processes (MDPs) with the Monte Carlo Tree Search (MCTS) algorithm. The MDP serves as a foundational formulation for modeling various types of processes, and it can be applied to model reasoning \textit{chains}, which constitute the reasoning structure of the RLMs. Simultaneously, MCTS serves as an efficient search algorithm for exploring and navigating the extensive space of possible reasoning chains. The resulting state space is then used as a basis for modeling the RLM. An overview of the notation used in this section is provided in Table~\ref{tab:notation}.

\subsection{Markov Decision Process} \label{MDP_def}

A \textbf{Markov Decision Process (MDP)} is defined as a 5-tuple $\mathcal{M} = (\mathcal{S}, \mathcal{A}, p, r, \gamma)$, where $\mathcal{S}$ is the state space, $\mathcal{A}$ is the action space with $\mathcal{A}_s \subseteq  \mathcal{A}$ denoting the set of actions which can be taken in the state $s$, $p$ represents the dynamics of transitions between states, i.e., $p: \mathcal{S} \times \mathcal{A} \times \mathcal{S} \rightarrow [ 0, 1 ]$ where $p(s, a, s')$  is the probability of transitioning to state $s'$ when action $a$ was selected in state $s$, $r: \mathcal{S} \times \mathcal{A} \times \mathcal{S} \rightarrow \mathbb{R}$ is the reward function, i.e., $r(s,a,s')$ represents the reward for arriving in state $s'$ after selecting action $a$ in state $s$, and $\gamma \in [0,1]$ is a discount factor.

\subsubsection{Solving an MDP}

Before stating what it means formally to \textit{solve an MDP}, we first need several definitions.

A \textbf{trajectory} $\tau_{\pi} = (s_0, a_0, \ldots, s_T, a_{T}, s_{T+1})$ is a sequence of interleaved states and actions, selected according to the policy $\pi$ (see below for the policy definition).
Each trajectory starts at an initial state $s_0 \in \mathcal{S}$ and ends with $s_{T+1}\in \mathcal{S}$ which represents the terminal state where no further actions can be taken.

A \textbf{policy $\pi(s)$} is a function assigning a probability distribution over the action space to a given state $s$; $\pi: \mathcal{S} \rightarrow \Delta(\mathcal{A})$ where $\Delta(\mathcal{A})$ is a set of probability distributions over action space $\mathcal{A}$. The expression $\pi(a\mid s )$ denotes the probability of selecting the action $a$ in the state $s$ according to the policy $\pi$.

A \textbf{state value function $V_\pi (s_t)$} represents the expected \textit{cumulative} future reward for a given state $s_t$ under policy $\pi$:
\[
    V_\pi(s_t) = \mathbb{E}\left[ \sum_{k=t}^{T}\gamma^{k-t} r(s_k,a_k,s_{k+1} ) \mid s_t  \right],
\]
where $T$ is a predefined time-horizon. Note that, in order to obtain the state $s_{k+1}$, an action $a_k$ is first derived by sampling from a distribution $\pi(s_k)$. Once the action $a_k$ is chosen, the environment dynamics $p(s_{k+1} \mid s_k, a_k)$ determine the probability distribution of the next state $s_{k+1}$.

The goal of \textbf{solving an MDP} is to find a policy $\pi^*$ which maximizes the value function as defined above for all states $s \in \mathcal{S}$, $\pi^{*} = \underset{\pi}{\arg\max}\  V_{\pi}\left(s\right)$.

Oftentimes, it is useful to use a \textbf{state-action value function $Q(s_t, a_t)$} instead of a state value function. Specifically, the state-action value function $Q(s_t, a_t)$ extends the state value function so that the function value is defined on a state \textit{and} a specific action $a_t$:
\begin{align*}
    Q_{\pi}(s_t, a_t) &= \mathbb{E}_{\pi} \left[ \sum_{k=t}^{T} \gamma^{k-t} r(s_k, a_k, s_{k+1}) \mid s_t, a_t \right] \\
    &\stackrel{}{=} r(s_t, a_t) + \gamma \mathbb{E}_{s_{t+1}} \left[ V_{\pi}(s_{t+1}) \mid s_t, a_t \right],
\end{align*}
where Bellman's equation is used in the second equality.

\subsubsection{MDPs in the RLM Setting}

In the context of RLMs, a state $s \in \mathcal{S}$ is typically defined as a sequence of reasoning steps $s = (z_0 \dots z_n)$, where each reasoning step $z_i$ is a sequence of $M_i$ tokens $z_i = (t_i^0, \dots, t_i^{M_i})$. Each $t_i^j$ is a token from the RLM's vocabulary, and the total number of tokens per reasoning step $M_i$ can vary. One can use a special token $t_{M_i} = t_{end}$ to indicate the end of the reasoning step. Typically, the initial query $q$ is used as the first reasoning step $z_0=q$. In the study of RLMs, an action $a \in \mathcal{A}_s$ usually represents appending a new reasoning step $z^{(a)}$ to the current state $s = (z_0, ..., z_n)$  resulting in a new state $s' = \left(z_0, ..., z_n, z^{(a)}\right)$. Since every action  $a$ is uniquely associated with exactly one reasoning step $z^{(a)}$ for every $s=(z_0, ..., z_{n})$ and $s'=(z_0, ..., z_{n}, z_{n+1})$, we have
\begin{equation*}
  p(s, a, s') =
    \begin{cases}
      1 & \text{ if } z_{n+1} = z^{(a)} \, \\
      0 & \text{ if } z_{n+1} \neq z^{(a)} \\
    \end{cases}.
\end{equation*}

The definition of the reward function depends on the specific task. A reward commonly seen in reasoning tasks assigns non-zero reward only in the terminal states and hence only at the final reasoning step. This approach reflects the fact that for most tasks, only the final answer can be evaluated against the ground truth solution to the original query. We call such reward functions \textit{sparse} to clearly distinguish it from other settings in which intermediate rewards can be observed by the algorithm in the non-terminal states.
The discount factor $\gamma$ determines how future rewards influence the current decision-making process. A higher discount factor ($\gamma \rightarrow 1$) places greater emphasis on long-term reasoning success, allowing the model to generate long reasoning sequences, while a lower discount factor prioritizes immediate rewards, incentivizing faster progress and shorter reasoning sequences.

In the RLM setting, a trajectory $\tau_\pi = (s_0, a_0, \ldots, s_T, a_T, s_{T+1})$ represents the progression of states $s_t$ and actions $a_t$ ending with a terminal state $s_{T+1}$ in which no further reasoning steps can be added. The final reasoning step contains the RLM's answer to the original query.


The policy \(\pi(a \mid s)\) in the context of RLMs defines the probability of selecting an action \(a\) that corresponds to appending a reasoning step \(z^{(a)}\) to the current reasoning sequence represented by the state \(s\). Since there exists a bijective mapping \(f: \mathcal{A} \to \mathcal{Z}\) between the action space \(\mathcal{A}\) and the reasoning step space \(\mathcal{Z}\), the probability distributions can be equated using the change of variables. Formally:
\[
\pi(a \mid s) = \pi(z \mid s), \quad \text{where } z = f(a).
\]


Based on the definition of the reasoning step and applying the chain rule we can then rewrite the policy as: 
\[
\pi(z_{t+1} \mid s_t) = \prod_{j=0}^{M_{t+1}} \pi(t_{t+1}^j \mid s_t, z_{t+1}^0, \dots, z_{t+1}^{j-1}).
\]



In the RLM setting, the state value function $V(s_t)$ assesses the expected cumulative reward of a partial reasoning sequence $s_t$, estimating its overall potential to lead to a successful solution. The state-action value function $Q(s_t,a_t)$  extends this by quantifying the expected cumulative reward for taking a specific action $a_t$ (e.g., appending a reasoning step $z_{t+1}$) to the current state $s_t$ and then following the policy $\pi$. It incorporates both the immediate reward for appending the reasoning step and the anticipated future rewards from completing the reasoning sequence.
Together, these functions inform and guide the policy $\pi$ to prioritize actions that maximize the expected cumulative reward. By leveraging $V(s_t)$ or $Q(s_t,a_t)$, the policy can be trained to select reasoning steps that progress toward correct and complete solutions, transforming an LLM into a RLM.




\subsection{Monte Carlo Tree Search}
\label{sec:MCTS_math}

\textbf{Monte Carlo Tree Search (MCTS)} is a heuristic search algorithm used for solving MDP problems. MCTS iteratively builds a search tree, representing the underlying MDP state-action space, by aggregating the information obtained from executed MDP trajectories. Let  \( \mathcal{T}  = (N, E) \) denote the MCTS search tree where \( N \subseteq \mathcal{S}\) is the set of nodes and \( E \subseteq N \times \mathcal{A} \times N \) is the set of directed edges between the nodes. Every node in the MCTS search tree corresponds to a single state in the MDP and every edge corresponds to a single action. Every path from the root to the leaf of the search tree \( \mathcal{T} \) corresponds to a single trajectory in the underlying MDP.

\textbf{Edge statistics} The MCTS algorithm stores the following values for every edge \( s, a \) in the search tree:
\begin{itemize}[noitemsep, leftmargin=0.75em]
    \item \( N(s,a) \) - the visit count of the edge \( (s,a)\) by the algorithm,
    \item \( q(s,a) \) - the estimated state action value of \( (s,a) \),
    \item \(r(s, a) = r(s, a, s') \) - the reward received after taking action \( a \) in state \( s\) leading to state \( s' \),
    \item \(\beta(s,a)\) - the terminality function indicating if action $a$ leads to a terminal state.
\end{itemize} 

\textbf{Algorithm} At the high level, the MCTS begins by initializing the tree with a single starting state $s_0$ as a root node and performing the following three phases in a loop:
\begin{enumerate}[noitemsep, leftmargin=0.75em]
    \item \textbf{Selection} - a leaf node in the current tree is selected for expanding its child (children).
    \item \textbf{Expansion} - if the selected node does not correspond to a terminal state, it is expanded by taking an action (or multiple actions) in the underlying MDP and by adding the resulting state (states) to the tree as children of the current node. A trajectory unroll is performed for every added node to obtain a reward. ``Unroll'' refers to simulating a sequence of steps from a newly added node in the tree down to a terminal state. This simulated trajectory represents a hypothetical path the system might take if it continued from the current node. Once the simulation reaches a terminal state, a reward value is calculated based on the outcome of that path.
    \item \textbf{Backpropagation} - update the value estimates and the visit counts for the selected node and all its ancestors based on the obtained reward.
\end{enumerate}

The MCTS algorithm finishes when the stop criterion such as the the number of iterations, the predefined computational budget, or the convergence criterion is met.

\section{Value and Reward Models}
\label{sec:value-reward-models}

We now proceed to discuss the details of value and reward models.

\subsection{Outcome- vs. Process-Based Reward Models}
\label{PBSvsOBS}

\begin{figure}[t]
    \centering
    \includegraphics[width=0.9\linewidth]{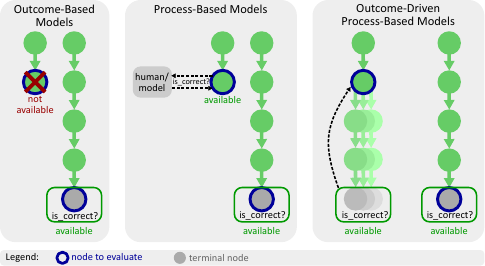}
    \caption{Comparison of outcome- vs.~process-based label generation, and the introduction of \textbf{Outcome-Driven Process-Based Reward Models (O-PRMs)}. Gray nodes mark terminal nodes.}
    \label{fig:orm_prm}
\end{figure}

In reinforcement learning environments, reward models estimate the reward for taking an action $a$ in state $s$ which leads to state $s'$. For reasoning tasks and algorithms like MCTS, which rely on evaluating intermediate steps, it is essential to have models capable of estimating the quality of each step. Two primary families of reward models for such tasks are Outcome-Based Reward Models (ORMs) and Process-Based Reward Models (PRMs). Figure~\ref{fig:orm_prm} compares both classes of models.

\textbf{Outcome-Based Reward Models}, first introduced by Uesato et al.~\cite{uesato2022solving}, evaluate the reasoning process solely based on the final outcome. These models estimate the reward of the final step in the chain, often modeled in the literature as the likelihood of a correct final answer given the entire reasoning chain $P(correct(z_{T+1} )\mid z_0,...,z_{T+1})$~\cite{uesato2022solving, lightman2023let} where $s_{T+1} := z_0,...,z_{T+1}$ is the complete reasoning chain consisting of reasoning steps $z_i$ and $T+1$ marks the last reasoning step.
ORMs are particularly ill-suited for evaluating intermediate steps for several reasons. First, the training data and objective are inherently misaligned with step-wise evaluation, as they focus exclusively on final outcomes. Second, ORM evaluations tend to be overly pessimistic for intermediate steps since a subsequent erroneous step can obscure the correctness of earlier steps. This observation aligns with Havrilla et al.~\cite{havrilla2024glore}, who noted that ORMs often underestimate the solvability of a problem from an intermediate state and are prone to a high false-negative rate. Furthermore, ORMs lack robustness against false positives, potentially favoring erroneous reasoning steps and misleading the evaluation process.

\textbf{Process-Based Reward Models}, introduced by Lightman et al.~\cite{lightman2023let} and Uesato et al.~\cite{uesato2022solving}, evaluate reasoning on a step-by-step basis. These models estimate the reward of a step, which can be seen as the likelihood of correctness for the $t$-th step given its preceding context $P(correct(z_t )\mid z_0,...,z_t)$ where $s_t := z_0,...,z_t$ is a potentially incomplete reasoning chain and $z_i$ are reasoning steps and $z_0$ is the query. PRMs provide more fine-grained feedback and can pinpoint errors in the chain. This step-wise evaluation provides dense rewards given partial responses and helps identify where reasoning deviates from correctness, offering improved interpretability and enabling more targeted improvements in reasoning processes. However, PRMs are computationally expensive to train and require extensive annotations of reasoning steps. These annotations, whether provided by humans or other LLMs, often suffer from limitations: human annotations are scarce, costly, and prone to bias, while prompted LLM-generated annotations~\cite{wang2023mathcoder} are typically of lower quality due to their limited self-evaluation capabilities~\cite{madaan2023self}. Automated methods using for example MCTS such as~\cite{wang2024math, luo2024improve} introduce large computational costs and are prone to false negatives.

\subsection{Outcome-\textit{Driven} Process-Based Reward Models} 

Motivated by the need for process-based reward models but constrained by the lack of annotated step-wise labels, certain models that we will refer to as \textit{Outcome-Driven Process-Based Reward Models (\textbf{O-PRMs})} have been proposed; they \textit{combine outcome-based signals with process-based objectives}. We also illustrate these models in Figure~\ref{fig:orm_prm}. These models rely on process-based data, often automatically generated using MCTS algorithms, where simulations starting from a given step $s_t$ are performed. The final correctness of these simulated paths is aggregated to create step-wise labels~\cite{wang2024math, luo2024improve} (for other, non-MCTS approaches see~\cite{havrilla2024glore}). This automation enables scalable data generation for O-PRMs, eliminating the need for extensive human annotation. Although O-PRMs can be categorized as process-based models due to their approximation of step-wise rewards, they remain inherently tied to outcome signals. Some authors~\cite{uesato2022solving} suggest that, under certain conditions, outcome signals in mathematical domains can approximate intermediate labels. However, O-PRMs inherit many limitations of ORMs, including susceptibility to false negatives, false positives, and an over-reliance on terminal outcomes. While the aggregation of multiple simulations helps reduce variance, the backtracking process may still oversimplify complex dependencies within reasoning chains.

\subsection{Reward Models vs. Value Models}\label{sec:Model_types}
\begin{figure}[t]
    \centering
    \includegraphics[width=0.85\linewidth]{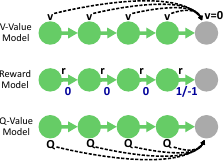}
    \caption{Comparison of reward, v-value and q-value models in a sparse reward setting (only terminal states receive non-zero rewards). Gray nodes mark terminal nodes. The reward model should predict the rewards for transitioning from one state to another which is 0 for non-terminal states. V-VMs and Q-VMs however, predict a global value and are therefore informative for non-terminal states.}
    \label{fig:reward_q_v}
\end{figure}

While the distinction between reward models and value models is often blurred in the literature—and their terminology is sometimes used interchangeably—we explicitly differentiate between these model types for evaluating reasoning steps. Additionally, we distinguish two variants of value modes: v-value and q-value models. This differentiation arises from the distinct roles these models play in reinforcement learning environments. We provide an overview over the differences between reward and value models in Figure~\ref{fig:reward_q_v}.

\subsubsection{Reward Models}

A Reward Model (RM) predicts immediate rewards. In RL, this corresponds to the reward obtained for a transition $(s,a, s')$ from state $s$ when taking action $a$ which results in step $s'$. For reasoning, this corresponds to adding a new reasoning step $a$ to the structure. The new structure is then represented by $s'$. Specifically, PRMs -- which are preferred over ORMs for MCTS due to the need for action-based evaluation -- learn these rewards and can be used to evaluate states (or the transition into a state). This formulation provides a localized, step-level evaluation independent of the overall outcome of the reasoning chain. The reward model is typically trained using labeled data where individual reasoning steps are associated with reward values. While this localized view is advantageous for step-by-step evaluation, it lacks the ability to consider how the current step contributes to the long-term success of the reasoning process. This limitation motivates the introduction of value models.

\subsubsection{Value Models}
\label{sec:value_model}

Value Models (VMs) provide a more abstract, global evaluation of states and actions by estimating their contribution to future rewards. Unlike reward models, which focus on immediate outcomes, value models consider both current and future rewards, enabling a broader perspective on reasoning quality.
For example in reinforcement learning and MCTS, value models play a critical role in guiding the search process. By providing estimates of state or state-action values, they enable more informed decisions about which nodes to expand and explore.
%
%
We now discuss variants of value models.

One such variant is the \textbf{V-Value Model (V-VM)}, which predicts the expected cumulative future reward of a state, denoted as $V(s)$. This is equivalent to the state value function in reinforcement learning, which evaluates the long-term potential of the current state $s$. A key advantage of V-VMs is their global perspective, as they aggregate future rewards across all possible trajectories originating from the current state. However, V-VMs do not explicitly evaluate individual actions, which may limit their utility in step-level decision-making. Additionally, v-values are often ill-defined at terminal states, where rewards may substitute for state values during training.

\textbf{Q-Value Models (Q-VMs)} are another variant, which predict instead the expected cumulative future reward of taking a specific action $a$ in a given state $s$, denoted as $Q(s,a)$. Unlike V-VMs, Q-VMs explicitly associate values with state-action pairs, offering a more granular evaluation.
This granularity makes Q-VMs particularly useful for MCTS, where decisions about which edge (action) to expand at a given node (state) are critical. By directly evaluating actions, Q-VMs align naturally with the selection mechanisms in MCTS, guiding the search toward promising paths. Similar to V-VMs, Q-VMs can also be categorized as Process-Based Q-Value Models (PQVMs), Outcome-Based Q-Value Models (OQVMs), and Outcome-Driven Process-Based Q-Value Models (O-PQVMs).

The choice between V-VMs and Q-VMs depends on the reasoning task and the specific requirements of the evaluation framework. While V-VMs provide a broader, state-centric evaluation, Q-VMs enable more precise, action-specific guidance. In practice, MCTS often benefits from the use of Q-VMs due to their compatibility with edge-based selection.

\subsubsection{Example: Solving a Mathematical Equation}  

To illustrate the differences between reward models, value models, and q-value models, consider the task of solving \(x^2  + y^2= 1\) step-by-step.
\begin{itemize}[noitemsep, leftmargin=0.75em]
    \item \textbf{Reward Model:}  A process-based reward model might assign a reward \(r(s_t, a_t, s_{t+1})\) for the reasoning step \(a_t = \text{"Substitute } y = \sqrt{1 - x^2} \text{"}\). This reward quantifies the quality of the resulting state \(s_{t+1}\), independent of whether it leads to a correct solution. However, in sparse reward settings (only final steps receive a reward), this reward would be $0$.
    \item \textbf{V-Value Model:}  A V-VM estimates \(V(s_t)\), representing the expected cumulative reward for the entire expected solution process starting from \(s_t\). For instance, if \(s_t = (\text{"Start with } x^2 + y^2 = 1\text{"})\), \(V(s_t)\) considers the long-term potential of all reasoning paths originating from this state.
    \item \textbf{Q-Value Model:}  A Q-VM evaluates \(Q(s_t, a_t)\), predicting the cumulative reward of taking a specific action \(a_t\) (e.g., substituting \(y = \sqrt{1 - x^2}\)) in state \(s_t\). This value directly informs whether the action \(a_t\) is likely to lead to a high-quality solution, providing a more granular evaluation compared to the V-VM.
\end{itemize}

\subsubsection{Summary}

By differentiating reward models and value models, and further categorizing value models into V-VMs and Q-VMs, we provide a nuanced framework for evaluating reasoning steps. Reward models offer localized evaluations, while value models incorporate global, long-term perspectives. This global evaluation enables the model to better prioritize reasoning paths that are likely to lead to correct solutions while mitigating the challenges posed by sparse or delayed rewards. Therefore, we advocate for the use of a process-based value model due to the sparsity of reward signals for reasoning tasks. Among value models, Q-VMs are particularly well-suited for MCTS due to their action-specific granularity, which aligns naturally with the tree’s edge-based exploration mechanism. 
We will demonstrate the practical implications of these distinctions in Appendix~\ref{QM_data}.

\subsection{Evaluation Schemes}

We also provide additional categorizations and details regarding overall evaluation.

\subsubsection{Evaluation Types}

Evaluating reasoning steps in RLMs involves assessing their quality and contribution toward solving a task. Numerical evaluations can be categorized as relative or absolute. 

\textbf{Relative evaluations} compare multiple steps, often using ranking mechanisms and can be created with, for example, the Bradley-Terry model \cite{bradley1952rank}, which is optimized based on pairwise preferences by maximizing the reward gap between chosen and rejected steps. 
%

\textbf{Absolute evaluations} assign scalar values to each step, assessing aspects such as coherence, correctness, or helpfulness, using regression-based models. Moreover, evaluation dimensions can also be modeled as binary with classification models. While regression models provide more information, classification models capture correctness more naturally since a statement is usually correct or incorrect. On the other hand, the former ones are more suitable for measuring quality, such as the degree of coherence. Depending on the specific quality being evaluated, the choice between regression and classification models should align with the evaluation's goals. Additionally, absolute scores can be transformed into rankings if needed, providing flexibility across various applications.

In addition to numerical evaluations, there are \textbf{text-based evaluations}, which are commonly used to provide detailed feedback and guidance for refining reasoning steps. Examples include ``LLM-as-a-Judge''~\cite{zheng2023judging} (which uses a larger LLM to provide a pairwise comparison or a single graded answer with an explanation) and self-critique approaches~\cite{saunders2022self} that allow models to reflect on and evaluate their own reasoning. These textual evaluations, often including rationales, are particularly useful for structural transformations rather than numerical guidance, enhancing interpretability by offering context and detail.

\subsubsection{Evaluation of Reasoning Steps}
\label{sec:eval_steps}

Step-wise evaluations are vital for integrating reasoning into MCTS. Numerical evaluations-—whether relative or absolute-—provide straightforward metrics to compare nodes and steer exploitation and exploration. Text-based evaluations, in contrast, are better suited for guiding structural refinements rather than directly influencing search paths.

Given that reasoning steps are typically textual sequences, language models are a natural fit for such evaluation tasks. LLM-based approaches can involve \emph{external model} approaches, where a dedicated value model is trained to predict scores, or \emph{internal model} approaches, which leverage existing policy models.

\textbf{External model approaches} include value models that predict scalar reward signals (reward models)~\cite{christiano2023deep, uesato2022solving, lightman2023let}, reinforcement learning values like state-values (v-value models)~\cite{silver2018general}, state-action values (q-value models), or pairwise models like the Bradley-Terry and PairRM~\cite{jiang2023llm} frameworks. A more detailed comparison of reward models, v-value, and q-value models can be found in Appendix~\ref{sec:value_model}.

There exist a large range of \textbf{internal model approaches} as substitutes for value models. They typically rely on methods like prompting the policy model to output scores. Examples include MCT Self-Refine (MCTSr)~\cite{zhang2024accessing}, querying for a binary feedback (e.g., ``Is the answer correct? answer yes or no'')~\cite{zhang2024generative} and evaluating the probability of the output, leveraging uncertainty metrics such as token entropy or aggregated probabilities~\cite{zhao2024marco}, and others~\cite{zhang2024restmcts}.

\textbf{Heuristics} may also serve as substitutes for evaluations in resource-constrained scenarios.

\textbf{Simulating} reasoning steps to terminal states for evaluation against golden answers is another option as done for example in MCTS, though often computationally prohibitive.

\textbf{External tools} provide an alternative path for evaluation, especially in domain-specific tasks. For programming, compilers can supervise tasks, as seen in Codex~\cite{chen2021evaluating}, self-debugging~\cite{chen2023teaching}, and similar methods. Program-of-Thought~\cite{chen2023program} and Program-aided-Language (PAL)~\cite{gao2023pal} use a formal language and Python interpreters to evaluate solutions. In mathematical tasks, ensemble approaches like MathPrompter~\cite{imani2023mathprompter} generate multiple algebraic expressions or Python functions to validate steps. These tool-based approaches excel at detecting errors due to their reliance on precise domain-specific rules, such as compilers for programming or interpreters for mathematics. While their applicability is limited to well-defined domains, they provide objective and verifiable feedback that complements language models. By injecting precise knowledge into the evaluation process, external tools mitigate model-specific limitations like hallucinations and offer actionable feedback for iterative refinement. This hybrid approach enhances reliability and ensures that the evaluation benefits from both the flexibility of language models and the precision of formal systems.


\section{Algorithmic Descriptions}
\label{sec:appendix-algo}

\subsection{Reasoning with Monte Carlo Tree Search}
\label{sec:mcts_algo_description}
\subsubsection{Setup and Notation}
We will now present the details of the \schemenameAS training pipeline.

The \textbf{MDP Design} of \schemenameAS follows the definition presented in Appendix~\ref{MDP_def} with the $\gamma$ values between $[0.95, 1]$ to avoid over-penalizing long reasoning sequences. In the RLM setup, the state and action spaces of the underlying MDP constitute a tree in which every state $s$ other than the starting state $s_0$ has exactly one action $a_s$ leading to it. This allows us to simplify the notation by omitting actions wherever it's clear from the context that we are referring only to an action leading to a given state. For every action $a$ leading from the state $s$ to the state $s'$ we will write:
\begin{align*}
\pi(s' \mid s) &:= \pi(a_{s'} | s) \\
r(s') &:= r(s,a,s') \\
q(s') &:= q(s,a) \\
\tau &:= (s_0, s_1, \dots, s_{T+1})
\end{align*}


The final reasoning step in the terminal state contains the RLM's answer to the original query. The final answer is compared to the ground truth solution, commonly referred to as the golden answer. 
This matches the common setup in many reasoning tasks and math problems, where no ground truth and no reward source is available for the intermediate reasoning steps.


Consider a trajectory $\tau := (s_0, s_1, \dots, s_{T+1})$. We assign a reward of $r(s_{T+1}) = 1$ if the last reasoning step in the final state $s_{T+1}$ contains the correct answer and $r(s_{T+1})=-1$  otherwise. The state value function simplifies to
\begin{align*}
V_{\pi}(s_t) &= 
\mathbb{E}_{\pi}\left[\gamma^{T-t} r(s_{T+1}) \right] \quad \in [-1, 1]
\end{align*}
and the state action function can be rewritten as:
\begin{equation}
    Q_{\pi}(s_t)\stackrel{}{=} 
    \begin{cases} 
        r(s_{T+1}), & \text{if } t = T + 1 \\
        \gamma V_{\pi}(s_{t+1}), & \text{otherwise}
    \end{cases} \quad \in [-1, 1] \label{eq_qv}
\end{equation}
hence both the value and the state-action value functions are bounded between -1 and 1 for all states and state-action pairs.

\textbf{MCTS Design} We define the MCTS tree as in Appendix~\ref{sec:MCTS_math} as $\mathcal{T}=(N,E)$, where $N$ is a set of nodes, and $E$ is the set of edges.
We use the notation of a node-edge-node relationship denoted by \((s, a', s')\) where \(s\) represents the origin node, \(a'\) describes the action corresponding to an edge, and \(s'\) denotes the target node. This notation symbolically ties the action and the target state together, as the action uniquely identifies the target state and is therefore indicative of it.

We use a pre-trained LM with parameters $\theta$ as a \textbf{policy model}, which we denote as $\pi_\theta$. The model autoregressively generates a sequence of tokens. We use a special token 'End of Intermediate Step' (eois) to indicate the end of the reasoning step. We use a standard end-of-sequence (eos) token to indicate the end of the final reasoning step concluding the reasoning trajectory.

A parametric \textbf{value model} is used to evaluate the quality of states. While MCTS traditionally approximates these values through extensive simulations, such an approach is computationally expensive and impractical in the RLM context.
Inspired by AlphaZero~\cite{silver2018general}, which replaces simulations with a parameterized value model, we estimate the state-action values, i.e. the q-values, for reasoning sequences using a value model — effectively employing a \textbf{process-based q-value model $Q_\phi$} (see Appendix~\ref{sec:Model_types}).
The value model is instantiated as a pre-trained transformer-based LM, modified by adding three linear layers and a shifted, rescaled sigmoid activation to align the output domain to the state action function domain $[-1,1]$ (see Eq.~\ref{eq_qv}). 
This setup proved more stable than alternatives, such as a tanh activation or a cropped linear layer. We will show in the following how such a model can be trained and provide a description for the data generation process in Appendix~\ref{sec:appendix_data}. During training, we assume access to a final answer verifier, which evaluates the correctness of the model's final answer and provides the true reward.

\subsubsection{MCTS Algorithm}

We now present the algorithmic steps of a Monte Carlo Tree Search variant similar to AlphaZero as implemented in the \schemenameAS reasoning framework, which we detail in Algorithm~\ref{alg:mcts_star}. The MCTS search operates in two distinct modes: training and inference. The core difference is that, during training, a final answer verifier evaluates and scores the final reasoning steps, providing a true reward signal that is backpropagated through the MCTS tree. This reward serves as a reliable learning signal for the value model $Q_\phi$. During inference, however, the verifier is unavailable, and decisions rely solely on the value model.

\textbf{Notation} We chose to store all values in nodes instead of edges, which defines the following set of statistics saved for each node \(s\):
\begin{itemize}[noitemsep, leftmargin=0.75em]
    \item \( N(s) \) - the visit count of node \(s\),
    \item \( q(s) \) - the running estimate of the q-value of the transition leading to state \(s\),
    \item \(\beta(s)\) - the binary terminality function, returns $1$ if the node $s$ is terminal and $0$ otherwise.
\end{itemize}  

The \textbf{Selection} phase iteratively identifies the most promising child node with a selection policy, which in \schemenameAS is a node-based variant of the PUCT algorithm in AlphaZero~\cite{silver2017mastering} (which is defined on edge-based values) without a prior for selecting a child of $s$:
\begin{align*}
        \argmax_{s_c\in\mathcal{C}(s)} \quad  q(s_{c}) +\frac{\sqrt{N(s)-1}}{1 + N(s_c)} \cdot \left(c_1 + \log{\frac{N(s) + c_2}{c_2}} \right),
\end{align*}
where $c_{1}$ and $c_{2}$ are hyperparameters controlling the exploration bias, and the other values can be taken from the node statistics.

\textbf{Expansion}
We append $M$ nodes to the selected leaf, $M$ being a hyperparameter. 
One of the major challenges in applying RLMs is maintaining the diversity of reasoning paths.
By adding $M$ nodes, we increase the exploration of alternative reasoning paths.

The \textbf{Backpropagation} step serves to propagate information from the terminal nodes back to their ancestors.
In our implementation, we update the running estimates of the q-values using the following formula:
\begin{align*}
    q(s) \leftarrow &(1-\alpha) q(s) + \alpha \gamma \left( \sum_{s_c\mathcal{C}(s)}w_s(s_c) \cdot q(s_c)\right) ,
\end{align*}
where we look at the node-edge-node tuples $(s,a_c,s_c)$ and $s_c\in \mathcal{C}(s)$. The weights \(w_s(s_c)\) for combining the children q-values are defined over the visit scores of the nodes as follows:
\begin{align*}
    w_s(s_c) = \frac{N(s_c)}{\sum_{s_{\tilde{c}} \in \mathcal{C}(s)} N(s_{\tilde{c}})}.
\end{align*}

\textbf{True Reward Propagation} We improve the quality of the q-values by propagating the real final rewards back through the tree when a terminal state $s_{T+1}$ is reached.
During training, terminal nodes can be evaluated against a reference golden answer $g^*$ using an external verifier. 
For actions leading to terminal states, the associated reward is equal to the q-value (see Eq.~\ref{eq_qv}). Therefore, instead of using the prediction of the q-value model, we initialize $q(s_{T+1})$ with the true reward $r(s_{T+1})$ based on the evaluation of the external verifier.
The reward is then backpropagated via the q-values through the tree with our backpropagation operator. This adjustment anchors the q-value model predictions with real reward signals and prevents the q-value model predictions to diverge. 


\textbf{Best Path Selection} After $N$ iterations, MCTS will have formed a tree in which every path corresponds to one of the explored reasoning trajectories. The final reasoning step in a path with the highest terminal value estimate is returned as the final solution.



\begin{algorithm}[!htbp]
\caption{MCTS for Reasoning \textcolor{red}{(Training mode in red)}}
\label{alg:mcts_star}
\textbf{Input:} Policy model $\pi_\theta$, value model $Q_\phi$, question $z_0$, \textcolor{red}{golden answer $g^*$, binary correctness verifier $\Gamma$}, number of MCTS iterations $N$, number of children expanded in every selection phase $M$, exploration constants $c_1, c_2$, Backpropagation weight $\alpha$.\\
\textbf{Output:} Search tree $\mathcal{T} = (\mathcal{N},\mathcal{E})$ containing the best path $\tau^{*}$.
\begin{algorithmic}[1]
\STATE $s_0 \gets (z_0)$ \hfill\COMMENT{\textit{Initialize root node}}
\STATE $N(s_0) = 0$
\STATE $\mathcal{N} \gets \{s_0\}$ \hfill\COMMENT{\textit{Initialize node set}}
\STATE $\mathcal{E} \gets \emptyset$ \hfill\COMMENT{\textit{Initialize edge set}}
\STATE $i \gets 1$
\WHILE{$i\leq N$ or $\beta(s) \neq 1$}
    \STATE $s \gets s_0$\hfill \COMMENT{\textit{Start from root node}}
    \STATE \textbf{-------------- Selection ------------------------------------------}
    \WHILE{$s$ is not a leaf node}
        \STATE \hfill\COMMENT{\textit{Select child $s_c \in \mathcal{C}(s)$ with highest selection score}}
        \STATE $s_c \gets \argmax\limits_{s_c\in\mathcal{C}(s)} q(s_c)+\frac{\sqrt{N(s)-1}}{1 + N(s_c)} \left(c_1 + \log{\frac{N(s) +     c_2}{c_2}} \right)$ 
        \STATE $s \gets s_c$ \hfill\COMMENT{\textit{Move to the selected child}}
    \ENDWHILE
    \STATE \textbf{-------------- Expansion -----------------------------------------}
    \FOR{$j=1$ to $M$}
        \STATE $z_c \gets (t^1,\ldots t^{M_{z_c}}) \sim \pi_\theta$\COMMENT{\textit{Sample a new reasoning step}}
        \STATE $s_c \gets s \frown z_c $  \hfill\COMMENT{\textit{Append $z_c$ to the current state $s$}}
        \STATE $q(s_c) \gets Q_\phi(s)$ \hfill \COMMENT{\textit{Predict with the Q-VM}} 
        \STATE $N(s_c) \gets 1$ \hfill \COMMENT{\textit{Initialize visit count}}
        \STATE $\beta(s_c) \gets 0$  \hfill\COMMENT{\textit{Initialize terminality function}}
        \IF{$s_c$ terminal}
        \STATE $\beta(s_c) \gets 1$ \hfill\COMMENT{\textit{Mark as terminal}}
        \textcolor{red}{\STATE
        $
        r(s_c) \gets
        \begin{cases} 
        1, & \text{if } \Gamma(s_c, g^*) = 1, \\
        -1, & \text{if } \Gamma(s_c, g^*) = 0.
        \end{cases}
        $
        \COMMENT{\textit{Check for correctness to determine the reward}}
        \STATE $q(s_c) \gets r(s_c)$ \hfill\COMMENT{\textit{Overwrite by true reward}}}
        \ENDIF
        \STATE $\mathcal{N} \gets \mathcal{N} \cup \{s_c\}$ \hfill\COMMENT{\textit{Add the node to the tree}}
        \STATE $\mathcal{E} \gets \mathcal{E} \cup \{(s, s_c)\}$ \hfill\COMMENT{\textit{Add the edge to the tree}}
    \ENDFOR
    \STATE \textbf{-------------- Backpropagation --------------------------------}
    \WHILE{$s \neq s_0$}
        \STATE $N(s) \gets N(s) + 1$ \hfill\COMMENT{\textit{Update the visit count}}
        \STATE $q(s) \gets (1-\alpha)q(s) + \alpha \gamma \sum_{s_c\in\mathcal{C}(s)}w_s(s_c) q(s_c)$ \STATE\hfill\COMMENT{\textit{Update the value}}
        \STATE $s \gets s_p$ \hfill\COMMENT{\textit{Move to the parent}}
    \ENDWHILE
    \STATE $i\gets i + 1$
\ENDWHILE
\STATE \textbf{Best Path Selection:}
\STATE Select the best reasoning sequence $s^{*}_T$. 
\STATE \RETURN $s^{*}_T$, all reasoning sequences $\{s^{(i)}_j\}_j$
\end{algorithmic}
\end{algorithm}

\subsection{Training Phase 1}
\label{app:phase1_training}

To adequately employ the MCTS-based reasoning scheme introduced in Appendix~\ref{sec:mcts_algo_description}, the policy model must be fine-tuned to generate responses in the format of semantically-relevant reasoning steps. The value model -- a q-value model in our case -- must be trained to accurately estimate the values of the sequences of reasoning steps.

\begin{figure*}[t]
    \centering
\includegraphics[width=0.95\textwidth]{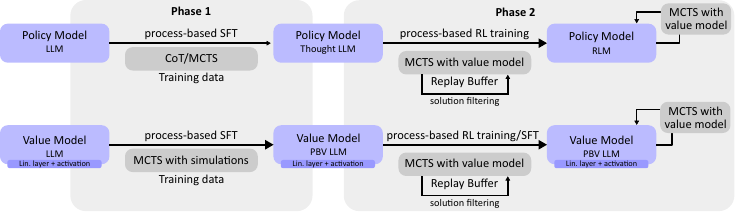}
    \caption{\textbf{The two phases of the training pipeline.}}
    \label{fig:two-phase-training}
\end{figure*}

We propose a two-phase training approach designed to let the policy effectively leverage the structured exploration and iterative refinement capabilities of the search process to generate optimal sequences of reasoning steps. A detailed algorithmic description of the pipeline is in Figure~\ref{fig:two-phase-training}.

The first phase focuses on preparing the policy and value models to generate and evaluate reasoning trajectories effectively. This is achieved by supervised fine-tuning (SFT) training on a dataset of example sequences of reasoning steps (where intermediate reasoning steps are terminated by an "End of Intermediate Step" eois token). The objective is twofold: (1) to fine-tune the policy model $\pi_\theta$ to produce semantically coherent reasoning steps, and (2) to train the q-value model $Q_\phi$ to accurately assign scalar scores to reasoning trajectories, distinguishing between high-quality and suboptimal reasoning paths.

This supervised fine-tuning phase ensures that the policy model can generate reasoning steps consistent with the structured format required for downstream MCTS-based exploration, while the q-value model provides reliable evaluations of intermediate and terminal states. Together, these components form the foundation for the subsequent online reinforcement learning in the second phase, where the policy and q-value models are further refined through interaction with the reasoning framework.

\subsubsection{Datasets Generation and Preparation}

Performing SFT of the policy model requires a dataset of high-quality reasoning sequences denoted as $ D_{\text{SFT}}=\{ \left( x_{\text{SFT}}^{(i)},\, y_{\text{SFT}}^{(i)}\right)\}$. Each pair in the dataset consists of a prompt $x_{\text{SFT}}^{(i)}$ composed of a sequence of reasoning steps (for example $x_{\text{SFT}}^{(i)} = (z_0^{(i)}, ..., z_j^{(i)})$), and a target completion $y_{\text{SFT}}^{(i)} = z_{j+1}^{(i)}$ which is the subsequent reasoning step or final answer.
Appendix \ref{sec:appendix_data} contains a detailed account of the dataset creation and processing. It covers how the special eois token is appended to reasoning steps to mark the end of a step during inference.

Similarly to the policy model, training the q-value model requires a supervised dataset of reasoning sequences and corresponding scores. We denote this dataset $D_{\text{QVM-train}}=\{ ( x_{\text{QVM-train}}^{(i)}, y_{\text{QVM-train}}^{(i)} ) \}$, with reasoning sequences
$x_{\text{QVM-train}}^{(i)} = (z_0^{(i)}, ..., z_t^{(i)})$ and target q-value $y_{\text{QVM-train}}^{(i)}$. Appendix \ref{sec:appendix_data} explains how this dataset can be generated using an initial list of questions, a base LLM for querying, and a verifier program to label reasoning sequences as conducive to a correct final answer or not. 

\begin{algorithm}[!htbp]
  \caption{SFT of Policy Model $\pi_\theta$ (completion-only)}
  \label{algo:sft_policy_model}
  \textbf{Input:} Policy Model $\pi_\theta$, tokenized dataset $D_{\text{SFT}} = \{(x^{(i)}, y^{(i)}) \}$, training hyperparameters (optimizer, learning rate $\eta$, batch size $B$, and maximum number of epochs $E$).\\
  \textbf{Output:} Fine-tuned policy model $\pi_\theta$.\\
  \begin{algorithmic}[1]
  \FOR{epoch $e = 1$ to $E$}
    \STATE Shuffle dataset $D_{\text{SFT}}$.
    \STATE Divide $D_{\text{SFT}}$ into batches $\{\mathcal{B}_k\}$ of size $B$.
    \FOR{each batch $\mathcal{B}_k$}
      \STATE Initialize batch loss: $\mathcal{L}_{\text{batch}} = 0$.
      \FOR{each sample $(x^{(i)}, y^{(i)}) \in \mathcal{B}_k$}
        \STATE Iteratively predict completion tokens: \[\hat{y}_t^{(i)} \sim \pi_\theta(x^{(i)}_{1:t-1}), \]
        where $x^{(i)}_{1:t-1}$ represents the context (prompt + previously predicted tokens).
        \STATE Compute CE loss for each completion token:\newline
        $
        \mathcal{L}^{(i)} = -\sum_{t=1}^{|y^{(i)}|} \log P(\hat{y}_t^{(i)} = y_t^{(i)} | x^{(i)}, \pi_\theta).
        $
        \STATE Accumulate the loss: $\mathcal{L}_{\text{batch}} \mathrel{+}= \mathcal{L}^{(i)}$.
      \ENDFOR
      \STATE Normalize batch loss: $
      \mathcal{L}_{\text{batch}} = \mathcal{L}_{\text{batch}} / |\mathcal{B}_k|$.
      \STATE Backpropagate gradients, update $\theta$ via optimizer.
    \ENDFOR
  \ENDFOR
  \end{algorithmic}
\end{algorithm}
 
\subsubsection{SFT of the Policy Model}
The supervised fine-tuning of the policy model, which we illustrate in Algorithm~\ref{algo:sft_policy_model}, is performed on the dataset $D_{\text{SFT}}$ of prompts and target completions of the next reasoning step. 
The policy $ \pi_\theta $ is instantiated as a general pre-trained LLM. 
Specifically, we perform 'completion-only' SFT such that for every (prompt, target completion) pair, the base model is trained to minimize the cross-entropy loss between its predicted token probabilities and the ground truth target completion.

\subsubsection{Q-Value Model Training}
The q-value model $Q_\phi$ is trained on $D_\text{QVM-train}$ to assign appropriate scalar scores to the candidate reasoning trajectories with the details being shown in Algorithm~\ref{algo:finetune_value_model}. It is instantiated as a pre-trained LLM with additional linear layers and to which a shifted and rescaled classification head is added; we denote all of its trainable weights as $\phi$. Depending on the reward design, the q-value model can be trained via scalar (least squares) regression if continuous rewards are chosen, or with a classification objective such as the Binary Cross-Entropy (BCE) loss, if trajectories are labeled with binary rewards or as chosen-rejected preference pairs.

By the end of training, $ Q_\phi $ should output accurate q-value scores, which will later guide policy refinement in the second phase and improve the search accuracy when used in the MCTS.

\begin{algorithm}[!htbp]
  \caption{Fine-tuning of the Q-Value Model $Q_\phi$}
  \label{algo:finetune_value_model}
  \textbf{Input:} Q-value model $Q_\phi$ (Q-VM), dataset $D_{\text{QVM-train}} = \{(x^{(i)}, y^{(i)}) \}$, training hyperparameters (optimizer, learning rate $\eta$, batch size $B$, and maximum number of epochs $E$).
  \newline \textbf{Output:} Fine-tuned q-value model \(Q_\phi\).\\
  \begin{algorithmic}[1]
  \FOR{epoch $e = 1$ to $E$}
    \STATE Shuffle the dataset $D_{\text{QVM-train}}$.
    \STATE Divide $D_{\text{QVM-train}}$ into batches $\{\mathcal{B}_k\}$ of size $B$.
    \FOR{each batch $\mathcal{B}_k$}
      \FOR{each sample $(x^{(i)}, y^{(i)}) \in \mathcal{B}_k$}
        \STATE Predict the q-value with QVM $\hat{y}^{(i)} = Q_\phi(x^{(i)})$. 
        \STATE \COMMENT{\textit{Compute the loss:}}
        \IF{Regression Loss}
          \STATE $\mathcal{L} = \frac{1}{B} \sum_{(x^{(i)}, y^{(i)})} (\hat{y}^{(i)} - y^{(i)})^2$.
        \ENDIF
        \IF{Classification Loss}
          \STATE $\mathcal{L} = \frac{1}{B} \sum_{(x^{(i)}, y^{(i)})} \texttt{BCE}(\hat{y}^{(i)}, y^{(i)})$.
        \ENDIF
        \STATE Backpropagate gradients, update $\phi$ via optimizer.
      \ENDFOR
    \ENDFOR
  \ENDFOR
  \end{algorithmic}
\end{algorithm}


\subsection{Training Phase 2}
\label{sec:phase2-appendix}

Phase 2 involves generating reasoning sequences from the policy model with MCTS and the q-value model, and fine-tuning the policy model with an RL-based alignment algorithm to generate better completions. The q-value model must also be continually updated in this training loop to keep in-distribution with the policy model's outputs. Sufficient pre-training of the policy and q-value models in the first phase is crucial to ensure stable training of these models in the second phase. The MCTS structure which provides a balanced exploration-exploitation search combined with repeated sampling of the policy model ensures sufficient exploration during this online-RL phase. This final training phase returns the fine-tuned policy and q-value models.


\subsubsection{Training Algorithm}


Phase 2 uses a set $D_p = \{p^{(i)}\}$ of prompt questions, which may be isolated from the phase 1 dataset $D_{\text{SFT}}$. The training process, displayed in Algorithm~\ref{algo:rl_finetuning}, involves a repetition of a MCTS rollout phase followed by a training (reinforcement) phase.

To obtain data for the training, a MCTS tree $\mathcal{T}^{(i)}$ is build w.r.t. each question $p^{(i)}$ using Algorithm~\ref{alg:mcts_star} in training mode. 
The set of hyperparameters for MCTS $\Xi_{MCTS}$, denotes the number of MCTS iterations $N$ (per question), the number of children expanded in every selection phase $M$, exploration constants $c_1, c_2$, and backpropagation weight $\alpha$.
To enhance the quality of the data, we prune the generated MCTS tree $\tilde{\mathcal{T}}^{(i)} = (\tilde{N}^{(i)}, \tilde{E}^{(i)})$ to only include paths that reached a terminal state since only these paths received the reward. Then, we extract all nodes and a set of node characteristics from the pruned tree. The dataset comprises of state, action and q-value triplets of the pruned tree: $ \{(s_j^{(i)}, z_j^{(i)}, q(s_j^{(i)})\}_{s_j \in \tilde{N}^{(i)}}$.
The data is stored in a replay buffer $\mathcal{R}$.

The reinforcement phase samples a batch of reasoning sequences from the replay buffer. From each trajectory, constituent states, actions, value estimates and the corresponding values attributed during MCTS are used to perform RL training (for example with PPO or reinforcement). Alternative schemes may involve selecting preference pairs among trajectories and then aligning the policy using DPO, or simply selecting the most desirable trajectory per question and performing further SFT training.

During this reinforcement phase, the value model is updated to mimic the (backpropagated) values from the MCTS process, which we illustrate in Algorithm~\ref{algo:value_model_update}.

\begin{algorithm}[!tbp]
  \caption{Phase 2: RL of the Policy and Q-Value Model}
  \label{algo:rl_finetuning}
  \textbf{Input:} Policy $\pi_\theta$, q-value model $Q_\phi$, dataset $D_p = \{p^{(i)}\}$, MCTS hyperparameters $\Xi_{MCTS}$. \newline
  \textbf{Output:} Trained $\pi_{\theta}$ and updated $Q_{\phi}$.\\
  \begin{algorithmic}[1]
  \FOR{each training iteration}
      \STATE \textbf{
      -------------- Rollout ---------------------------------}
      \FOR{each question $p^{(i)} \in D_p$}
        \STATE \COMMENT{\textit{Generate MCTS tree with $\pi_\theta$ and $Q_\phi$  (\underline{Algorithm~\ref{alg:mcts_star}})}}
        \STATE $\mathcal{T}^{(i)} \gets MCTS(p^{(i)}, Q_\phi,\pi_\theta, \Xi_{MCTS})$ 
        \STATE \COMMENT{\textit{Remove incomplete paths from the tree}}
        \STATE $\tilde{\mathcal{T}}^{(i)} \gets Prune(\mathcal{T}^{(i)})$ 
        \STATE \COMMENT{\textit{Extract nodes and values, store them in replay buffer}}
        \STATE  $\mathcal{R} \gets \mathcal{R} \cup \{(s_j^{(i)}, z_j^{(i)}, q(s_j^{(i)})\}_{s_j \in \tilde{N}^{(i)}}$ 
        
      \ENDFOR
      \STATE \textbf{
      -------------- Training ---------------------------------}
      \FOR{each epoch}
        \STATE Sample a batch $\mathcal{B}$ from replay buffer $\mathcal{R}$.
        \STATE Update policy $\pi_\theta$ (\underline{Algorithm~\ref{algo:policy_update}}).
        \STATE Update q-value model $Q_\phi$ (\underline{Algorithm~\ref{algo:value_model_update}}).
      \ENDFOR
    \ENDFOR
  \end{algorithmic}
\end{algorithm}

\subsubsection{Policy Update}
The policy update, which we present in Algorithm~\ref{algo:policy_update}, is performed on a batch $\mathcal{B}$ of reasoning sequences. As mentioned above, the reasoning sequences can be decomposed into state-action-value triplets to then perform RL training.
We will discuss now three different methods (standard RL, preference-based RL, and SFT training) to improve the policy model during the second phase training.

\begin{algorithm}[!htbp]
  \caption{Policy Update (PPO, DPO, or SFT)}
  \label{algo:policy_update}
  \textbf{Input:} Batch \(\mathcal{B}\), policy \(\pi_\theta\), reference policy \(\pi_{\text{ref}}\), learning rate \(\eta\), clipping parameter \(\epsilon\), preference data \(\mathcal{B}_{\text{pref}}\) for DPO.
  \newline
  \textbf{Output:} Updated policy \(\pi_\theta\).
  \newline
  \begin{algorithmic}[1]
    \STATE \textbf{
      -------------- Train via PPO ---------------------------------}
    \STATE Select state-action-value triplets from sequences in 
    $\mathcal{B}$
    \FOR{each \((s_t, a_t, q_t) \in \mathcal{B}\)}
      \STATE Compute the policy ratio: \(r_\theta = \frac{\pi_\theta(a_t|s_t)}{\pi_{\theta_{\text{ref}}}(a_t|s_t)}\).
      \STATE Compute the advantages $\hat{A}(s_t)$ (\underline{Algorithm~\ref{algo:advantage_calc}}).
      \STATE Compute the PPO loss: \newline 
      \(
      \mathcal{L}_{\text{PPO}} = \min(r_\theta \hat{A}(s_t), \text{clip}(r_\theta, 1-\epsilon, 1+\epsilon) \hat{A}(s_t))\).
    \ENDFOR
    \STATE (Optional) Add KL divergence or entropy regularization:
    $\mathcal{L}_{\text{PPO}} \leftarrow \mathcal{L}_{\text{PPO}} + \lambda_{\text{KL}} \texttt{KL}(\pi_\theta || \pi_{\text{ref}}) + \lambda_H \mathcal{L}_{H}$.
    \STATE Perform gradient update to refine \(\pi_\theta\).
  \STATE
  \STATE \textbf{
      -------------- Train via DPO (pairwise preferences)  ------}
    \STATE Select preference pairs of reasoning sequences in $\mathcal{B}$
    \FOR{each pair \((s^+, s^-) \in \mathcal{B}_{\text{pref}}\)}
      \STATE Compute DPO objective:
      \[
        \mathcal{L}_{\text{DPO}} = \frac{1}{|\mathcal{B}_{\text{pref}}|} \sum_{(s^+, s^-)} \log \sigma \left(\beta \left( \log \frac{\pi_\theta(s^+)}{\pi_\theta(s^-)} \right)\right).
      \]
    \ENDFOR
  \STATE Perform gradient update to refine \(\pi_\theta\).
  \STATE
  \STATE \textbf{
      -------------- Train via SFT (single target sequence)  ------}
    \STATE Select high-value reasoning sequences \(s^+\) from $\mathcal{B}$
    \FOR{each reasoning sequence \(s^+\)}
      \STATE Perform SFT on \(s^+\)
    \ENDFOR
  \end{algorithmic}
\end{algorithm}

\textbf{Standard Policy Gradient Methods} such as Proximal Policy Optimization (PPO)~\cite{schulman2017proximal} or REINFORCE~\cite{ahmadian2024back, sutton2015reinforcement} are particularly suited for tasks where trajectories are collected (online) and reliably evaluated by the q-value model $Q_\phi$.

PPO relies on the computation of trajectory (reasoning sequence) advantages \(\hat{A}(s_t)\), which quantify how much better or worse an action taken in a given state is compared to the expected baseline value of that state. The advantage function is estimated by:
\begin{equation*}
\hat{A}(s_t) = R_t + \gamma V(s_{t+1}) - V(s_t),
\end{equation*}
where \(R_t\) is the immediate environment reward at step \(t\), \(V(s_t)\) is the state value of state \(s_t\), and \(\gamma\) is the discount factor. We can derive the state value easily from the q-values obtained via the q-value model or the running estimates in the MCTS as follows:
\[
V(s_{t+1}) = \frac{1}{\gamma}Q_{\phi}(s_t, a_t),
\]
since rewards are sparse. The standard PPO approach trains the critic model from scratch on bootstrapped rewards for this purpose. We introduce an alternative advantage computation scheme that leverages the backpropagated values from the MCTS in conjunction with $Q_\phi$, as detailed in Algorithm~\ref{algo:advantage_calc}. This integration combines MCTS's exploration and evaluation capabilities with the RL update, enhancing robustness and efficiency in reasoning tasks.

Further regularization can be imposed on the PPO training procedure. To align the policy $\pi_\theta$ with a reference policy $\pi_{\text{ref}}$ (usually instantiated as $\pi_\theta$ before the second phase) during training, the KL divergence $\texttt{KL}(\pi_\theta || \pi_{\text{ref}})$
between the two distributions can be added to the training loss. Additionally, to maintain the diversity of policy generations (and exploration during training), the entropy of the policy distribution can be enhanced by subtracting it from the loss.
The entropy penalty is estimated over a batch $\mathcal{B}$ of state-action pairs $(s,a)$, where $s$ denotes a reasoning sequence and $a$ the next reasoning step. The entropy of a single completion $a$ is computed by summing the entropy of its individual tokens $a_{1:|a|}$ of $a$: \[  \mathcal{L}_{H} = -\frac{1}{|\mathcal{B}|} \sum_{(s, a) \in \mathcal{D}} \sum_{a_i \in a} \pi_\theta(a_i|[s, a_{1:{i-1}}]) \log \pi_\theta(a|[s, a_{1:{i-1}}]). \]

\textbf{Direct Preference Optimization (DPO)}~\cite{rafailov2023direct} aligns the policy to user preferences expressed as pairwise comparisons between reasoning sequences. Given pairs \((s^+, s^-)\), where \(s^+\) is preferred over \(s^-\). This method may not require a process-based reward/value model. The loss involves the sigmoid function which we denote as $\sigma$.

\textbf{Supervised Fine-Tuning (SFT)}
As a straightforward alternative to RL, high-value reasoning sequences can be selected to perform SFT, i.e. train the policy to maximize the likelihood of these reasoning steps. The high-value reasoning sequences may be selected as terminal nodes having the highest q-value, or highest aggregated intermediate-step values. This approach is inspired by AlphaZero-like frameworks, focusing on iteratively refining the policy to generate high-quality reasoning trajectories without requiring explicit rewards.


\subsubsection{Advantage Calculation}

While standard advantage computation in PPO (e.g., via Generalized Advantage Estimation (GAE) \cite{schulman2017proximal}) is widely applicable, we propose an alternative approach tailored to our reasoning framework in Algorithm \ref{algo:advantage_calc}. Specifically, for each state/node $s$, we leverage the q-value estimates $q(s)$
obtained during the MCTS process. They were updated in the backpropagation phase to provide a more informed estimate of the q-values incorporating the estimates of the children and potentially true reward signals from terminal paths in the tree. We expect these MCTS-derived values to be more reliable as they incorporate the ground truth terminal reward, propagated back through the tree, ensuring that a node's value reflects both its immediate reward and the aggregated values of subsequent child states.


\begin{algorithm}[!tbp]
  \caption{Advantage Calculation in MCTS Framework}
  \label{algo:advantage_calc}
  \textbf{Input:} MCTS Tree $\mathcal{T}= (N,E)$, node statistics: rewards and q-values, q-value model $Q_\phi$, discount factor $\gamma$, and $\lambda$.\newline
  \textbf{Output:} Advantages $\{\hat{A}(s_t)\}$.\newline
  \begin{algorithmic}[1]
  \FOR{each node $s_i \in N$}
      \STATE Compute state values: $v_{s_{i+1}}^{\text{MCTS}} = \frac{1}{\gamma} q^\text{MCTS}(s_i)$
      \STATE Compute state values: $v_{s_{i}}^{\text{MCTS}} = \frac{1}{\gamma} q^\text{MCTS}(s_{i-1})$
      \STATE Compute the advantage on the TD error: $\hat{A}(s_i) = r(s_i,a_i) + \gamma v_{s_{i+1}}^{\text{MCTS}} - v_{s_i}^{\text{MCTS}}$.
  \ENDFOR
  \end{algorithmic}
\end{algorithm}

\subsubsection{Q-Value Model Update}
During the second training phase, the q-value model $Q_\phi$ is also updated (see Algorithm~\ref{algo:value_model_update}) to track the MCTS-backpropagated value estimates $q^{\text{MCTS}}(s_t)$,
which should be of higher quality (thanks to the final answer verifier and score aggregation from child nodes).
For each state-action pair $(s,a)$, we train the q-value model $Q_\phi$ via squared error minimization, to match its q-value $Q_\phi(s,a)$ as closely as possible to the corresponding MCTS-value $q^{\text{MCTS}}(s')$, i.e. updated q-value of action $a$ taken in state $s$ leading to state $s'$.

This has the benefit of both improving the accuracy of the value model, and keeping it "in-distribution" with the new policy outputs during this online-RL training.

\begin{algorithm}[!tbp]
  \caption{Q-Value Model Update}
  \label{algo:value_model_update}
  \textbf{Input:} Batch $\mathcal{B}$, q-value model $Q_\phi$, learning rate $\eta$.\newline
  \textbf{Output:} Updated $Q_\phi$.
  \begin{algorithmic}[1]
    \STATE Compute loss:\newline$\mathcal{L}_q = \frac{1}{|\mathcal{B}|} \sum_{(s,a,s')} (Q_\phi(s,a) - q^{\text{MCTS}} (s'))^2$. 
  \STATE Perform gradient update on $\mathcal{L}_q$.
  \end{algorithmic}
\end{algorithm}

\section{Data Generation}
\label{sec:appendix_data}

\subsection{Phase 1 Policy Model Training}


The objective of this training process is to introduce a new 'End of Intermediate Step' (eois) token that serves to delimit individual reasoning steps while preserving the original distribution of the model as much as possible. To achieve this, the model is trained on data generated by itself using greedy decoding.

The training data is derived from eight chain-of-thought (CoT) completions generated for 1,000 questions sampled from the training split of the MATH dataset \cite{hendrycks2021math}. These completions are produced using the same model intended for subsequent training with greedy decoding. During this generation process, the reasoning steps in the data are observed to be separated by two consecutive '\textbackslash n\textbackslash n'. This observation informs the method of delimitation used to construct pairs of questions and their corresponding sequences of reasoning steps.

For each data point, consisting of a question prompt and its associated target response comprising multiple reasoning steps $(q^{(i)}, [z_1^{(i)}, \dots, z_{T+1}^{(i)}])$, additional tokens are introduced to explicitly mark the boundaries of the reasoning steps. Specifically, the 'End of Intermediate Step' (eois) token is defined and inserted after each reasoning step $z_j^{(i)}$, resulting in a modified step $z_j^{(i)*}$. Additionally, the 'End of Sequence' (eos) token is appended to the final reasoning step $z_{T+1}^{(i)}$, yielding $z_{T+1}^{(i)*} = [z_{T+1}^{(i)}; \text{eos}]$. This augmentation ensures that the model can consistently identify when a final solution has been reached during inference.

For LLaMA models, it has been empirically observed that introducing an 'assistant' token after each reasoning step enhances the model's effective utilization of the eois token. However, this behavior may not generalize to other base models, necessitating careful consideration when applying this approach.

Accordingly, the target sequence for supervised fine-tuning (SFT) is constructed as:
\[
y_{\text{SFT}}^{(i)} = [z_1^{(i)}, \text{eois}, \text{assistant}, z_2^{(i)}, \dots, z_{T+1}^{(i)}, \text{eos}].
\]

This approach yields a training dataset comprising pairs of prompts and their corresponding target completions, formally represented as:
\[
D_{\text{SFT}} = \{(q^{(i)}, y_{\text{SFT}}^{(i)})\}.
\]

\subsection{Phase 1 Value Model Training}
\label{value_model_phase1_data}

The original MCTS framework relies on simulations to evaluate a state. Given the state, $n$ rollouts are performed till a terminal state is reached. The terminal states usually can be evaluated (e.g., in math by comparing it with the golden answer). This enables the distribution of  terminal rewards based on their success which are then aggregated to provide a value estimate of the state. These Monte Carlo simulations serve as an estimate of a state's ability to lead to a correct answer. The value estimated in this manner corresponds to the expected cumulative future reward for a given state:
\[
V_{\pi_\theta}(s) = \mathbb{E}_{\tau\sim\pi_\theta} \left[ \sum_{t=i}^{T} \gamma^{t-i} r(s_t, a_t) \mid s_i = s \right],
\]
where \( T \) is the terminal step of the partial reasoning chain \( \tau = (s_i, a_i, r_i, s_{i+1}, \ldots, s_T, a_T, r_T, s_{T+1}) \).

Since rewards are sparse (i.e., \( r(s_t, a_t) = 0 \) for all \( t < T \)), the value function simplifies to:
\[
V_{\pi_\theta}(s_t) = \mathbb{E}_{\pi_\theta}\left[\gamma^{T-t} r(s_{T}, a_T) \mid s_t\right].
\]

This represents the expected terminal reward, which can be empirically estimated using Monte Carlo (MC) estimates:
\[
V_{\pi_\theta}(s_t) \approx \frac{1}{N} \sum_{i=1}^N \gamma^{T-t} r(s_{T}^{(i)}, a_T^{(i)}) := \hat{V}(s_t),
\]
where \( N \) is the number of sampled reasoning chains, and \( s_T^{(i)}, a_T^{(i)}, s_{T+1}^{(i)} \) denote the last transition of the simulation trajectory \( \tau^{(i)} = (s_t, a_t^{(i)}, s_{t+1}^{(i)}, \ldots, s_T^{(i)}, a_T^{(i)}, s_{T+1}^{(i)}) \) for \( i \in \{1, \ldots, N\} \).

To avoid sample inefficiencies and high computational burdens, AlphaGo Zero~\cite{silver2017mastering} and AlphaZero~\cite{silver2018general} introduce a value model to replace simulations by using its predictions for a state. We follow this approach by defining a process-based value model \( V_\phi \). Notably, we train this model with simulation data (instead of true value functions), thereby building a model that predicts state value function estimates \( \hat{V} \). We denote this model as \( \hat{V}_\phi \), parameterized by \( \phi \).

Given that the input of a value model is a sequence of reasoning steps - therefore a sequence of tokens, the natural value model architecture is to use a language model on which one adds linear layer(s) and a suitable output activation function. Typically, it is designed to output a scalar value \( \hat{V}_\phi(s_t) \in  \mathcal{C}\subseteq\mathbb{R} \).

The core distinction between different modeling approaches to state value functions lies in how rewards are modeled. Depending on whether a binary reward setting or a continuous (bounded) one is used, the aggregation mechanism, model architecture, training loss, and interpretation of the predictions vary. We provide an overview of both scenarios and, although often omitted for simplicity, we consider both \( \gamma = 1 \) and \( \gamma \in (0,1] \) for continuous rewards in our analysis.

\subsubsection{Binary Rewards: Modeling the Likelihood of a Correct Terminal State}

For this approach the rewards are modeled binary, therefore \( r(s_T, a_T) = +1 \) for correct solutions and \( r(s_T, a_T) = 0 \) for incorrect solutions. We will adopt a discount factor of \( \gamma = 1 \) which we will see aligns more with the interpretation this reward model provides and is widely adopted in literature. This approach corresponds to the value model proposed in AlphaGo Zero~\cite{silver2017mastering}.

\paragraph{State Value Estimation} 
The value function then further simplifies to:
\[
V_{\pi_\theta}(s_t) = \mathbb{E}_{\pi_\theta}\left[r(s_T, a_T) \mid s_t\right] = \mathbb{P}_{\pi_\theta}\left(r(s_T, a_T) = 1 \mid s_t\right)
\]
This formulation represents the probability of reaching a correct terminal state from a given state \( s_t \). Empirically, this probability is estimated using simulations as follows:
\[
V_{\pi_\theta}(s_t) \approx \frac{\#\text{correct simulations}}{\#\text{simulations}} := \hat{V}(s_t).
\]

\paragraph{Data Generation} To generate labels for estimating the state-value function during the training of a value model, we use MCTS with simulations till a terminal node is reached and calculate the ratio between the number of correct simulations to the number of simulations. There is one very important detail, for a trajectory \( \tau = (s_i, a_i, r_i, s_{i+1}, \ldots, s_{T+1}) \) where $s_{T+1}$ is a terminal state. 
By definition, the true state value function at $s_{T+1}$ is zero. However, in training the value model, we avoid instructing it to output zero for terminal states. 
Instead, in a supervised learning setting, we can identify terminal states and directly compare the model's predictions against the known correct outcomes (golden answers). This comparison negates the need to rely solely on the value model to estimate the value of terminal states or to determine the reward associated with transitioning into these states. 
During inference, while we can still recognize terminal states, we cannot evaluate them by comparing the model's output to a golden answer. Therefore, an alternative metric is necessary. 
We train the value model to predict whether transitioning to $s_{T+1}$ leads to a correct terminal outcome. 
By learning the relationship between a node's content and the correctness of the resulting terminal state, the model can estimate the likelihood that a terminal state leads to a correct answer.
To approximate the terminal reward during inference, we define: $r(s_T, a_T, s_{T+1}) \approx \mathbbm{1}_{[0.5,1]}(\hat{V}_\phi(s_{T+1}))$.
Here $\hat{V}_\phi(s_{T+1})$ represents the value predicted by the value model for the terminal state $s_{T+1}$. If this predicted likelihood exceeds a threshold (e.g., 0.5), we assign a terminal reward of 1; otherwise, we assign a reward of 0. This approach allows the value model to indirectly influence the terminal reward by predicting the likelihood of a correct outcome. Consequently, during training, terminal rewards serve as labels for terminal states in the value model. It is important to note that $\hat{V}_\phi(s_{T+1})$ is not used in any other context but solely to estimate the terminal reward.
\[
    \hat{V}_\phi(s_{T+1}) \neq \hat{V}(s_{T+1})
\]
This distinction clarifies that the predicted value for the terminal state $\hat{V}_\phi(s_{T+1})$ differs from the standard value function's definition $\hat{V}(s_{T+1})=0$.

\paragraph{Model Training $\hat{V}_\phi: \mathcal{S} \rightarrow [0,1]$} 
When trained with these labels we obtain a value model \( \hat{V}_\phi \), parameterized by \( \phi \), that represents the likelihood of a correct terminal state emanating from state \( s_t \). Therefore, the model will output values between $0$ and $1$. To accommodate the binary classification nature of this task, the model should employ a sigmoid activation function in the output layer. The training objective is then to minimize the binary cross-entropy (CE) loss between the predicted probabilities and the empirical estimates derived from the simulations:
\begin{flalign*}
&\mathcal{L}(\phi) = \\
&-\frac{1}{N} \sum_{i=1}^N \left[ y_i \log(\hat{V}_\phi(s_t^{(i)})) + (1 - y_i) \log(1 - \hat{V}_\phi(s_t^{(i)})) \right],
\end{flalign*}
where \( y_i \in \{0,1\} \) denotes the binary label indicating whether the \( i \)-th simulation resulted in a correct terminal state.

\paragraph{Summary} Employing a binary reward structure offers several benefits. First of all, simplicity since binary rewards simplify the learning process, reducing the complexity associated with continuous reward signals. Moreover, the clear distinction between correct and incorrect states facilitates faster convergence during training making this approach effective. In addition, binary classification is less susceptible to noise in reward signals, ensuring more stable value estimates. Furthermore, this approach aligns with the objectives of reinforcement learning in achieving clear and unambiguous rewards, thereby streamlining the optimization of the policy \( \pi_\theta \).

\subsubsection{Continuous and Bounded Rewards: Modeling the Expected Future Reward}
We model the rewards to be continuous and bounded by allowing values in $[a,b]$:
\[
V_{\pi_\theta}(s_t) \in [a,b].
\]
A common design, is to set the borders to $-1$ and $1$ such that a terminal reward is \( r(s_T, a_T) = +1 \) for correct terminal states and \( r(s_T, a_T) = -1 \) for incorrect states.
This approach models the expected future reward as a continuous and bounded value, capturing the degree of correctness or quality of the terminal state. In contrast to the binary reward structure, continuous and bounded rewards provide a more nuanced representation of the outcomes in reasoning tasks. Note, that without discounting this approach resembles the proposed value model of AlphaZero~\cite{silver2018general}. 

\paragraph{Bounded rewards} By constraining rewards within a predefined interval $[a,b]$, we effectively create a correctness scale where the extremities represent the definitive outcomes of the reasoning process. Specifically, the lower bound $a$ corresponds to reaching an incorrect terminal state, while the upper bound $b$ signifies a correct terminal state. This bounded framework mirrors the spectrum of possible correctness, allowing the model to capture varying degrees of solution quality between these extremes. Such a scale facilitates a more nuanced evaluation of intermediate states, reflecting partial correctness or varying levels of reasoning quality. Moreover, this approach ensures that the reward signals remain interpretable and consistent, fostering a clear distinction between successful and unsuccessful outcomes. 

\paragraph{State Value Estimation} 
With a discount factor \( \gamma \in (0,1]\), the value function is defined as:
\[
V_{\pi_\theta}(s_t) = \mathbb{E}\left[\gamma^{T-t} r(s_T, a_T) \mid s_t\right],
\]
where \( r(s_T, a_T) = b \)  for correct terminal states and \( r(s_T, a_T) = a \) for incorrect ones. 
Empirically, this expectation is approximated by averaging the rewards of the simulations:
\[
V_{\pi_\theta}(s_t) \approx \frac{1}{N} \sum_{i=1}^N \gamma^{T-t} r(s_T^{(i)}, a_T^{(i)}) := \hat{V}(s_t),
\]
where \( N \) denotes the number of sampled reasoning chains, and \( (s_T^{(i)}, a_T^{(i)}, s_{T+1}^{(i)}) \) represent the final transition of the \( i \)-th simulation trajectory \( \tau^{(i)} = (s_t, a_t^{(i)}, s_{t+1}^{(i)}, \ldots, s_T^{(i)}, a_T^{(i)}, s_{T+1}^{(i)}) \) for \( i \in \{1, \ldots, N\} \). If a discount factor is applied $\gamma \in (0,1)$ then each terminal reward is discounted proportional to the number of steps needed to reach the terminal state.
This corresponds to the soft estimation proposed by Wang et al.~\cite{wang2024math}. We want to note that this estimator typically underestimates $V$ due to its proneness to false negatives~\cite{havrilla2024glore, yuan2024free}.

\paragraph{Data Generation} To generate labels for state-value function estimate pairs to train a value model, we use MCTS with simulations and average the outcomes of the simulations. Therefore, at each newly generated node $s$ we simulate till a terminal node is reached and we record the depth - the number of steps needed starting from $s$ (since T is not identical per trajectory). We then record the terminal reward which in our case is \( r(s_T, a_T) = 1 \)  for correct and \( r(s_T, a_T) = -1 \) for incorrect answers. Discounted by the depth we can average these rewards and obtain an estimation of the node value which serves as a label for the initial value model training. 

\paragraph{Model Training \( \hat{V}_\phi: \mathcal{S} \rightarrow [a,b] \)} 
The value model \( \hat{V}_\phi \), parameterized by \( \phi \), is designed to predict the expected terminal reward from any given state \( s_t \). To accommodate the continuous and bounded nature of this task, the model employs a scaled and shifted sigmoid activation function in the output layer, ensuring that the predictions remain within the range \([a,b]\). The training objective is to minimize the mean squared error (MSE) loss between the predicted values and the empirical estimates derived from the simulations:
\[
\mathcal{L}(\phi) = \frac{1}{N} \sum_{i=1}^N \left( \hat{V}_\phi(s_t^{(i)}) - \gamma^{T-t} r(s_T^{(i)}, a_T^{(i)}) \right)^2.
\]
We also experimented with a tanh activation output and a linear layer with clipping of the values. However, both methods proved to be unstable in training in contrast to the scaled and shifted sigmoid layer. A tanh and sigmoid layer naturally bounds the output but also pushes values towards the extremes, enhancing the separation between high and low value estimates. This characteristic can improve the model's ability to distinguish between highly correct and highly incorrect states which is why we are particularly interested in these activation functions.

\paragraph{Discounting} Introducing a discount factor \( \gamma \) aligns the value function with the incremental nature of reasoning tasks. Unlike traditional games, where all moves contribute indirectly and trajectories are not penalized for length, reasoning benefits from discouraging unnecessary or redundant steps. The inclusion of the discount factor \( \gamma \) ensures that rewards achieved sooner have a greater impact on the value function, the model incentivizes reaching correct solutions with fewer steps which ultimately enhances efficiency and suppresses redundancies. Moreover, this models the uncertainty decay in the trajectories; the further into the future a reward lies, the more uncertain its prediction becomes. Discounting naturally reduces the reliance on these uncertain long-term rewards, thereby stabilizing the learning process by focusing on more predictable and immediate outcomes. However, the model's performance becomes sensitive to the choice of \( \gamma \), requiring careful tuning to balance the influence of immediate versus long-term rewards. Balancing the discount factor is essential to ensure that the model effectively captures the importance of both progress and the final correctness of the reasoning chain.

\paragraph{Summary} Employing a continuous and bounded reward structure offers several benefits. Unlike binary rewards, continuous rewards provide a finer distinction between varying degrees of correctness, allowing the model to capture subtle differences in terminal states. Continuous rewards can encode more information about the quality of solutions, facilitating more informed decision-making during the search process. Bounded rewards prevent extreme values, promoting numerical stability and consistent training dynamics. However, it also shows that the choice of reward values and their scaling can significantly impact the learning process, necessitating careful calibration to ensure effective training. 

\subsection{State Action Value Function Modeling}\label{QM_data}

The state-action value function, commonly denoted as \( Q_{\pi_\theta}(s_t, a_t) \), represents the expected cumulative reward of taking action \( a_t \) in state \( s_t \) under policy \( \pi_\theta \). Formally, it is defined in our framework as:
\begin{flalign*}
    &Q_{\pi_\theta}(s_t, a_t) \\
    &= \mathbb{E}_{\tau\sim\pi_\theta} \left[ \sum_{i=t}^{T} \gamma^{i-t} r(s_i, a_i) \mid s_t, a_t \right] \\
    &= r(s_t, a_t) + \gamma \mathbb{E}_{\tau\sim\pi_\theta} \left[ \sum_{i=t+1}^{T} \gamma^{i-(t+1)} r(s_i, a_i) \mid s_t, a_t \right] \\
    &= r(s_t, a_t) + \gamma \mathbb{E}_{s_{t+1}} \left[ V_{\pi_\theta}(s_{t+1}) \mid s_t, a_t \right] \\
    &\stackrel{\text{det. } \mathbb{P}}{=} r(s_t, a_t) + \gamma V_{\pi_\theta}(s_{t+1}),
\end{flalign*}
where \( T \) denotes the terminal step of the trajectory \( \tau = (s_t, a_t, r_t, s_{t+1}, \ldots, s_T, a_T, r_T, s_{T+1}) \). In environments characterized by sparse rewards, where \( r(s_t, a_t) = 0 \) for all \( t < T \), the q-value simplifies to:
\[
Q_{\pi_\theta}(s_t, a_t) = \gamma V_{\pi_\theta}(s_{t+1}).
\]
At terminal states, where the state value \( V_{\pi_\theta}(s_{T+1}) = 0 \), the q-value further reduces to:
\[
Q_{\pi_\theta}(s_T, a_T) = r(s_T, a_T).
\]

\subsubsection{Process-Based Q-Value Modeling}

A process-based q-value model utilizes the same architecture as a process-based value model, typically leveraging a LLM enhanced with additional linear layers and an appropriate output activation function. The output is a scalar value \( \hat{Q}_\phi(s_t, a_t) \in \mathcal{C} \subseteq \mathbb{R} \). Specifically, the q-value model takes a state-action pair—comprising a sequence of past steps and the current action—and predicts the corresponding q-value based on the aforementioned formulation.

\paragraph{Training Data Generation}

To train the q-value model, it is essential to compute the q-values for various state-action pairs. For \( t < T \), q-values can be estimated using \( N \) Monte Carlo simulations as follows:
\begin{align*}
    Q_{\pi_\theta}(s_t, a_t) &= r(s_t, a_t) + \gamma V_{\pi_\theta}(s_{t+1}) \\
    &= \gamma V_{\pi_\theta}(s_{t+1}) \quad \text{(since \( r(s_t, a_t) = 0 \))} \\
    &\approx \gamma \cdot \frac{1}{N} \sum_{i=1}^N \gamma^{T-(t+1)} r(s_T^{(i)}, a_T^{(i)}) \\
    &= \frac{1}{N} \sum_{i=1}^N \gamma^{T-t} r(s_T^{(i)}, a_T^{(i)}) := \hat{Q}(s_t, a_t),
\end{align*}
where \( N \) is the number of sampled reasoning chains, and \( \tau^{(i)} = (s_t, a_t^{(i)}, s_{t+1}^{(i)}, \ldots, s_T^{(i)}, a_T^{(i)}, s_{T+1}^{(i)}) \) represents the \( i \)-th simulation trajectory for \( i \in \{1, \ldots, N\} \). This estimation aligns with the state value estimation under the sparse reward formulation:
\[
\hat{Q}(s_t, a_t) = \hat{V}(s_t).
\]
For \( t = T \), the q-value is directly given by the immediate reward:
\begin{align*}
    Q_{\pi_\theta}(s_T, a_T) &= r(s_T, a_T) = V_{\pi_\theta}(s_{T+1})  \neq \hat{V}(s_T) = 0.
\end{align*}

\paragraph{Reward Modeling} The considerations about reward modeling apply to q-value models as well since these models are trained very similar, so we omit their discussion here.

\subsubsection{The Difference Between Value and Q-Value Models}

The difference of VMs and Q-VMs can be easily shown in how they are used in the evaluation processes of an MCTS algorithm. Q-VMs predict \(\hat{Q}_{\phi}(s_t, a_t)\), which evaluates the action \(a_t\) taken in state \(s_t\) that deterministically transitions to \(s_{t+1}\). Thus, the value \(\hat{Q}(s_t, a_t)\) is used to evaluate adding the node \(s_{t+1}\) to the tree. On the other hand, for VMs, adding a node \(s_{t+1}\) to the tree is determined by \(\hat{V}(s_{t+1}) = \frac{1}{\gamma}\hat{Q}_{\phi}(s_t, a_t)\), where \(\gamma\) is the discount factor.

This distinction is making the training processes different. Note that \(s_t \frown a_t = s_{t+1}\). For Q-VMs, the training tuples are \(((s_t, a_t), \hat{Q}(s_t, a_t)) = (s_{t+1}, \hat{Q}(s_t, a_t))\) due to the deterministic transition. For VMs, the corresponding training tuples are \((s_{t+1}, \hat{V}(s_{t+1}))\). Since we propose training VMs on terminal rewards for terminal states instead of assigning a label of 0, VMs and Q-VMs become equivalent under the following transformation for any \( t \in \{0, \ldots, T\} \) for evaluating adding node $s_{t+1}$:
\begin{align*}
    \hat{V}(s_{t+1}) = \frac{1}{\gamma}\hat{Q}_{\phi}(s_t, a_t).
\end{align*}

We introduced q-value models since they address a critical inconsistency of value models in terminal states. Specifically, while value models assign a flat value of zero to terminal states, q-value models provide a meaningful evaluation of the final action's correctness through \( Q_{\pi_\theta}(s_T, a_T) = r(s_T, a_T) \). This distinction is essential for accurately assessing whether a terminal step leads to a correct or incorrect response during inference.


\ifcnf

\fi

\bibliographystyle{abbrv}
\bibliography{references.complete}

\end{document}